\documentclass[12pt]{article}
\PassOptionsToPackage{numbers,super,sort&compress}{natbib}
\usepackage[preprint]{neurips_2026}
\usepackage[T1]{fontenc}
\DeclareUnicodeCharacter{0394}{\ensuremath{\Delta}}
\DeclareUnicodeCharacter{2212}{\ensuremath{-}}
\usepackage{microtype}
\usepackage{hyperref}
\usepackage{url}
\usepackage{multirow}
\usepackage{booktabs}
\usepackage{graphicx}
\usepackage{amsmath}
\usepackage{enumitem}
\usepackage[font=small,labelfont=bf]{caption}
\usepackage{placeins}
\usepackage{float}

\setcitestyle{super,open={},close={},citesep={,}}
\usepackage{lineno}
\usepackage{rotating}
\title{Reported Confidence in LLMs Tracks Commitment More Than Correctness}

\author{%
  Dharshan Kumaran\thanks{Corresponding author: \texttt{dkumaran@google.com}} \\
  Google DeepMind \\
}

\begin{document}
\maketitle
%\linenumbers
%\fontsize{12pt}{12pt}\selectfont
\section*{Abstract}
Confidence is defined as an estimate of the probability that a chosen answer is correct \cite{pouget2016confidence, kepecs2012computational, fleming2017self}. Verbal confidence reports are widely used as uncertainty measures in large language models \cite{xiong2023can, tian2023just, yoon2025reasoning}, but it remains unclear whether they are best understood as estimates of correctness at all. We test this using a two-stage abstention paradigm inspired by neuroscientific studies of perceptual decision making \cite{lak2014orbitofrontal, kepecs2008neural}: a model first answers a question and reports its confidence, then later decides whether to commit that answer to a user or abstain. Across four non-reasoning models, different prompt framings, and confidence formats, verbal confidence predicted the later commit/abstain decision substantially better than it predicted whether the answer was actually correct. Calibrated token log-probabilities showed a contrasting profile: their ability to predict abstention was closely coupled to their ability to discriminate correctness — the signature of an answer-evidence signal. We then asked whether VC's behavioural alignment is reducible to its correctness-tracking, by removing the variance in verbal confidence shared with calibrated log-probabilities and examining the residual. Residual verbal confidence remained preferentially aligned with later commitment behaviour, while its relationship to correctness collapsed to near chance levels. The verbal-confidence dissociation generalised to four reasoning models across four benchmarks of varying difficulty -- including challenging multiple choice questions and frontier-level freeform questions. Mechanistic analyses in Gemma~4 and Gemma~3 provided convergent evidence. A post-answer internal state previously shown to causally support verbal-confidence generation \citep{kumaran2026llms} already encoded the model's future abstention decision before the abstention prompt was given. This commit-readiness representation was organised primarily by the model's later abstention decision rather than by the correctness of its answer, with the decision and correctness components occupying approximately orthogonal directions in activation space. In Gemma~4, steering along a verbal-confidence-specific activation direction causally shifted subsequent abstention behaviour. Together, these findings show that verbal confidence and log-probability confidence are not interchangeable uncertainty measures. Calibrated log-probabilities are more closely tied to answer evidence and correctness; verbal confidence is better understood as a behaviour-facing readout of an internal commit-readiness state. This challenges the common practice of treating verbal confidence reports primarily as estimates of correctness, with implications for users and downstream systems that rely on them as proxies for reliability.

\section*{Introduction}
An important desideratum for large language models is the production of confidence signals that convey their internal estimate of the correctness of their answer \citep{pouget2016confidence, kepecs2012computational}. Such signals are operationally important: they communicate how much weight a user should place on a model's output, and they may also support the model's own downstream decisions, such as deferring to a human, seeking additional information, or revising a prior commitment \citep{stone2022second, fleming2017self, kepecs2012computational, balsdon2020confidence, webb2023natural}. Multiple elicitation methods have been proposed, including log-probability-based confidence, sampling-based methods, and \emph{verbal confidence} --- the model's explicit response to a direct question about how confident it is in its answer \citep{steyvers2025metacognition, xiong2023can, lin2022teaching, kadavath2022language, tian2023just}. Verbal confidence is of particular interest because it is a black-box measure, applicable in the many settings where the model's internal probabilities are not accessible. This makes it the only confidence measure available for many widely deployed reasoning and non-reasoning models, and accordingly the subject of growing interest as a default reliability signal.

The value of such signals rests on a simple assumption: confidence should track the likelihood that an answer is correct, assigning higher scores to correct responses than to wrong ones. Here we challenge the assumption that verbal confidence is best characterised in this way, and ask whether it instead functions as a commit-readiness signal --- an index of the model's behavioural disposition to stand behind its answer. This distinction matters both conceptually and operationally. A confidence signal more tightly coupled to commit-readiness than to truth functions as a behavioural readout rather than as a faithful indicator of answer correctness, with direct implications for how such signals should be interpreted by users and downstream systems, and for the training objectives needed to better anchor confidence to correctness. 

Prior work has shown that confidence-related signals predict abstention in LLMs, and that steering along a calibrated log-probability-derived activation direction can causally alter abstention behaviour \citep{kumaran2026causal}. Those findings establish that confidence-related internal signals can guide abstention; here we ask what kind of signal verbal confidence is. First, do verbal confidence signals predict the model's downstream commit/abstain decision more strongly than they predict the actual correctness of its answer? Second, do verbal and log-probability-based confidence signals differ in this respect? Third, is any such behavioural alignment reflected in an internal state associated with verbal-confidence generation? 

\begin{figure}[ht]
    \centering
    \includegraphics[width=\textwidth]{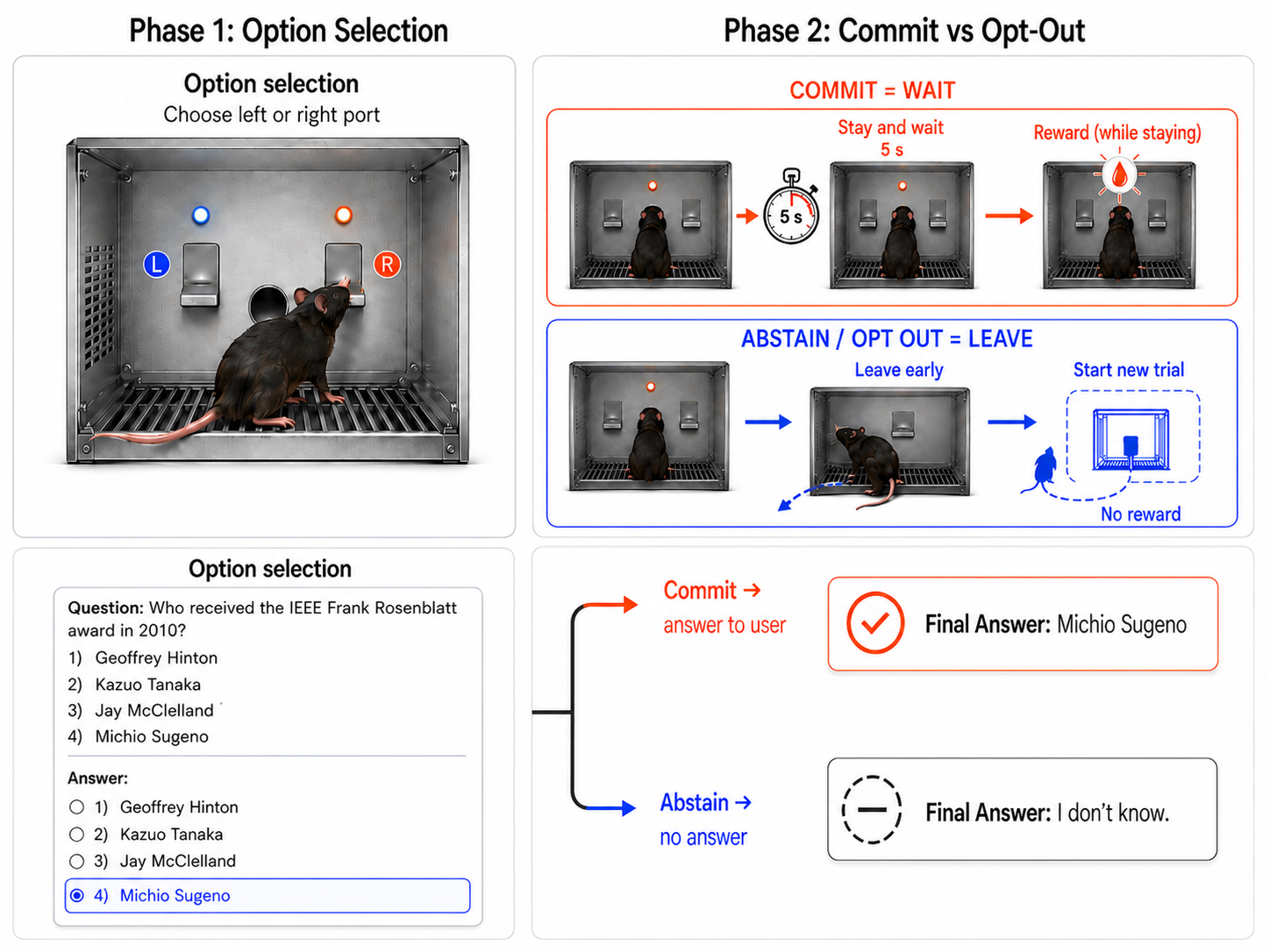}
  \caption{\textbf{Two-phase abstention paradigm inspired by post-decision confidence studies in rodents and primates.} \\
  \\
  \textbf{Schematic illustration of a rodent experiment \citep{kepecs2008neural, lak2014orbitofrontal} (top row)}: the animal first makes a right/left decision according to the dominant component of an odor mixture (Phase~1), and then (Phase~2) chooses whether to wait for a variable delayed reward (commit) or leave the port to begin a new trial (abstain). \\
  \textbf{Overview of LLM task (bottom row).} Phase~1 (left): The model selects from multiple-choice options (SimpleQA, MMLU-Pro, SuperGPQA) or generates a free-form response (HLE)(see Methods for exact prompts used). Three confidence signals are collected: log-probability-based confidence (Cal-LP), computed from the temperature-scaled softmax over the option letters, and verbal confidence (VC), elicited by directly asking the model to rate its confidence on a 10-class labelled scale (Class-VC), or a numerical scale of 0-100 (Num-VC). Phase~2 (right): The model is presented with its Phase~1 answer (and reasoning trace, in a reasoning model) and makes a binary commit-or-abstain decision. \emph{Commit} returns the chosen answer to a fictitious user; \emph{abstain} signals ``I don't know.'' Phase~1 confidence signals are tested as predictors of (a) the correctness of the Phase~1 answer and (b) the Phase~2 abstention decision on the same trials, allowing direct comparison of how strongly each signal tracks correctness versus the downstream commit decision. Notably, unlike in the animal paradigm, where the opt-out behaviour serves as the confidence readout itself, we measure Phase~1 confidence independently and treat abstention as the later behaviour to be predicted.}
    \label{fig:2-stage-paradigm}
\end{figure}

To address these questions, we use a two-stage abstention paradigm inspired by neuroscience studies of post-decision confidence in rodents and primates \citep{lak2014orbitofrontal, kiani2009representation, kepecs2008neural, beran2006rhesus} (Figure~\ref{fig:2-stage-paradigm}). In Phase~1, the model produces an answer --- either by selecting from multiple-choice options (SimpleQA \citep{kumaran2026llms, wei2024measuring}, MMLU-Pro \citep{wang2024mmlu}, SuperGPQA \citep{du2025supergpqa} datasets) or by generating a free-form response (HLE \citep{center2026benchmark} dataset) --- and in a separate elicitation step produces a verbal confidence rating in its own answer. Crucially, Phase~1 is run without any reference to the abstention task: the model is not informed that it will subsequently be asked whether to commit or abstain. In Phase~2, the model is shown its Phase~1 answer (and trace in the case of a reasoning model) and makes a binary commit-or-abstain decision: commit, in which case its answer is shown to a (fictitious) user, or abstain, in which case the response becomes ``I don't know.'' The Phase 1 confidence signals (verbal confidence and calibrated log-probability) are therefore generated without reference to the Phase 2 abstention task, and reflect the model's confidence in its answer in isolation. This two-phase structure separates the answer-production step from the meta-decision about whether to commit, allowing us to directly compare each Phase~1 confidence signal as a predictor of (a) Phase~1 correctness and (b) the Phase~2 abstention decision on the same trials. Unlike in animal paradigms, where opting out is itself the confidence readout, we measure Phase~1 confidence independently and ask whether it tracks later abstention or objective correctness more closely. 

We investigate these issues in four non-reasoning instruction-tuned models (Gemma~3 27B, Gemma~4 31B, GPT-4o, Qwen 3 235B no-think mode) on a SimpleQA-derived multiple-choice benchmark and MMLU-Pro, and in four reasoning models (Gemini Flash 3 preview, Kimi K2 Think, GPT-oss-120B and Qwen 3 235B think mode) across four benchmarks of varying difficulty (SimpleQA, MMLU-Pro, SuperGPQA, HLE). For Gemma~3 27B and Gemma~4 31B, where we have access to internal activations, we complement the behavioural analyses with mechanistic tests of the underlying confidence-related state. We ask whether Phase~1 activations measured immediately after answer production --- and before the abstention prompt --- already predict the later commit/abstain decision more strongly than they predict objective correctness. In Gemma~4, we further test causality: whether injecting a verbal-confidence-related direction derived from the Phase~1 state can shift the model's subsequent commit/abstain behaviour. The mechanistic claim is not that the Phase~1 verbal confidence report itself causes the later abstention decision, but that verbal confidence and later commitment behaviour reflect a shared internal state from which an extracted direction can causally bias the decision.

\section*{Results}
\subsection*{Confidence measures and baseline behaviour}
We first analysed the non-reasoning models, extracting three Phase~1 confidence signals to be used as predictors of Phase~2 abstention behaviour: class-based verbal confidence (Class-VC), numeric verbal confidence (Num-VC), and calibrated log-probability confidence (Cal-LP). Cal-LP was obtained by temperature-scaling the option-token distribution on a held-out Phase~0 calibration set \citep{guo2017calibration}(see Methods). In non-reasoning models the answer-token log-probability distribution provides a readout of the model's option-level distribution at the moment of answer selection in Phase~1 (cf reasoning models; see later). 

Following standard practice in prior work on verbal confidence \citep{yoon2025reasoning, tian2023just, xiong2023can}, we did not post-hoc calibrate class-VC or Num-VC. Because our primary analyses are based on AUROC --- a non-parametric discrimination measure invariant to monotonic transformations of the predictor --- post-hoc calibration would not change any of the reported results. Cal-LP is calibrated only to enable interpretable comparison of mean confidence on the same absolute scale across signals (Figure~10). Across SimpleQA and MMLU-Pro, Cal-LP discriminated correct from incorrect answers more strongly than either verbal measure (Cal-LP AUROC $0.62$--$0.80$; VC $0.53$--$0.65$; NumC $0.51$--$0.68$; Table~\ref{tab:vc_numc_callp_calibration}). The three signals were  partially correlated within cells, indicating shared confidence-related variance while leaving open whether they differ in their relationship to correctness and downstream abstention (Table~\ref{tab:three_way_correlation}).

The behavioural task elicited substantial variation in both accuracy and abstention. Phase~1 accuracy ranged from $0.539$ to $0.648$ on SimpleQA and from $0.292$ to $0.494$ on MMLU-Pro, while Phase~2 abstention rates ranged from $9\%$ to $72\%$ on SimpleQA and from $6\%$ to $83\%$ on MMLU-Pro (Table~\ref{tab:nrm_behavior}). Notably, in every model--dataset cell, abstention was more frequent after incorrect than correct answers (i.e. delta ranging from $+0.028$ to $+0.224$), indicating that the commit/abstain decision was partially discriminatory with regards to correctness. This sets up the central 
question of this paper: whether the verbal confidence reports models produce are more tightly coupled to truth or to this later commit/abstain behaviour.

\subsection*{Verbal confidence shows a robust decision--truth gap that Cal-LP does not}
A confidence signal that genuinely tracks correctness should predict abstention only insofar as it predicts correctness: low-confidence answers are abstained from because they are more likely to be wrong. A signal's predictive power for the commit/abstain decision should therefore closely track its predictive power for correctness --- a signal that only weakly separates right from wrong responses should not strongly separate commit from abstain decisions unless it carries additional behaviour-relevant information. A systematic excess of behavioural prediction over truth prediction therefore provides an initial signature of decision alignment, which we test more stringently below.

To quantify this, we define the \emph{decision--truth gap} as
\[
\Delta_{\mathrm{DT}} = \mathrm{AUROC}_{\mathrm{abstention}} - \mathrm{AUROC}_{\mathrm{correctness}},
\]
computed on the same trials with paired-bootstrap 95\% confidence intervals. Importantly, we used AUROC as a primary metric since it is non-parametric and independent of base rates, accuracy and ECE (i.e. confidence calibration); it represents the probability that a randomly chosen positive trial (correct or committed) receives higher confidence than a randomly chosen negative trial (incorrect or abstained).

\begin{figure}[!t]
    \centering
    \includegraphics[width=0.8\textwidth]{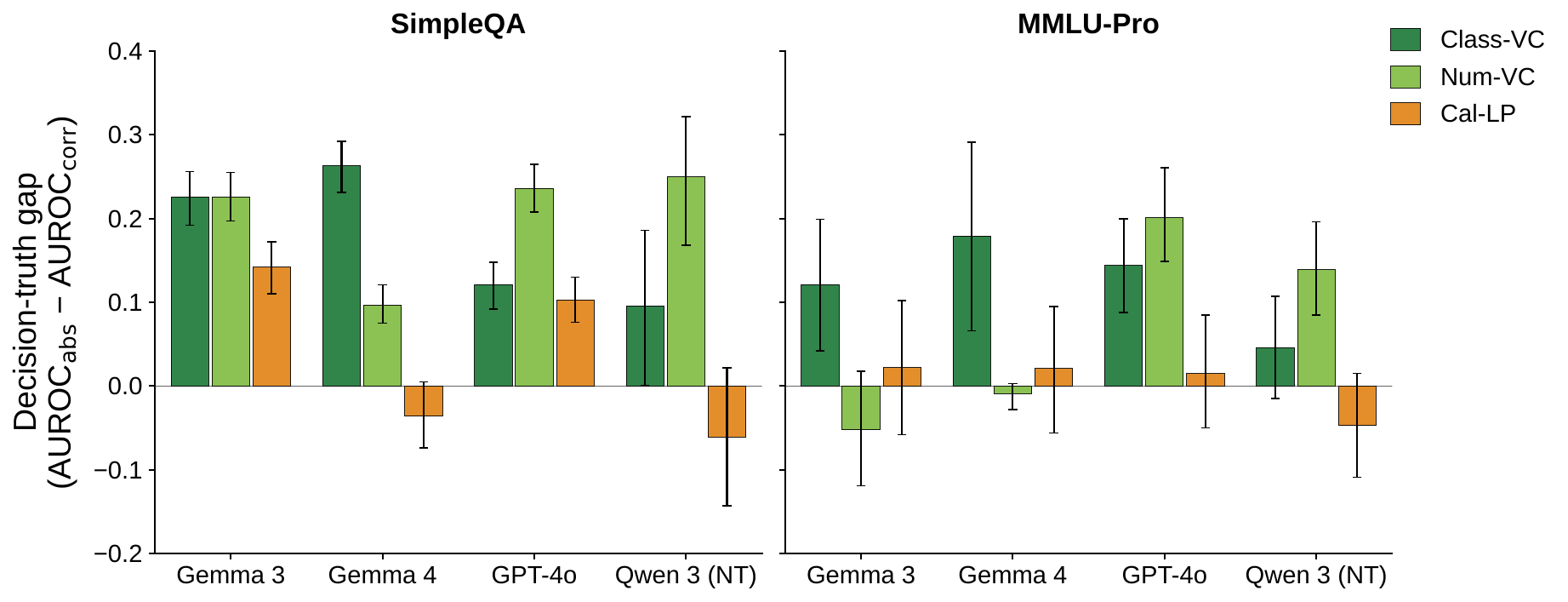}
    \caption{\textbf{Decision-truth gaps for each confidence signal across non-reasoning models.}
For each (model, dataset) cell, we plot the difference between the AUROC for predicting Phase~2 abstention and the AUROC for predicting Phase~1 correctness, computed on the same trials: $\text{AUROC}_{\text{abstention}} - \text{AUROC}_{\text{correctness}}$. Positive values indicate that the signal predicts abstention more strongly than correctness; near-zero or negative values indicate no systematic decision-over-truth advantage. Verbal measures (Class-VC, Num-VC) consistently show large positive gaps; calibrated log-probability (Cal-LP) shows small or negative gaps. Error bars are 95\% paired-bootstrap CIs from 1000 trial-level resamples within each (model, dataset) cell; per-cell trial counts are reported in Table~\ref{tab:nrm_behavior}. Gemma 4 and Qwen 3 (NT) are run in no-think mode. Note numeric confidence for Gemma~4 31B on MMLU-Pro is degenerate (>98\% scores are 100). See Table~\ref{tab:nrm_decision_truth_gap} for exact gap values and CIs.}
\label{fig:NRM_fig_decision_truth_gap}
\end{figure}

Both class-based and numeric verbal confidence showed positive decision--truth gaps in nearly every cell (Table~\ref{tab:nrm_decision_truth_gap}; Figure~\ref{fig:NRM_fig_decision_truth_gap}). Class-based VC gaps ranged from $+0.05$ to $+0.26$ across the eight model--dataset cells, and numeric VC gaps ranged from $-0.05$ to $+0.25$. By contrast, Cal-LP showed near-zero or negative gaps in 7 of 8 cells (range: $-0.06$ to $+0.14$). Thus, verbal confidence predicts abstention substantially better than correctness, whereas Cal-LP discriminates abstention and correctness comparably. 

For descriptive purposes, we treated gaps greater than $0.05$ AUROC as non-trivial and analysed whether this pattern held reliably across the 8 model--dataset cells. Class-VC gaps exceeded this threshold in 7/8 cells (median $+0.13$, one-sample Wilcoxon $W=35$, $p=0.008$). Num-VC gaps (excluding Gemma 4 31B on MMLU-Pro where numeric confidence was degenerate; see Figure~\ref{fig:all_NRMmodels_vcplot}; >98\% scores were 100) exceeded this threshold in 6/7 cells (median $+0.20$, $W=25$, $p=0.039$). Cal-LP gaps exceeded this threshold in only 2/8 cells (median $+0.018$, $W=10$, $p=0.88$), and did not reliably exceed zero ($W=21$, $p=0.37$). Pairwise comparisons confirmed the dissociation: verbal-confidence gaps reliably exceeded Cal-LP gaps across cells (Class-VC vs.\ Cal-LP: median $\Delta=+0.11$, $W=36$, $p=0.004$; Num-VC vs.\ Cal-LP: median $\Delta=+0.13$, $W=27$, $p=0.016$; one-sided Wilcoxon signed-rank).

Together, these results reveal a robust asymmetry in how the two confidence signals relate to truth and behaviour. Verbal confidence predicts the model's later commitment decision substantially more strongly than it predicts correctness (i.e. shows a positive decision-truth gap) --- whereas Cal-LP shows no comparable decision-over-truth advantage. This establishes the core behavioural dissociation: VC is disproportionately aligned with downstream commitment behaviour, while Cal-LP remains more comparably coupled to answer correctness and abstention. 

\subsection*{Verbal confidence and Cal-LP carry separable information about behaviour and truth}
The decision--truth gap shows that VC predicts abstention more strongly than correctness, while Cal-LP does not. We next asked how Class-VC, Num-VC, and Cal-LP divide predictive labour for the two outcomes of interest --- the model's later commit/abstain decision and the objective correctness of its answer --- when all available confidence readouts are considered jointly. If verbal confidence is preferentially decision-aligned, it should contribute unique variance to abstention prediction beyond Cal-LP; if Cal-LP is preferentially truth-aligned, it should contribute unique variance to correctness prediction beyond the verbal measures.

To test this, we fit nested logistic regressions with Class-VC, Num-VC, and Cal-LP as predictors of Phase~2 abstention and, separately, of Phase~1 correctness, and decomposed total McFadden $R^2$ into unique and shared components (Supplementary Table~\ref{tab:nrm_auroc_variance}; Methods). The two outcomes showed different variance structures. For correctness, Cal-LP dominated: it accounted for 42--87\% of total explained variance across cells, whereas the verbal measures contributed little unique variance (median $2\%$). For abstention, by contrast, a substantial fraction of explained variance was shared across confidence signals (35--56\% across cells, median 45\%), indicating that the later commit/abstain decision draws on structure common to both verbal confidence and Cal-LP. Nevertheless, verbal measures also contributed sizeable unique abstention-relevant variance, with the dominant verbal signal accounting for 16--64\% of total explained variance across cells (median 34\%) and varying by model: Class-VC contributed most strongly in Gemma~3 and Gemma~4, whereas Num-VC dominated in GPT-4o and Qwen. Cal-LP's unique contribution to abstention was generally smaller and more variable (0.5--24\% across cells; median 11\%) than its dominant unique contribution to correctness.

The two confidence families therefore partition by target. Verbal measures account for substantial unique variance in the model's later commit/abstain decision, whereas Cal-LP accounts for the dominant unique component of correctness prediction. The substantial shared variance for abstention is consistent with both signals partially tracking truth: because the model abstains more often on incorrect trials, any signal that tracks correctness will also predict abstention, and the shared component captures this common truth-derived contribution. What distinguishes verbal confidence is what lies \emph{beyond} this shared component: VC carries additional unique abstention-relevant variance that is not coupled to truth-tracking, whereas Cal-LP's behavioural relevance remains largely tied to its correctness signal. 

These decompositions establish \emph{which} outcome each signal preferentially predicts behaviourally. We next ask whether this behavioural-prediction asymmetry reflects an underlying structural asymmetry in how each signal's variance is \emph{internally} organised, using a complementary metric---partial $\eta^2$ from a 2$\times$2 ANOVA---that captures signal-variance organisation by factor rather than behavioural prediction. We also note that Cal-LP's dominance of VC's correctness-related variance (verbal measures contribute a median of $\approx 2\%$ unique correctness $R^2$) positions it as the empirical proxy for answer-evidence in residualisation analyses reported later.

\subsection*{Verbal confidence is organised by downstream commitment, whereas Cal-LP retains stronger correctness structure}
Having shown that VC and Cal-LP partition by target in their behavioural-prediction labour, we next ask whether this asymmetry reflects how each signal's variance is internally organised. If VC is primarily a truth-estimate, its trial-to-trial variance should be explained mainly by whether the answer is correct. If instead VC is a behavioural confidence readout, its variance should be explained more strongly by whether the model later commits to or abstains from the answer.

\begin{figure}[!t]
    \centering
    \includegraphics[width=1\textwidth]{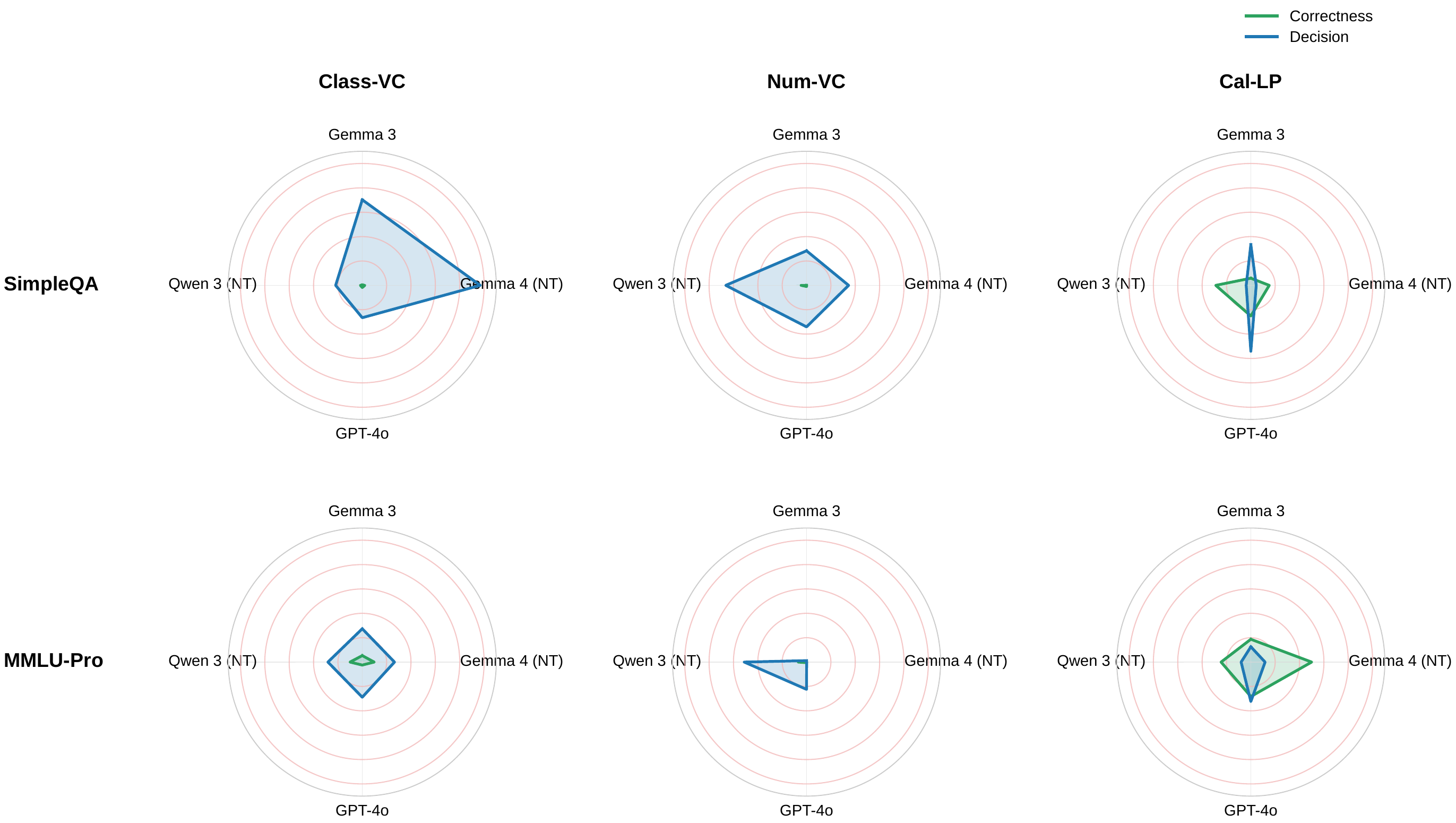}
    \caption{
    \textbf{Verbal confidence is dominated by the abstention decision, while Cal-LP also carries correctness-related variance.}
    Radar plots show partial $\eta^{2}$ (i.e. the proportion of variance in the signal explained by a factor, after accounting for the other factors in the model) associated with objective correctness (green) and the model's later commit/abstain decision (blue), from a two-way ANOVA of each confidence signal on correctness, decision, and their interaction. Each radar shows four spokes (one per model) for one signal $\times$ dataset combination. Concentric gridlines mark increments of $0.1$ in partial $\eta^{2}$, with the outer ring at $0.5$. For both verbal confidence measures (Class-VC, Num-VC), the decision effect dominates and the correctness effect is small in nearly every cell, indicating that verbal confidence is organised primarily by the later commit/abstain decision. Cal-LP shows a different pattern: it carries substantially more correctness-related variance than VC, with the relative balance of decision vs.\ correctness effects varying across models and datasets. This is consistent with verbal confidence being a decision-aligned signal and Cal-LP being a more truth-aligned signal that also carries decision-relevant information. See Table~\ref{tab:eta2_effect_sizes} for exact values. Numeric confidence for Gemma~4 31B on MMLU-Pro is degenerate ($>98\%$ scores are 100). Gemma~4 31B and Qwen~3 235B are run in no-think mode (NT).
}
    \label{fig:nrm_radar}
\end{figure}

For each confidence signal, we fit a two-way ANOVA with correctness, decision, and their interaction as predictors, and plotted partial $\eta^{2}$ for the correctness and decision main effects (Figure~\ref{fig:nrm_radar}; full values are reported in Table~\ref{tab:eta2_effect_sizes}; see Figure~\ref{fig:NRM_raw_means_grid} for raw mean-confidence plots). Partial $\eta^{2}$ quantifies the proportion of variance in the signal explained by a factor after accounting for the other terms in the model. For VC, the decision-related effect size substantially exceeded the correctness-related effect size in nearly every cell where the signal was available (Class-VC: decision $\eta^{2}$ 0.11--0.48 vs.\ correctness $\eta^{2}$ 0.00--0.05 across all 8 cells; Num-VC: decision $\eta^{2}$ 0.04--0.43 vs.\ correctness $\eta^{2}$ 0.00--0.06 across the seven non-degenerate cells, with Gemma~4 MMLU-Pro excluded). Trials that the model would later commit to therefore tended to receive higher verbal confidence than trials it would later abstain on, even though VC only weakly separated correct from incorrect answers. This is the variance-structure counterpart to the behavioural dissociation: VC is dominated, in its overall variance, by decision-aligned structure. 

Cal-LP showed a different profile: it carried substantially more correctness-related variance than VC, and the balance of decision versus correctness effects varied by model. In some cells Cal-LP was dominated by the correctness effect (e.g., SimpleQA Qwen~3: correctness $\eta^{2}=0.14$ vs.\ decision $\eta^{2}=0.02$), whereas in others it was dominated by the decision effect (e.g., SimpleQA GPT-4o: correctness $\eta^{2}=0.13$ vs.\ decision $\eta^{2}=0.27$).

The $\eta^{2}$ and the regression decomposition use different metrics---$\eta^2$ captures signal-variance organisation by factor, whereas regression $R^2$ captures behavioural prediction. Their convergence on the same dissociation establishes that VC and Cal-LP are differentially organised around the decision and correctness outcomes respectively. Whether VC's decision-aligned profile reflects a genuine non-truth component, however, cannot be settled by variance organisation alone — nor by the observed decision–truth gap alone. Both observations have the same potential alternative explanation: if the commit/abstain decision is a thresholded readout of VC, then VC's variance will appear decision-organised and VC will predict abstention more strongly than binary correctness, even when VC carries only truth content. The decision–truth gap and the decision-organised variance structure are therefore both consistent with two qualitatively different possibilities: that VC contains a genuine non-truth component aligned with the commit/abstain decision, or that VC carries only truth content driven through a thresholded decision rule.

\subsection*{Residualisation analysis: VC predicts abstention independently of correctness evidence}
We consider two generative accounts that formalise these two possibilities (see Figure 9 for details). Under Account 1, VC is a noisy readout of a latent truth signal $z_T$, and the model abstains whenever VC falls below an internal threshold. Under Account 2, VC is the sum of $z_T$ and an independent decision (i.e. commitment/abstention) component $z_D$ uncorrelated with correctness, and the model again abstains by thresholding VC. Both accounts produce a positive decision--truth gap, because a graded signal that drives a thresholded behaviour predicts that behaviour better than it predicts binary correctness: thresholding the graded signal recovers the abstention decision directly, whereas correct and incorrect answers can produce overlapping graded values that the binary correctness label cannot distinguish. Of note, we formalise the decision rule as a hard threshold because our motivating observation is VC's positive decision--truth gap, and a hard threshold captures the strongest version of the alternative interpretation we need to rule out---that VC's gap arises mechanically from thresholded use of a graded truth signal. This is not to say that every graded truth signal produces a gap: gap magnitude depends on decision-rule steepness, with softer rules attenuating the gap toward zero. The residualisation analyses, however, are invariant to decision-rule steepness. In summary, the raw gap is consistent with either Account 1 or Account 2, and the residual analyses provide the diagnostic test.

To distinguish them, we use calibrated log-probability confidence (Cal-LP) as the empirical proxy for $z_T$: as established above, Cal-LP is the strongest single-trial correctness-tracking signal available in our non-reasoning models (correctness AUROC $0.62$--$0.80$ vs.\ $0.51$--$0.65$ for VC) and captures essentially all of VC's correctness-relevant variance. The residualisation does not require Cal-LP to equal $z_T$; it requires only that Cal-LP captures the bulk of correctness-related variance shared with VC. We residualise VC on Cal-LP within each cell and ask whether residual VC predicts Phase 2 abstention, Phase 1 correctness, or neither. The two accounts make distinct predictions (see Figure~\ref{fig:residualization_schematic}). Under Account~1, the residual contains only measurement noise and should predict neither outcome. Under Account~2, the residual should be dominated by the decision component $z_D$ and should therefore predict the commit/abstain decision while showing no above-chance discrimination of correctness. The sign of the residual--correctness association is the critical test: a positive association would be consistent with VC tracking additional truth-evidence beyond Cal-LP, whereas a null or negative association rules this out.

\begin{figure}[!t]
    \centering
    \includegraphics[angle = -90, width=1\textwidth]{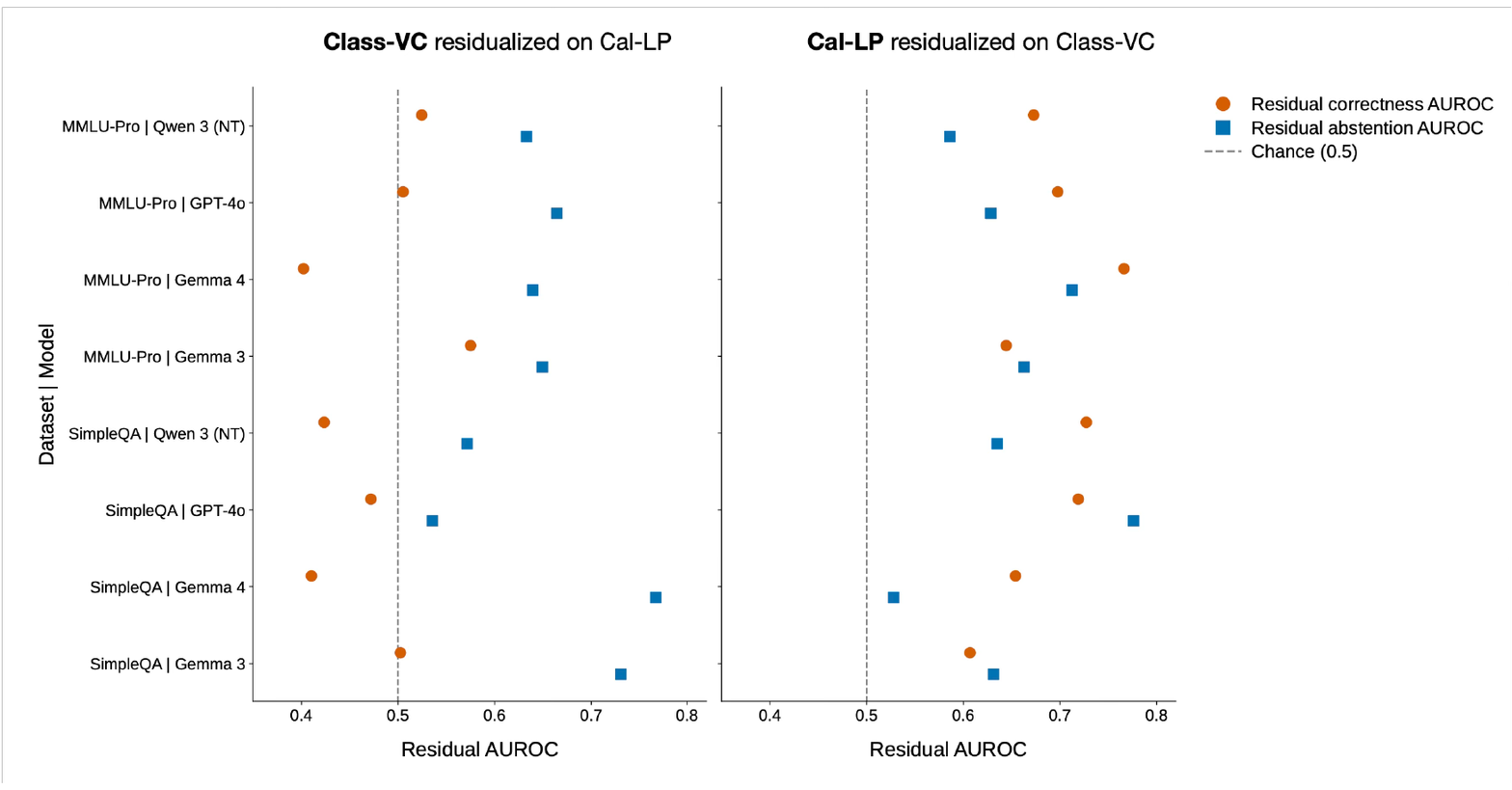}
    
\caption{\textbf{Reciprocal residual analyses distinguish the signal components unique to verbal confidence and Cal-LP.}
We remove the linear component shared between Class-VC and Cal-LP in each direction, then ask whether the remaining signal is more aligned with objective correctness or with the model's later commit/abstain decision.
\textbf{(a)} Residual Class-VC AUROCs after removing the component shared with Cal-LP. Across non-reasoning model cells, the VC-specific residual generally remains more predictive of the later commit/abstain decision than of objective answer correctness (which is at chance or below chance levels - see main text for details). 
\textbf{(b)} Residual Cal-LP AUROCs after removing the linear component shared with Class-VC. The Cal-LP-specific residual retains strong correctness discrimination across cells while also continuing to predict later abstention, consistent with a residual answer-evidence signal that remains behaviourally relevant. Circles show residual correctness AUROC; squares show residual abstention AUROC; the dashed vertical line marks chance performance ($0.5$). Point estimates are shown for clarity; bootstrap 95\% confidence intervals and residual decision--truth gaps are reported in Supplementary Table~\ref{tab:reciprocal_residual_analysis}. Gemma~4 31B and Qwen~3 235B are run in no-think mode (NT).}
\label{fig:residual_auroc}
\end{figure}

For Class-VC, the residual signal remained predictive of abstention in all eight non-reasoning model cells (median AUROC $=0.64$, range $0.54$--$0.77$), while its correctness discrimination was weak and directionally inconsistent (median AUROC $=0.49$, range $0.40$--$0.58$; Supplementary Table~\ref{tab:reciprocal_residual_analysis}; Figure~\ref{fig:residual_auroc}). The residual decision--truth gap was positive in 8/8 cells (median $+0.15$). Notably, residual correctness AUROC fell below $0.5$ in several cells, indicating that lower residual VC was associated with \emph{greater} objective correctness even though it predicted \emph{greater} abstention. This active reversal---not merely the absence of correctness alignment---rules out the strongest Account~1 interpretation available under this proxy: if residual VC were simply additional truth evidence not captured by Cal-LP, lower residual VC should predict both greater abstention and lower correctness, not opposite directions. Num-VC showed a broadly similar pattern, with positive residual decision--truth gaps in 6/8 cells; the two exceptions were MMLU-Pro cells in which raw Num-VC did not predict abstention above chance, including the degenerate Gemma~4 cell in which $>98\%$ of responses were 100 (Supplementary Table~\ref{tab:reciprocal_residual_analysis}).

The variance in verbal confidence not explained by Cal-LP is therefore not consistently aligned with truth; it is preferentially aligned with the model's later commit/abstain decision. This pattern matches Account~2: residual VC behaves as a decision-aligned component that exists in VC independently of any truth-tracking. The raw VC signal is therefore not reducible to the Account~1 picture of a noisy correctness estimate.

We then performed the converse analysis, residualising Cal-LP on verbal confidence. After removing variance shared with Class-VC, residual Cal-LP retained strong correctness discrimination (median AUROC $=0.69$) and abstention discrimination (median AUROC $=0.63$), with a residual decision--truth gap near zero (median $-0.06$). The same pattern held when Cal-LP was residualised on Num-VC (Supplementary Table~\ref{tab:reciprocal_residual_analysis}). Cal-LP's unique component therefore remained a coupled truth-and-behaviour signal---predicting both outcomes in proportion---rather than dissociating into truth-aligned and behaviour-aligned parts. This profile is consistent with Cal-LP behaving as a truth-evidence signal whose behavioural relevance is inherited from its truth-tracking, paralleling the variance-organisation finding above that Cal-LP's variance is correctness-organised. The two residuals therefore differ in kind: the VC-specific residual is preferentially commitment-aligned in a way that matches Account~2, whereas the Cal-LP-specific residual remains coupled across truth and behaviour, the profile expected if Cal-LP is primarily an answer-evidence signal whose behavioural relevance is coupled to its correctness-tracking.

Cell-level tests confirmed these patterns (see Supplemental Results for further details). Across the eight non-reasoning cells, residual Class-VC showed no reliable positive correctness alignment (mean AUROC $=0.473$; one-sided exact Wilcoxon signed-rank against $0.5$: $W=11$, $p=0.844$), whereas residual Cal-LP remained strongly correctness-aligned (mean AUROC $=0.686$; $W=36$, $p=0.0039$). Residual Cal-LP's correctness AUROC exceeded residual Class-VC's in every cell (mean difference $=+0.213$; $W=36$, $p=0.0039$; one-sided paired Wilcoxon).

Further, complementary residualisations of VC against binary correctness, Cal-LP, and item difficulty yielded the same conclusion (see Supplemental Results; Table~\ref{tab:residual_correctness_proxy_analysis}). Together, therefore, these analyses show that VC's alignment with later commitment is not reducible to the answer-evidence signal captured by Cal-LP, and remains detectable even after additionally controlling for observed correctness and question-level difficulty.

\paragraph*{Robustness across abstention prompts.}
We replicated the analysis under a neutral abstention prompt that omits the cautionary clause of the standard prompt (See Supplemental Methods for exact prompt used). Whilst overall, abstention rates were lower with the neutral prompt (Table~\ref{tab:nrm_abstention_by_prompt}), the difference in conditional abstention rates (i.e. for incorrect vs correct rates) was positive across cells. Importantly, the dissociation pattern was preserved: Class-VC and Num-VC decision--truth gaps remained positive in nearly every cell, while Cal-LP gaps remained small or negative (Table~\ref{tab:nrm_dt_gap_comparison}). The cross-cell pattern held: verbal decision-truth gaps exceeded Cal-LP gaps in 8/8 (Class-VC) and 6/7 (Num-VC) cells under the neutral prompt (Wilcoxon $p \leq 0.008$ and $p \leq 0.016$, one-sided, respectively; see Supplemental Results for details). The dissociation pattern is therefore preserved under the neutral prompt: verbal confidence remains preferentially decision-aligned, whereas Cal-LP shows no comparably robust decision-over-truth advantage.

\subsection*{The verbal-confidence dissociation generalises to reasoning models}
We next asked whether the decision--truth dissociation extends beyond non-reasoning models to systems that generate explicit reasoning traces before committing to an answer. We evaluated four reasoning models --- Gemini Flash~3, Qwen~235B Think, Kimi K2 Think, and GPT-Oss-120B --- across four benchmarks: SimpleQA, MMLU-Pro, SuperGPQA-hard, and HLE. Phase~1 allowed chain-of-thought reasoning; the resulting trace was included when eliciting verbal confidence and when prompting the Phase~2 commit/abstain decision. We elicited class-based verbal confidence only. Cal-LP was available for seven reasoning-model cells: Gemini Flash~3 on all four datasets and Qwen~235B Think on the three MCQ benchmarks (see Methods for details). Unlike in the non-reasoning models, Cal-LP did not uniformly outperform VC in correctness discrimination: VC's correctness AUROC exceeded Cal-LP's in 4 of the 7 reasoning-model cells. This pattern is consistent with single-token Cal-LP capturing only a narrow slice of a reasoning model's uncertainty \citep{fu2026multiple, farquhar2024detecting}: in reasoning models, uncertainty is distributed across the reasoning trajectory rather than concentrated at the answer token, and no consensus LP-based alternative has yet emerged \citep{kang2026scalable, fu2025deep}. Verbal confidence has therefore become a particularly valuable practical signal in reasoning models \citep{yoon2025reasoning} --- making its properties especially important to characterise. We therefore treat Cal-LP in reasoning models as a partial reference signal rather than a fully comparable uncertainty measure, and report it alongside VC for continuity with the non-reasoning model analysis.

The behavioural setting was heterogeneous, spanning Phase~1 accuracies from $0.07$ to $0.94$ and abstention rates from $0.04$ to $0.46$ across the 16 model--dataset cells (Table~\ref{tab:rm_behavior}). Nevertheless, abstention remained higher on incorrect than correct trials in every cell, with $\Delta=P(\mathrm{abstain}\mid\mathrm{incorrect})-P(\mathrm{abstain}\mid\mathrm{correct})$ ranging from $+0.047$ to $+0.259$ (see Supplemental Results for details). 

\begin{figure}[!t]
    \centering
    \includegraphics[width=0.8\textwidth]{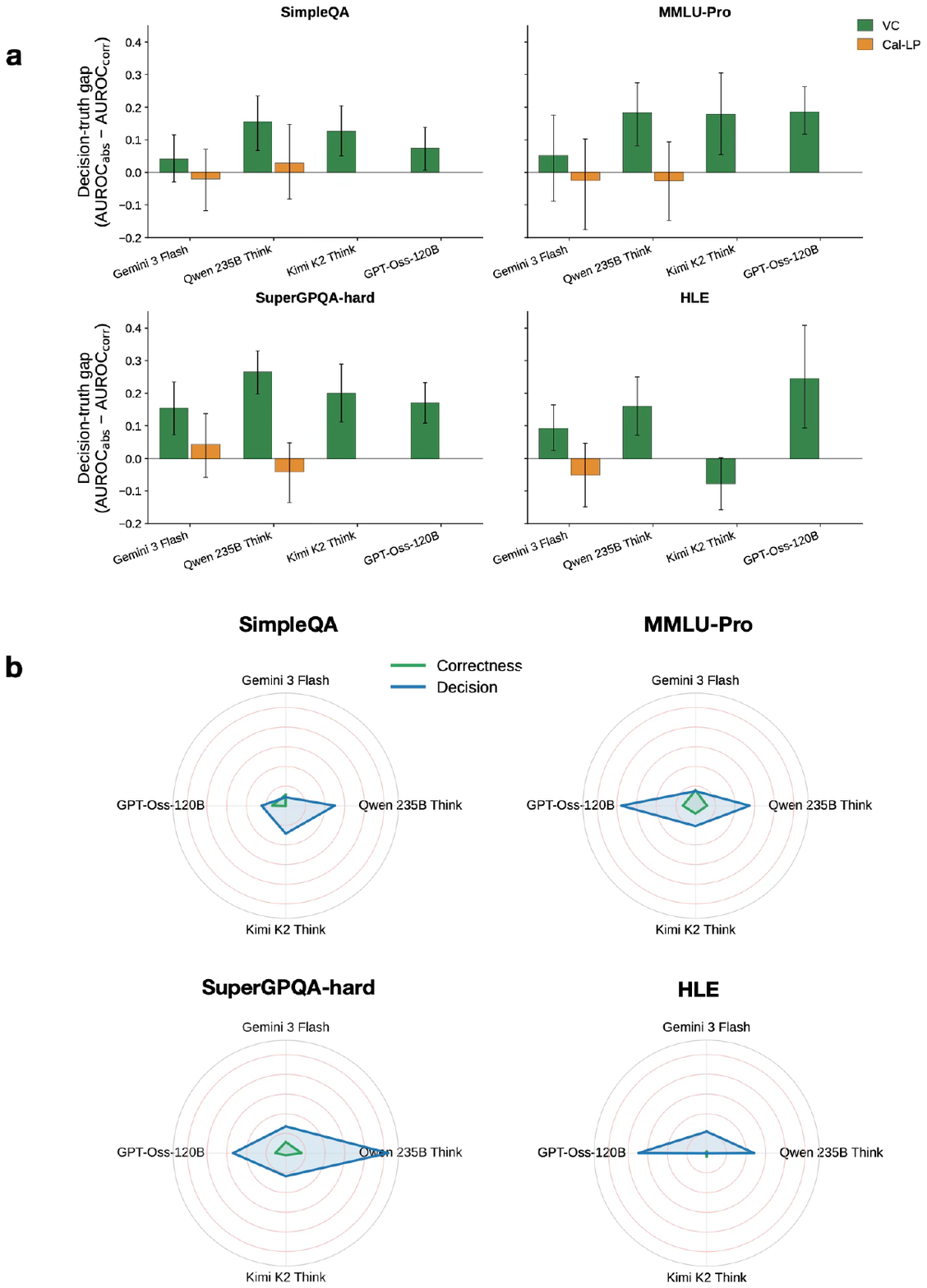}
\caption{\textbf{The decision--truth dissociation generalises to reasoning models.}
\textbf{(a)} Decision--truth gap for verbal confidence (Class-VC) and Cal-LP across reasoning-model cells. For each (model, dataset) cell, we plot $\mathrm{AUROC}_{\mathrm{abstention}}-\mathrm{AUROC}_{\mathrm{correctness}}$, computed on the same trials. Positive values indicate that the signal predicts abstention more strongly than correctness; near-zero or negative values indicate no systematic decision-over-truth advantage. Empty positions in the Cal-LP series indicate cells where Cal-LP was unavailable. Error bars are 95\% paired-bootstrap CIs from 1000 trial-level resamples within each (model, dataset) cell; per-cell trial counts ($N$) are reported in Table~\ref{tab:rm_auroc_variance}. Note that Cal-LP in reasoning models is computed from per-option log-probabilities at a checkpoint within or following the reasoning trace (see Methods), and is shown here as a partial reference signal rather than as a fully equivalent answer-evidence measure (see Main Text for details). 
\textbf{(b)} Variance decomposition of verbal confidence (Class-VC) across reasoning-model cells. Radar plots show partial $\eta^2$ from a two-way ANOVA of VC on correctness, decision, and their interaction. Each radar corresponds to one dataset; spokes denote models. Green polygons show variance explained by correctness; blue polygons show variance explained by the later commit/abstain decision. For VC, decision-related variance exceeds correctness-related variance in 13/16 cells. Cal-LP equivalents are reported in Supplementary Table~\ref{tab:rm_eta2_effect_sizes}.}
\label{fig:RM_fig_decision_truth_gap}
\end{figure}

The key dissociation replicated strongly (see Supplemental Results for further details of reasoning model analyses). Across the 16 reasoning-model cells, verbal confidence (Class-VC) showed a positive decision--truth gap in 15/16 cells (median $+0.16$, one-sample Wilcoxon $p = 1.1 \times 10^{-4}$), and exceeded the pre-specified $+0.05$ threshold in 14/16 cells ($p = 1.1 \times 10^{-3}$; Figure~\ref{fig:RM_fig_decision_truth_gap}a). Cal-LP, where available, showed no comparable pattern: its gap was centred near zero (median $-0.03$; 2/7 positive; $p = 0.77$). In the seven paired cells, the Class-VC gap exceeded the Cal-LP gap in 7/7 cases (median difference $+0.13$, $p = 0.008$), and exceeded it by at least 0.05 in all seven cells ($p=0.008$)(see Supplemental Results for details). The behavioural decision--truth dissociation therefore generalises robustly to reasoning models, including harder expert-level MCQ tasks and freeform answering.

The internal organisation of Class-VC showed the same profile. In a two-way ANOVA of verbal confidence on correctness and the later commit/abstain decision, the decision-related partial $\eta^2$ exceeded the correctness-related partial $\eta^2$ in 14/16 reasoning-model cells (decision $\eta^2$ median $0.25$; correctness $\eta^2$ median $0.06$; Figure~\ref{fig:RM_fig_decision_truth_gap}b)(see Supplemental Results for details). Thus, reasoning-model VC is organised predominantly by downstream commitment, not by whether the answer is objectively correct. By contrast, Cal-LP in reasoning models carried little organised variance for either factor where it was computable (median partial $\eta^2=0.006$ for correctness and $0.005$ for decision; Supplementary Table~\ref{tab:rm_eta2_effect_sizes}). This differs from the cleaner Cal-LP profile in non-reasoning models and likely reflects a measurement limitation: in reasoning systems, uncertainty is distributed across the generated reasoning trajectory rather than concentrated in a single answer-token distribution (e.g. \citep{kang2026scalable, fu2025deep, farquhar2024detecting}), making the checkpoint log-probabilities used (see Methods) a less complete proxy for answer evidence.

Additional analyses confirmed that the reasoning-model VC effect was not fully explained by correctness-related evidence. After residualising VC on binary correctness and Cal-LP, residual VC retained a median of approximately $87\%$ of its raw abstention-predictive power across the seven Cal-LP-available cells (range: 36--98\%; Supplementary Table~\ref{tab:rm_residual_abstention})(see Supplemental Results for details). Joint logistic regressions likewise showed that VC carried the dominant unique contribution to abstention prediction in these cells, accounting for a median of $96\%$ of total McFadden $R^2$, whereas Cal-LP's unique contribution was negligible in most cases (Table~\ref{tab:rm_auroc_variance}). Together, these results show that the core finding is not restricted to non-reasoning models: verbal confidence in reasoning models remains strongly and preferentially aligned with later commitment behaviour.

\paragraph*{Across experiments, VC shows a robust decision--truth gap absent for Cal-LP}
Across all 46 VC cells (8 models, 4 datasets, 2 prompts, 2 formats), the average decision--truth gap was $+0.16$ (significantly greater than zero; $t = 11.6$, $p < 0.001$, one-sample $t$-test). The magnitude of the gap did not differ between non-reasoning and reasoning models (Mann--Whitney $U=60.0$, $p=0.83$; NRM median $+0.13$, $n=8$; RM median $+0.16$, $n=16$; matched Class-VC, standard-prompt cells). There was no significant effect of prompt or format in a mixed-effects regression with model as random intercept; for Cal-LP (23 cells), the average gap was indistinguishable from zero ($p = 0.36$). Restricting to the 16 non-reasoning-model Cal-LP cells, the decision--truth gap remained indistinguishable from zero (mean $+0.015$; mixed-effects intercept 95\% CI $[-0.04, +0.07]$, $p = 0.60$; one-sample $t(15) = 0.86$, $p = 0.40$; Wilcoxon $W = 54$, $p = 0.50$), confirming that the Cal-LP profile is not an artefact of the partial single-token uncertainty signal in reasoning models.

\begin{figure}[t]
\centering
\includegraphics[angle = -90, width=\linewidth]{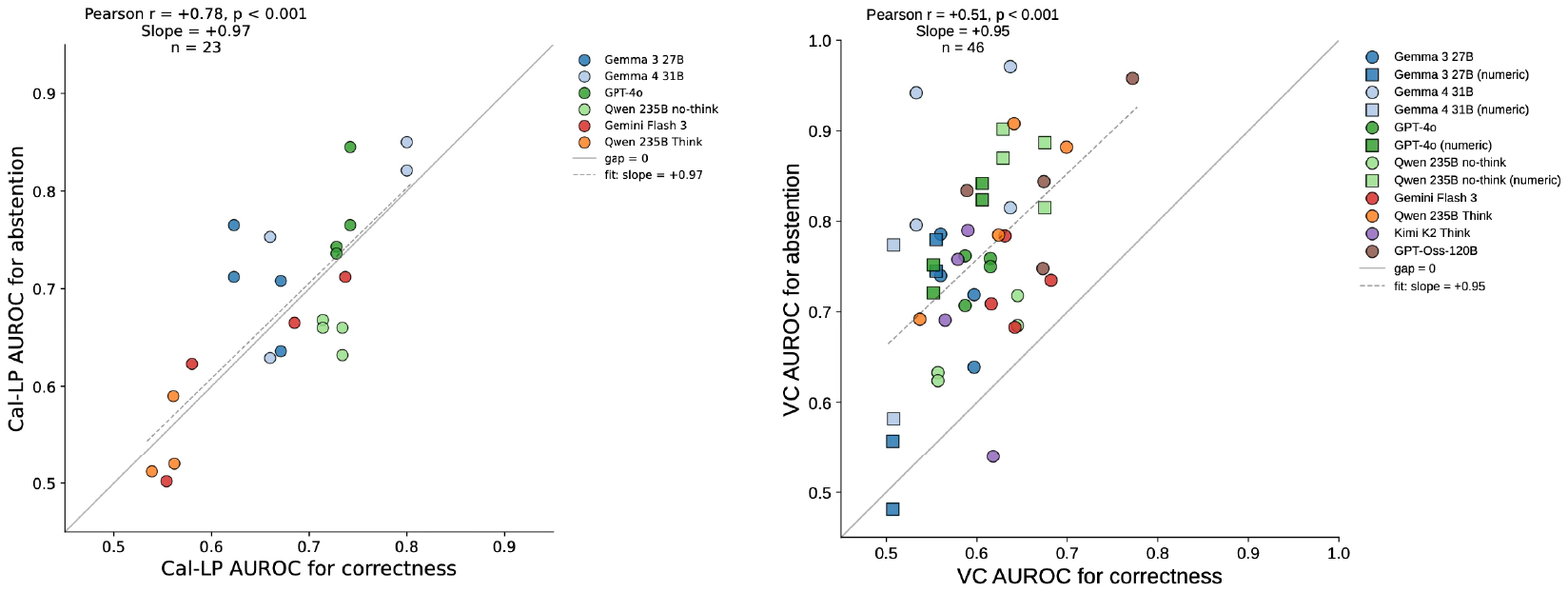}
\caption{\textbf{Cal-LP and VC differ in how their abstention prediction scales with their truth-tracking.} Each point represents one (model $\times$ dataset $\times$ prompt $\times$ confidence-format) cell. \textbf{(Left) Cal-LP} ($n = 23$ cells across 6 models with available log-probabilities). Cal-LP abstention AUROC scales tightly with Cal-LP correctness AUROC (Pearson $r = +0.78$; mixed-effects slope $\beta = +0.91$, 95\% CI $[+0.43, +1.39]$, $p < 0.001$, with model as random intercept), and the fitted line (dashed) lies close to the $y = x$ reference (solid grey). \textbf{(Right) VC} ($n = 46$ cells across 8 models, including labelled and numeric formats and both standard and neutral prompts). VC abstention AUROC also scales with VC correctness AUROC (Pearson $r = +0.51$; mixed-effects $\beta = +1.31$, 95\% CI $[+0.72, +1.89]$, $p < 0.001$), but the fitted line sits systematically above the $y = x$ reference: across the full range of VC correctness AUROCs we observe (0.51--0.77), abstention AUROC exceeds correctness AUROC by a substantial margin (mean gap $= +0.16$). The cross-cell pattern is therefore qualitatively different for the two signals. Cell-level Pearson correlations are descriptive; statistical inference accounts for non-independence of cells within model via the mixed-effects regression (see Supplementary Table~\ref{tab:mixed-effects} for full model specification and covariates).}
\label{fig:scatter-dissociation}
\end{figure}

The two signals also differed in how their abstention AUROC scaled with their correctness AUROC across cells (Figure~\ref{fig:scatter-dissociation}). For Cal-LP, abstention AUROC scaled tightly with correctness AUROC along a slope close to one (mixed-effects $\beta = +0.91$, 95\% CI $[+0.43, +1.39]$, $p < 0.001$; cell-level Pearson $r = +0.78$), with fitted values lying close to the $y = x$ reference. This pattern held when restricted to non-reasoning models (mixed-effects $\beta = +0.86$, 95\% CI $[+0.09, +1.63]$, $p = 0.029$; mean gap $+0.015$). For VC, abstention AUROC also rose with correctness AUROC ($\beta = +1.31$, 95\% CI $[+0.72, +1.89]$, $p < 0.001$; Pearson $r = +0.51$), but with the fitted line sitting systematically above the $y = x$ reference, indicating a systematic decision-over-truth advantage not explained by truth-tracking quality alone. The VC decision--truth gap was not significantly modulated by VC's correctness AUROC ($\beta = +0.31$, $p = 0.30$), indicating that the gap is approximately constant across the observed range of VC truth-tracking quality.

The cross-cell pattern is therefore qualitatively different for the two signals. This pattern is consistent with Cal-LP behaving primarily as an answer-evidence signal whose decision relevance scales with its truth-tracking quality, whereas VC behaves as a behaviour-anchored signal whose decision relevance is partly independent of truth-tracking. Full mixed-effects results, including dataset, prompt, and format covariates, are reported in Table~\ref{tab:mixed-effects}.

\section*{Activation-based Analyses: a post-answer confidence state predicts and causally controls abstention in Gemma~4}
The behavioural analyses establish that verbal confidence predicts the model's later commit/abstain decision more strongly than it predicts the correctness of its answer. We next asked whether this asymmetry is reflected in the internal representation from which verbal confidence is generated. We analysed Gemma~4 31B Phase~1 activations at the position immediately following the model's committed answer, which we term the post-answer (PA) position, in the SimpleQA task, and report convergent representational analyses in Gemma~3 27B in the Supplemental Results. This position is especially relevant for the present question: prior work showed that it is causally involved in generating verbal-confidence reports \citep{kumaran2026llms}, and our behavioural results suggest that such confidence reports are more tightly coupled to later commitment than to truth. If the same internal state underlies this dissociation, PA activations should encode the model's future commit/abstain decision more strongly than the correctness of its answer. 

We therefore extracted PA-position residual-stream activations before the model was presented with the Phase~2 abstention prompt (see Methods), and at each layer trained two linear probes: a \emph{correctness probe} predicting whether the Phase~1 answer was correct, and a \emph{decision probe} predicting the subsequent Phase~2 commit/abstain outcome (5-fold cross-validated logistic regression; see Methods).

\subsection*{Decoding correctness and decision across layers}
The two probes diverge substantially with depth (Figure~\ref{fig:activation}a). The correctness probe rises modestly to AUROC $\approx 0.66$ by L30 and plateaus, closely matching the joint behavioural baseline obtained by combining all three confidence signals (Class-VC + Num-VC + Cal-LP; dashed line). The decision probe rises steeply through mid-network layers, reaching AUROC $\approx 0.94$ at L38 --- substantially exceeding its corresponding behavioural baseline by $\approx 0.12$ AUROC, indicating that PA activations contain commit/abstain-relevant signal that the model's own verbalisable and Cal-LP confidence channels do not fully express. The post-answer state therefore encodes the model's later commit/abstain decision far more strongly than the correctness of the answer that has just been generated.

A complementary 2$\times$2 ANOVA on each probe's output (factors: correctness $\times$ decision) confirms the dissociation in variance terms (Figure~\ref{fig:activation}b). The decision probe's partial $\eta^2$ for the decision factor climbs steeply through mid layers, peaking at $\eta^2 = 0.49$ at L38, while the correctness probe's partial $\eta^2$ for correctness plateaus at a much lower level: $\eta^2 \approx 0.06$. The two signals are therefore not equally represented at the PA position: decision-related variance dominates correctness-related variance by approximately an order of magnitude at peak layers. The same probe-level dissociation was observed in Gemma~3 27B: decision decoding again far exceeded correctness decoding across the relevant post-answer layers, and at the peak decision layer (L32), decision-related variance in the decision probe ($\eta^2=0.584$) exceeded correctness-related variance in the correctness probe ($\eta^2=0.034$) by approximately $17$-fold (Supplemental Results; Figure~\ref{fig:activation_G3}).

As a control, we performed the same analysis at the question-last-token (QLT) position, which precedes answer generation. At QLT, the model has seen the question but has not yet committed to an answer, so neither truth (defined relative to that answer) nor commit-readiness should be strongly encoded. As expected, the correctness probe at QLT was almost at chance (peak AUROC $= 0.53$; partial $\eta^2 < 0.01$), and the decision probe was substantially weaker than at PA (peak AUROC $= 0.68$ vs.\ $0.94$; partial $\eta^2 = 0.07$ vs.\ $0.49$). The PA decision signal is therefore not simply a readout of question-level features already available before answer commitment; it emerges largely after the model has generated and committed to its answer.

\subsection*{Geometric and structural relationships between the two representations}
\begin{figure}[!t]
    \centering
    \includegraphics[angle=-90, width=\textwidth]{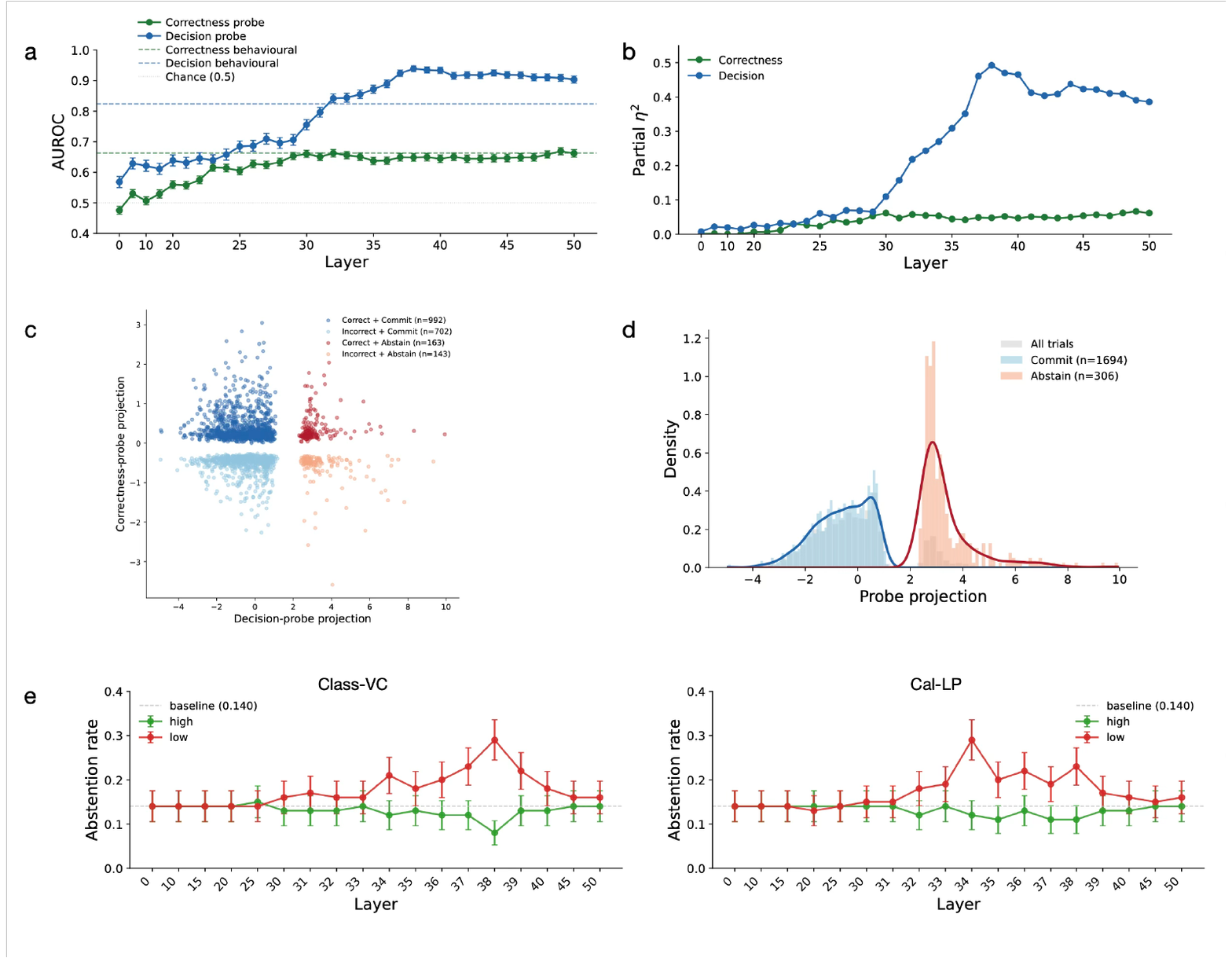}
    \caption{\textbf{Activation-level structure of decision and truth representations, and causal control of the commit/abstain decision in Gemma~4 31B.} Panels (a--d) analyse residual-stream Phase~1 activations at the post-answer (PA) position; steering in panel (e) is applied at the final token of the Phase~2 prompt, where the commit/abstain decision is generated (see Methods).
    \textbf{(a)} \textbf{Linear-probe AUROC across layers.} PA Decision probe (blue) and correctness probe (green), trained to predict Phase~2 abstention and Phase~1 correctness respectively. Error bars are Hanley-McNeil SEMs. Dashed lines indicate joint behavioural baselines obtained by fitting a logistic regression on Class-VC + Num-VC + Cal-LP for each target. The correctness probe approaches its behavioural ceiling at $\approx 0.66$; the decision probe exceeds its behavioural baseline by $\approx 0.12$ AUROC, reaching $0.94$ at L38.
    \textbf{(b)} \textbf{Partial $\eta^2$ from a 2$\times$2 ANOVA on each probe's output} (factors: correctness $\times$ decision). The decision probe's variance is dominated by the decision factor ($\eta^2 = 0.49$ at L38); the correctness probe's variance attributable to correctness plateaus at $\eta^2 \approx 0.06$.
    \textbf{(c)} \textbf{Trial-level decomposition at L38.} Each point is one of 2000 trials, plotted by its projection onto the decision-probe weight vector (x-axis) and the correctness-probe weight vector (y-axis)(see Methods). The four (correctness $\times$ decision) quadrants are populated approximately independently; clusters along the horizontal axis correspond to commit and abstain classes, separated along the direction defined by the decision probe (cosine similarity with correctness direction $= -0.01$).
    \textbf{(d)} \textbf{Distribution of the decision-probe projection at L38.} Histograms show all trials (grey) and class-conditional distributions for commit (blue) and abstain (red); coloured lines show class-conditional kernel density estimates. The decision projection is strongly bimodal, with the two modes separated by 3.06 pooled-SD units and matching the commit/abstain classes in both location and weight (mode weights 84.3\% / 15.7\% vs.\ class proportions 84.7\% / 15.3\%).
    \textbf{(e)} \textbf{Causal steering in Gemma~4 with the orthogonal components VC$_\mathrm{orth}$ (e) and LP$_\mathrm{orth}$ (f)} (scale = 4). VC$_\mathrm{orth}$ produces a clear effect on the commit/abstain decision: low-confidence steering elevates abstention above baseline, while high-confidence steering has a smaller effect, plausibly reflecting a floor imposed by the already low baseline abstention rate. LP$_\mathrm{orth}$ also shifts abstention, showing that the decision is causally sensitive to a separable Cal-LP-related direction. Error bars are SEM across 200 test trials.}
    \label{fig:activation}
\end{figure}

\begin{figure}[!t]
    \centering
    \includegraphics[width=0.8\textwidth]{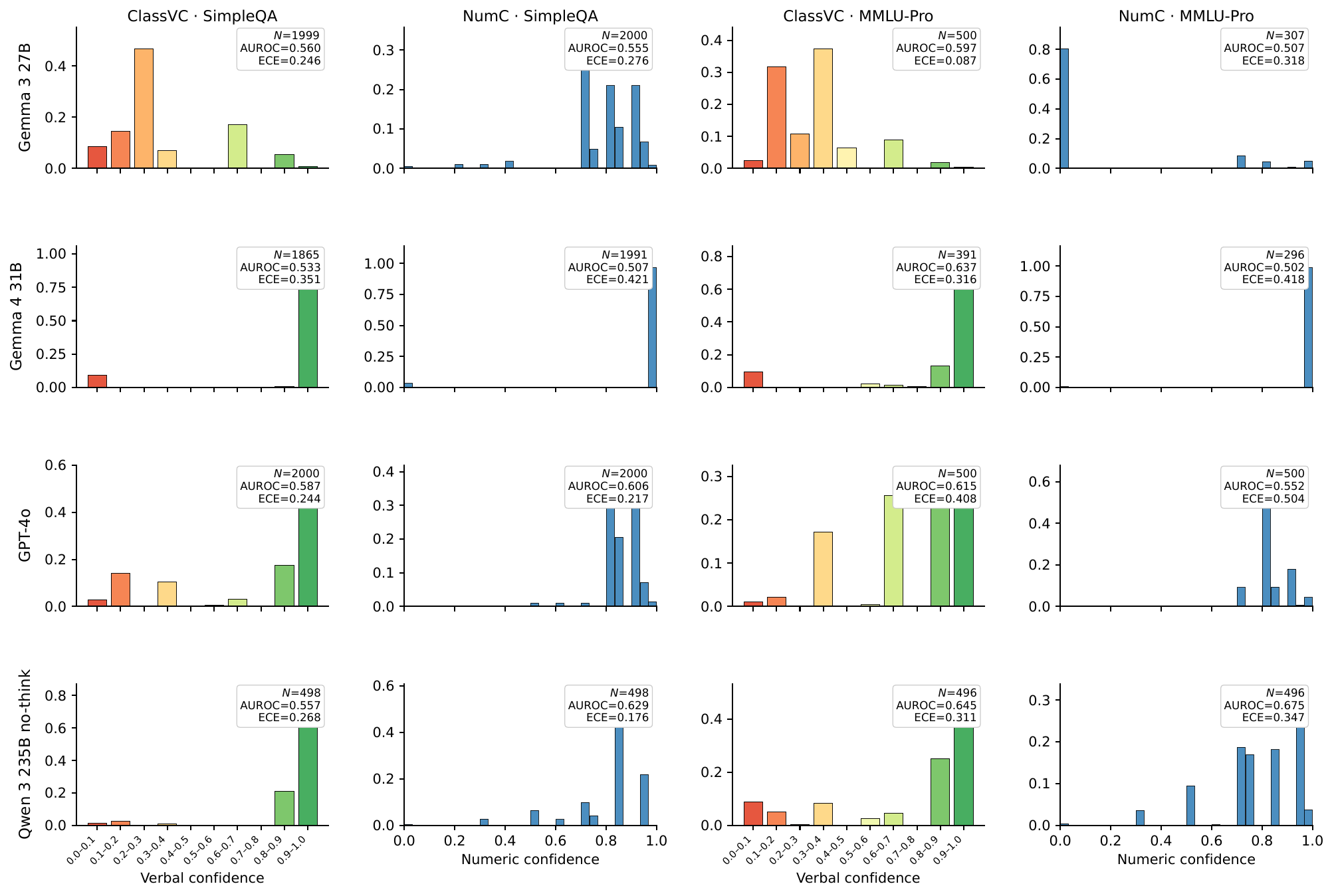}
    \caption{\textbf{Non reasoning models, SimpleQA and MMLU-Pro: verbal confidence distributions and metrics (class-based prompt and numeric prompt)}. Inset boxes report metrics}
    \label{fig:all_NRMmodels_vcplot}
\end{figure}

We next characterised the relationship between the two probe directions. The cosine similarity between the correctness-probe weight vector $\mathbf{w}_\mathrm{corr}$ and the decision-probe weight vector $\mathbf{w}_\mathrm{dec}$ at L38 (i.e. the peak layer for predicting abstention) was $-0.01$ --- the two directions are approximately orthogonal in activation space. Visualising trials projected onto both directions confirms this: trials separate cleanly along the decision axis (horizontal) while showing comparatively weak separation along the correctness axis (vertical), with the four (correctness $\times$ decision) quadrants populated approximately independently (Figure~\ref{fig:activation}c).

Notably, the decision projection at L38 is strongly bimodal (Hartigan's dip test $D = 0.13$, $p < 10^{-6}$; Gaussian mixture model with two components preferred over one by $\Delta\mathrm{BIC} = 348$)(Figure~\ref{fig:activation}d). The two modes are separated by 3.06 pooled-SD units --- near-categorical separation. Most strikingly, the smaller mode's weight (15.7\%) matches the model's behavioural abstention rate (15.3\%), and the mode means align with the commit and abstain class means within 0.02 SD units. The model's decision-aligned representation at the PA position is therefore not merely a graded confidence gradient, but a near-categorical commit/abstain signal --- and this binary organisation is already present after answer commitment, before the model is ever presented with the Phase~2 abstention prompt. Gemma~3 27B showed the same broad geometric pattern: approximately orthogonal decision and correctness directions, together with a strongly structured decision projection aligned with the later commit/abstain classes (Supplemental Results; Figure~\ref{fig:activation_G3}).

\subsection*{Causal steering: a VC-specific activation direction controls abstention in Gemma~4}
The preceding analyses establish a strong behavioural link between verbal confidence and abstention. The decoding and structural analyses further show that the PA residual stream --- a position previously implicated causally in verbal-confidence generation \citep{kumaran2026llms} --- contains a representation that strongly predicts the later commit/abstain decision and is approximately orthogonal to the correctness representation. We next asked whether this verbal-confidence-related internal state plays a causal role in shaping the later commit/abstain decision.

To test this, we used activation steering \citep{turner2023steering}: a technique in which a candidate direction in activation space is injected into the residual stream during a forward pass, and the resulting behavioural change is measured. If a direction is causally implicated in producing a behaviour, perturbations along it should systematically shift that behaviour. Our central question was whether a verbal-confidence-specific direction can alter abstention independently of Cal-LP. We therefore constructed two Phase~1 PA-position steering vectors: a Class-VC vector, defined as the difference between mean activations on high-VC and low-VC trials, and a Cal-LP vector, defined analogously from high- and low-Cal-LP trials (top vs.\ bottom $N=100$ trials for each measure; see Methods). 

Class-VC and Cal-LP steering vectors in Gemma~4 were moderately aligned in the layers where abstention was most decodable, with cosine similarity ranging from $0.35$ to $0.48$ across L31--L40. We therefore decomposed each vector into a component shared with the other and an orthogonal residual. Steering along the orthogonal residuals, VC$_\mathrm{orth}$ and LP$_\mathrm{orth}$, tests the causal contribution of signal-specific activation structure: whether verbal-confidence-related activity can alter abstention independently of Cal-LP-related activity, and conversely.

We restricted this orthogonalised steering analysis to Gemma~4. In Gemma~3, the Class-VC and Cal-LP steering vectors were highly aligned in the peak abstention-decoding region, reaching cosine similarity of approximately $0.80$ at L32. Because such strong collinearity makes the orthogonal residuals a less interpretable basis for isolating VC-specific versus Cal-LP-specific causal effects, we did not pursue orthogonalised steering in Gemma~3. Layerwise cosine similarities for both models are shown in Figure~\ref{fig:activation_G3}.

Across L31--L40, low-confidence steering along VC$_\mathrm{orth}$ increased abstention above baseline, whereas high-confidence steering produced a smaller reduction, plausibly reflecting the already low baseline abstention rate in Gemma~4. Steering along LP$_\mathrm{orth}$ also altered abstention behaviour (Figure~\ref{fig:activation}f), consistent with prior work showing that steering along a Cal-LP-derived confidence direction can causally shift abstention rates in a related paradigm \citep{kumaran2026causal}. Critically, the present result goes beyond that finding: the VC-specific component alone was sufficient to causally shift the later commit/abstain decision, even after removing variance shared with Cal-LP. Thus, verbal confidence is not merely correlated with abstention at the behavioural or representational level: a residual-stream direction unique to VC carries functionally operative commitment signal in Gemma~4.

\section*{Discussion}
\begin{figure}[!t]
    \centering
    \includegraphics[width=\textwidth]{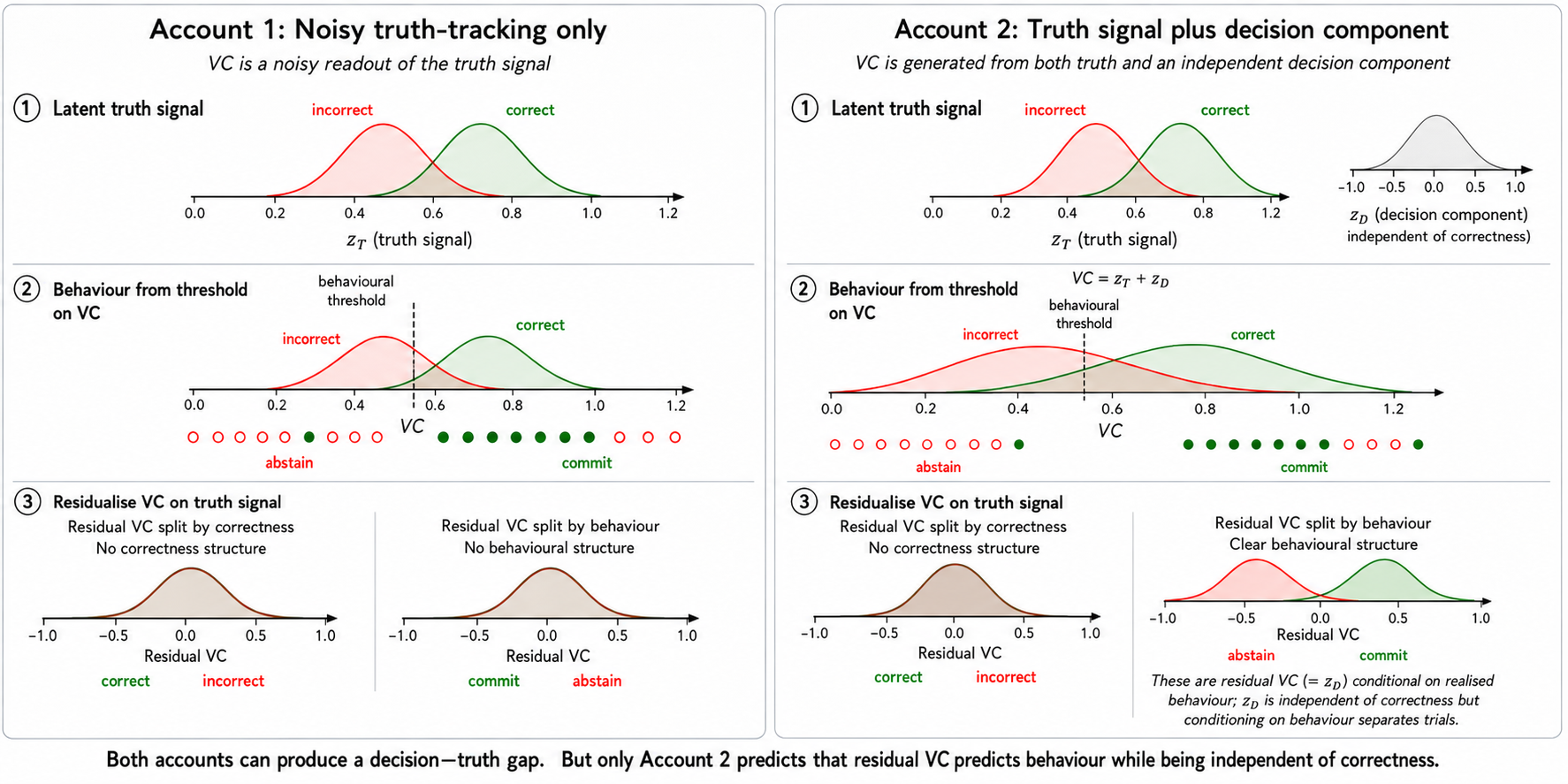}
    \caption{\textbf{Two generative accounts of verbal confidence (VC) that can both produce a decision--truth gap, distinguished by residualisation.} 
    Both accounts share the same basic structure: a latent truth signal $z_T$ separates correct and incorrect trials, and the model commits to its answer when VC exceeds a behavioural threshold and abstains otherwise. They differ in what VC is: in Account~1, VC is a noisy readout of $z_T$ alone; in Account~2, VC is the sum of $z_T$ and an independent decision component $z_D$ that is uncorrelated with correctness.
    \textbf{Row 1:} The latent truth signal $z_T$ is the same under both accounts: correct trials (green) have higher $z_T$ on average than incorrect trials (red), with substantial overlap. Account~2 additionally posits an independent decision component $z_D$ (top right, grey), drawn from a distribution centred at zero and uncorrelated with correctness.
    \textbf{Row 2:} VC is generated from the latent signal(s) and a behavioural threshold (dashed line) determines the commit/abstain decision. In Account~1, $\mathrm{VC} = z_T$ plus measurement noise, so the VC distribution closely tracks $z_T$. In Account~2, $\mathrm{VC} = z_T + z_D$, so the VC distribution is wider than $z_T$ alone, reflecting added variance from $z_D$. Under both accounts, the same thresholding rule produces a behavioural pattern in which incorrect responses are more often abstained from and correct responses more often committed to, while neither class is purely correct or purely incorrect --- consistent with the partial truth--abstention coupling observed empirically (median $\Delta = P(\mathrm{abstain}\mid\mathrm{incorrect}) - P(\mathrm{abstain}\mid\mathrm{correct}) \approx +0.13$ across non-reasoning model cells). Both accounts therefore produce a positive decision--truth gap: a signal that drives behaviour through a threshold predicts commit/abstain more strongly than it predicts the underlying truth.
    \textbf{Row 3:}  The two accounts diverge sharply when VC is residualised on the latent truth signal $z_T$ and the residual is examined as a predictor of correctness and behaviour. In practice, $z_T$ is unobserved; we use calibrated log-probability confidence (Cal-LP) as an empirical proxy. Cal-LP is the strongest single-trial correctness signal available in our non-reasoning models (correctness AUROC $0.62$--$0.80$) and accounts for the dominant unique correctness-related variance in joint regressions including VC (42--87\% of total $R^2$; Table~\ref{tab:nrm_auroc_variance}). The schematic's prediction does not require Cal-LP to equal $z_T$; it requires only that Cal-LP captures the bulk of correctness-related variance shared with VC, which our variance partitioning analysis confirms. Residualisation is performed by OLS regression of VC on Cal-LP within each cell. \textbf{\emph{Under Account~1}}, residual VC contains only measurement noise: it has no structure with respect to correctness (left density) and no structure with respect to behaviour (right density). \textbf{\emph{Under Account~2}}, the residual should be dominated by the decision component \(z_D\), and should therefore show no above-chance discrimination of correctness(left density: distributions overlap entirely); however, conditioning on the realised commit/abstain decision induces apparent separation in the residual (right density: shifted distributions for abstain and commit). This separation is not because $z_D$ carries correctness information --- it does not --- but because trials in which $z_D$ pushed VC above the behavioural threshold were classified as commit, while trials in which $z_D$ pulled VC below threshold were classified as abstain. Selection on a thresholded behaviour therefore induces a behavioural difference in $z_D$ even though $z_D$ is unconditionally independent of correctness.
    \textbf{The empirical signature that distinguishes the accounts is therefore the following:} Under Account~1, residual VC predicts neither correctness nor behaviour. Under Account~2, residual VC does not predict correctness but reliably predicts the commit/abstain decision. The reciprocal residual analyses in Figure~\ref{fig:residual_auroc} and Table~\ref{tab:reciprocal_residual_analysis} test this prediction empirically: across the eight non-reasoning model cells, residual Class-VC continues to predict abstention reliably (median residual abstention AUROC $= 0.64$) while showing weak or directionally inconsistent correctness alignment (median residual correctness AUROC $= 0.49$), matching the qualitative pattern predicted by Account~2.}
    \label{fig:residualization_schematic}
\end{figure}

\begin{figure}[!t]
    \centering
    \includegraphics[width=0.8\textwidth]{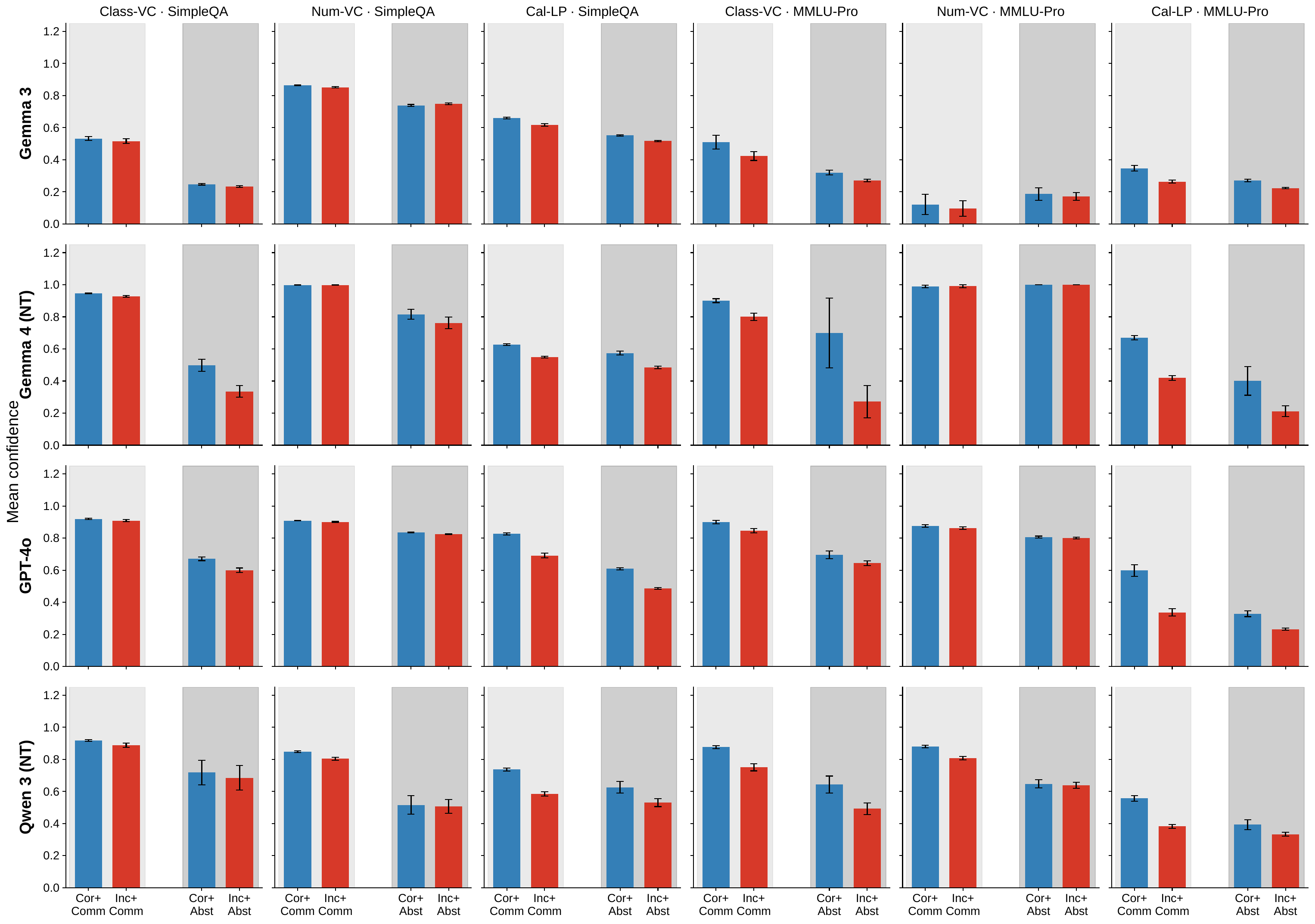}
\caption{\textbf{Mean confidence in each (correctness $\times$ decision) quadrant, by signal, model, and dataset.}
For each (model, dataset) cell, mean confidence is shown for the four (correctness $\times$ decision) quadrants: correct $+$ commit, incorrect $+$ commit, correct $+$ abstain, incorrect $+$ abstain. Bars are ordered with the two commit quadrants on the left (light grey background) and the two abstain quadrants on the right (dark grey background); within each block, blue bars are correct trials and red bars are incorrect trials. Error bars are SEM. A purely decision-aligned signal has a blockwise \emph{same--same, lower--lower} pattern: the two commit bars are similarly high, while the two abstain bars are similarly low, regardless of correctness. A purely truth-aligned signal has a within-block \emph{sawtooth} pattern: correct bars exceed incorrect bars within both the commit and abstain blocks. A signal carrying both decision- and truth-related structure tends to show a mixed \emph{high, lower, lower, lowest} profile, with correct committed trials highest, incorrect abstained trials lowest, and the two intermediate quadrants separating according to the relative strength of the two effects. The verbal measures (Class-VC, Num-VC) tend to show the decision-aligned step pattern; Cal-LP shows a more variable mixture of decision- and truth-related structure. These raw means underlie the partial $\eta^{2}$ decomposition in Figure~\ref{fig:nrm_radar}a and Table~\ref{tab:eta2_effect_sizes}.}
\label{fig:NRM_raw_means_grid}
\end{figure}

We asked whether verbal confidence (VC) in large language models is best understood as an estimate of answer correctness or as a signal more closely tied to downstream behavioural control. Across four non-reasoning models, two datasets, two abstention prompt framings, and two confidence formats, the evidence consistently favoured the latter interpretation. VC predicted the model's subsequent commit/abstain decision substantially more strongly than it predicted objective correctness, whereas calibrated log-probability confidence (Cal-LP), derived from the answer-token distribution, showed a contrasting profile: its prediction of abstention remained closely coupled to its prediction of correctness. These contrasting profiles point to structurally different signals: Cal-LP behaves as a truth-evidence signal whose behavioural relevance is inherited from its correctness-tracking, whereas VC carries an additional decision-aligned component that is not reducible to truth-tracking. Figure~\ref{fig:residualization_schematic} formalises these two structural possibilities for VC---a pure truth signal (Account~1) versus a truth-plus-decision signal (Account~2)---and shows how the reciprocal residualisations distinguish them. 
After removing the variance in VC shared with Cal-LP, residual VC retained its abstention prediction while its correctness discrimination collapsed to near chance---a directional pattern that rules out the strongest Account 1 interpretation under the Cal-LP-as-answer-evidence proxy. The converse residualisation preserved Cal-LP's correctness alignment with its behavioural relevance remaining coupled to its truth-tracking, the profile expected of a signal whose behavioural relevance is downstream of its correctness-tracking.

Multiple lines of evidence converged on this conclusion. Analyses of variance structure showed that VC was organised primarily by the later commit/abstain decision, whereas Cal-LP retained substantially stronger correctness-related structure. The VC decision--truth gap --- the excess of abstention prediction over correctness prediction --- generalised to four reasoning models across four benchmarks of varying difficulty, with a median AUROC gap of +0.16 across all models combined. Mechanistic analyses in Gemma~3 and Gemma~4 provided convergent evidence. Post-answer activations, previously implicated in verbal-confidence generation \citep{kumaran2026llms}, encoded the model's future commit/abstain decision before the abstention prompt was presented, and steering along the VC direction orthogonal to Cal-LP causally shifted abstention behaviour in Gemma~4. Together, these findings indicate that VC and Cal-LP index distinct quantities in non-reasoning models: Cal-LP behaves more like an answer-evidence signal, whereas VC is better understood as a behaviour-facing readout of internal commit-readiness, rather than a noisy estimate of correctness. The behavioural dissociation extends to reasoning models, where verbal confidence is widely treated as the primary practical confidence signal (e.g. \citep{yoon2025reasoning}). Where earlier work established that confidence-related signals can predict and causally regulate abstention \citep{kumaran2026causal}, the present results clarify the nature of verbal confidence itself: it is preferentially aligned with commitment behaviour rather than with answer correctness.

Our claim is not that VC is devoid of truth information. VC still discriminates correct from incorrect answers above chance, consistent with prior reports that verbal confidence carries some objective correctness tracking information \citep{steyvers2025metacognition, yoon2025reasoning, tian2023just, xiong2023can}. The critical observation is comparative: VC's behavioural prediction substantially exceeds its truth prediction. This pattern fits a simple interpretation: Cal-LP reflects answer evidence that is relevant both to whether the answer is correct and to whether the model later commits to it, whereas VC reflects a downstream post-answer readout in which commitment-related information is amplified relative to truth-related information. Notably, a subjective estimate of $P(\text{correct})$ need not be well calibrated to objective truth to count as a confidence signal \citep{pouget2016confidence, kepecs2012computational}; our claim is therefore not that VC is ``not confidence'', but that VC is best understood as a behaviour-facing confidence signal --- organised more strongly around the model's later commitment behaviour than around whether the answer is actually correct.

Cal-LP's coupled profile in non-reasoning models is itself a non-trivial finding. Prior work has shown that answer-token log-probabilities can discriminate correctness in LLMs \cite{steyvers2025metacognition}, and that confidence-like signals can predict abstention \citep{kumaran2026causal}. The new and substantive observation here is that Cal-LP predicts these two outcomes in proportion: its decision--truth gap is centred near zero, indicating that its behavioural relevance remains closely coupled to its truth-tracking rather than showing the separate decision-over-truth component observed for VC. This coupling is not a priori predictable from how Cal-LP is computed --- a confidence signal extracted at the answer-token position could in principle show the same decision-over-truth asymmetry observed for VC --- and it is the structural contrast that makes VC's dissociation interpretable: VC's excess behavioural prediction is not a generic property of LLM confidence signals, but a property specific to verbal confidence that distinguishes it from option-distribution-derived measures.

In reasoning models, by contrast, the relationship between log-probability-derived confidence measures and answer evidence is less direct. The single-token answer-position distribution that yielded a clean answer-evidence proxy in non-reasoning models is heavily saturated in reasoning models, in which uncertainty is distributed across an extended reasoning trajectory rather than concentrated at the answer token \citep{fu2026multiple, farquhar2024detecting}. This has motivated growing interest in sequence-level and distributional confidence measures \citep{kang2026scalable, fu2025deep}, although no consensus measure has yet emerged. These measures index uncertainty over the generation as a whole and are not as tightly tied to the moment of answer selection as the option-token distribution in a non-reasoning model. Whether such sequence-level signals retain the answer-evidence-like profile of Cal-LP in non-reasoning models --- predicting correctness and abstention through a single underlying truth-evidence channel --- is an open question, and an important target for future work. What our reasoning-model results do establish is that the VC half of the dissociation is robust: even in reasoning models, where verbal confidence has become a leading uncertainty measure in practice \citep{yoon2025reasoning}, VC's alignment with later commitment substantially exceeds its alignment with answer correctness. This finding is therefore not a particular property of non-reasoning models, but a feature of how verbal confidence relates to behaviour across the model classes we tested.

The mechanistic results further constrain this interpretation. At the post-answer position --- previously shown to causally support verbal-confidence generation \citep{kumaran2026llms} --- linear probes in Gemma~3 and Gemma~4 decoded the model's later abstention decision with high accuracy (i.e. AUROC $=0.94$ in Gemma~4), substantially exceeding decoding of answer correctness (AUROC $=0.66$). This is striking because the activations were extracted immediately after the model had produced its answer and before it had seen the Phase~2 abstention prompt. A representation strongly aligned with the future commit/abstain decision was therefore already present in the residual stream before the model was explicitly asked to make that decision. The structure of this representation mirrored the behavioural dissociation. Post-answer activations were organised predominantly by the later commit/abstain decision (partial $\eta^2=0.49$) rather than by correctness ($\eta^2=0.06$), and the decision and correctness directions were approximately orthogonal in activation space. Moreover, the decision-related signal was not merely a weak graded confidence axis: it formed a near-categorical separation between trials the model would later commit to and trials it would later abstain from, with modes matching the two behavioural classes. Crucially, this state was not merely predictive. Steering along the VC-specific component of the post-answer activation space causally shifted later abstention behaviour in Gemma~4. Together, these results suggest the following account. The model first selects an answer using an option-distribution-derived evidence signal, reflected most directly in Cal-LP. After answer commitment, this evidence is transformed into a post-answer self-evaluation state that is more tightly organised around future commitment than around objective correctness. Verbal confidence is one readout of this state, but not the whole state: the post-answer representation predicts abstention beyond surface confidence measures and can causally influence the later decision. Thus, the relevant internal signal is behaviourally operative and confidence-like, but imperfectly truth-anchored. Our mechanistic results also bear on the broader question of LLM introspection \citep{lindsey2026emergent, macar2026mechanisms}: verbal confidence appears to be a partial self-report of an internal commit-readiness state rather than of a pure correctness estimate. This illustrates a broader caveat for studies that use LLM self-reports as evidence about internal states: the prima facie content of a self-report may not correspond to what the report actually tracks. 

Our results also connect to first-order versus second-order accounts of confidence in humans and non-human primates \citep{fleming2017self, webb2023natural}. In second-order accounts, reported confidence is not a direct readout of the evidence that drove the initial choice, but is constructed downstream from that choice and other cues about the decision process, such as fluency or apparent response quality \citep{fleming2017self}. This framework offers a useful conceptual scaffold for our findings. Cal-LP resembles a first-order answer-evidence signal, in that its behavioural relevance scales with its ability to discriminate correctness. VC, by contrast, has the key signature of a downstream confidence readout: it is generated after the choice, only partially overlaps with Cal-LP, retains a commitment-aligned component after that overlap is removed, and is disproportionately aligned with the model's later behavioural disposition to commit. Identifying the cues that LLMs use in constructing this post-answer confidence report --- analogues of fluency, generation ease, or pattern-match strength --- is an important question for future work.

Human metacognition research also provides an important precedent for treating confidence as a control signal for future behaviour, rather than only as a retrospective report of accuracy \citep{yeung2012metacognition}. Confidence predicts later changes of mind in value-based decisions \citep{folke2016explicit} and regulates subsequent speed--accuracy policy after low-confidence choices \citep{desender2019confidence}. These studies establish that confidence can guide behaviour. However, they typically do not test the specific dissociation examined here: whether the same confidence signal predicts a later behavioural decision more strongly than it predicts objective correctness on the same trials. Our decision--truth gap directly compares confidence--behaviour coupling with confidence--truth coupling, raising a broader question for biological metacognition: whether confidence reports and confidence-guided behaviours are always best understood as estimates of correctness, or whether they can sometimes reflect behaviourally useful control signals whose relationship to objective truth is weaker.

The human metacognition literature has developed tools for measuring confidence--accuracy coupling --- in particular, type-2 sensitivity and meta-$d'$ \citep{maniscalco2012signal, fleming2017self} --- typically applied in binary-choice discrimination tasks. The question we address here is structurally different: rather than measuring how well a single confidence signal tracks accuracy at a given task, we ask how the same confidence signal differentially aligns with two outcomes --- objective correctness and the model's later commitment behaviour --- on the same trials. Our setting also spans 4-option, 10-option, variable-option, and freeform answers, outside the binary-choice paradigm in which meta-$d'$ is typically applied. AUROC provides a common non-parametric discrimination measure for both outcomes, allowing the decision--truth gap to be defined comparably across tasks and model classes.

Our results have direct implications for how verbal confidence is interpreted in current LLMs. Users and downstream systems often treat verbal confidence as a proxy for the probability that an answer is correct, but our findings show that this interpretation is not generally warranted: VC's predictive power for behaviour does not imply comparable predictive power for truth. A model may verbally express high confidence in an answer it is prepared to commit to, even when that answer is incorrect — and our findings indicate that this misalignment between expressed confidence and objective accuracy is not merely a calibration failure but a more fundamental feature of how verbal confidence is internally organised.

These findings also raise a training question. Although our experiments do not identify the training mechanisms that produce this dissociation, one possibility is that current instruction-tuning and reinforcement-learning objectives reward confidence-like signals that regulate response style, caution, and abstention without requiring those signals to be tightly anchored to correctness. Improving uncertainty communication may therefore require objectives that target not only surface verbal reports, but also the internal post-answer representation from which confidence and abstention behaviour are read out. A central goal for future work is to determine whether this representation can be anchored more tightly to correctness while preserving its capacity to support adaptive behaviours such as withholding unreliable answers.

\clearpage
\section*{Methods}
\subsection*{Datasets}
We evaluated two model classes: non-reasoning models (NRMs) and reasoning models (RMs). NRMs were evaluated on SimpleQA and MMLU-Pro. RMs were evaluated on the same two datasets and additionally on SuperGPQA-hard and HLE. 

\paragraph*{SimpleQA.}
SimpleQA is an open-domain factual question-answering benchmark \citep{wei2024measuring}(\url{https://osf.io/sh6z3/overview?view_only=323639b7ba714ec9926913269f3ae0b2}). We converted SimpleQA into a 4-option multiple-choice benchmark by pairing each correct answer with three foils \citep{kumaran2026causal}. For the NRM analyses, we used a 2{,}000-question version split into a disjoint $N=1{,}000$ Phase~0 calibration set and $N=2{,}000$ Phase~1 behavioural set. For the RM analyses, we used $N=1{,}000$ Phase~1 questions and a disjoint $N=1{,}000$ Phase~0 calibration set.

\paragraph*{MMLU-Pro.}
MMLU-Pro is a 10-option multiple-choice reasoning benchmark \citep{wang2024mmlu}(\url{https://huggingface.co/datasets/TIGER-Lab/MMLU-Pro}). We used a STEM-category subset sampled from the cleaned dataset. MMLU-Pro was used in both the NRM and RM analyses, with $N=500$ Phase~1 questions and a disjoint $N=500$ Phase~0 calibration set.

\paragraph*{SuperGPQA-hard.}
SuperGPQA-hard was used only in the RM analyses (\url{https://huggingface.co/datasets/m-a-p/SuperGPQA}). It is a 500-question subset drawn from the hard tier of SuperGPQA \citep{du2025supergpqa}, spanning the natural sciences, engineering, medicine, economics, philosophy, and law. Option counts vary across items, with 4--10 options per question; in our subset, the median number of options was 10 and 86\% of items had 10 options. We used $N=500$ Phase~1 questions plus a disjoint $N=500$ Phase~0 calibration set drawn from the same hard tier.

\paragraph*{Humanities Last Exam (HLE).}
HLE is a frontier-difficulty freeform benchmark spanning mathematics, physics, biology, chemistry, computer science, and the humanities \citep{center2026benchmark}(\url{https://huggingface.co/datasets/cais/hle}). Unlike the MCQ datasets, HLE answers are arbitrary text strings rather than option labels. We used the 1{,}451-question text-only subset, assigning 500 questions to Phase~0 calibration and the next 500 to Phase~1 behavioural evaluation. For correctness evaluation, model answers were judged against the reference answer using GPT-4o as judge.

\paragraph{Dataset scope and bias controls}
The datasets used in this study were selected as controlled experimental probes of the relationship between model confidence, answer correctness, and later commit/abstain behaviour. As with any benchmark-based evaluation, results may be influenced by dataset- and task-specific factors. To reduce this concern, our primary analyses compare confidence signals within the same model, dataset, and trial set, ensuring that each signal is evaluated against the same questions, answers, correctness labels, and abstention decisions. We also use disjoint calibration and evaluation data where calibration is required, evaluate the central effects across multiple datasets and model classes, and test robustness to alternative abstention-prompt wording.

\paragraph*{Question formats.}
SimpleQA was presented as a 4-option MCQ; MMLU-Pro as a 10-option MCQ; SuperGPQA-hard as a variable-option MCQ with answer labels A--J; and HLE as a freeform answer task. Thus, the dataset suite spans short factual MCQ, harder reasoning MCQ, variable-option expert MCQ, and open-ended frontier-difficulty questions.

\subsection*{Models}
\subsubsection*{Non-reasoning models.}
We evaluated four non-reasoning model configurations: Gemma~3 27B, Gemma~4 31B in no-think mode, GPT-4o, and Qwen~3 235B A22B in no-think mode. Gemma~3 and Gemma~4 were run locally using the HuggingFace \texttt{transformers} library. For Gemma~3, activations were accessed using the layer path \texttt{m.language\_model.model.layers}. For Gemma~4, activations were accessed using \texttt{m.model.language\_model.layers}, and we set \texttt{enable\_thinking=False} in the chat template to suppress the model's native thinking-channel output. GPT-4o was accessed via the OpenAI API. Qwen~3 235B A22B was accessed via Vertex AI. In the NRM condition, Qwen was prompted to output only a single option and not to reason, thereby placing it in its no-think operating mode.

\subsubsection*{Reasoning models.}
We evaluated four reasoning-model configurations: Gemini~3 Flash preview, Qwen~3 235B A22B in think mode, Kimi K2 Think, and GPT-Oss-120B. Gemini~3 Flash preview was accessed via Google's internal Beyond API. Qwen~3 235B A22B and Kimi K2 Think were evaluated via Vertex AI Model-as-a-Service endpoints. GPT-Oss-120B was evaluated in reasoning mode. Per-option log-probabilities were available for Gemini~3 Flash on all four datasets and for Qwen~3 235B Think on the three MCQ datasets, but not for Kimi K2 Think or GPT-Oss-120B. Consequently, Cal-LP analyses for RMs are restricted to cells where suitable log-probability information was exposed.

Qwen~3 235B A22B was evaluated in both no-think and think modes. The operating mode was controlled by prompt format: prompts that requested only a single option without reasoning elicited no-think behaviour, whereas prompts that instructed the model to think step by step before providing a final answer elicited think-mode reasoning. Thus, Qwen contributes both to the NRM analyses in no-think mode and to the RM analyses in think mode.

\paragraph{Computational resources}
Non-reasoning model experiments and local analyses were run using Google Colab with NVIDIA A100 GPU acceleration where required. Reasoning-model experiments were run through Google-managed model-serving infrastructure; the exact hardware configuration for these internally served models was not exposed to the author. 

\subsection*{Experimental phases}
All experiments followed the same four-phase structure. Phase~0 was used only for calibration of log-probability-based confidence. Phase~1 generated the model's answer (an option in multiple choice (SimpleQA, MMLU-Pro, SuperGPQA-Hard) or freeform (HLE)). Phase~1b elicited verbal confidence (class-based or numeric) in that answer. Phase~2 elicited the model's binary commit-versus-abstain decision.

\paragraph*{Phase~0: calibration.}
Phase~0 used a held-out calibration set drawn from the same dataset distribution as the corresponding Phase~1 behavioural set. The Phase~0 questions were disjoint from all Phase~1 questions. Models were prompted to answer the question in the same format as Phase~1, but no verbal confidence or abstention decision was elicited. For MCQ datasets, we recorded the model's log-probabilities over the available answer options at the relevant answer-readout position. These Phase~0 log-probabilities were used to fit the temperature (for MCQ datasets) or Platt-scaling (for HLE) parameters for Cal-LP. 

\paragraph*{Phase~1: answer generation.}
In Phase~1, the model answered each question. For non-reasoning models, prompts instructed the model to output only the answer option, with no preamble, reasoning, or explanation. This produced a constrained answer token from which option-token log-probabilities could be read directly. For reasoning models, prompts allowed chain-of-thought reasoning and required a final answer in a fixed format. Dataset-specific answer formats were used: digits for SimpleQA and MMLU-Pro, letters A--J for SuperGPQA-hard, and freeform text for HLE. The Phase~1 answer was treated as the model's committed answer for all subsequent phases.

\paragraph{Phase~0 / Phase~1: answer elicitation (non-reasoning models).}
\begin{quote}\small\ttfamily
You will be given a 4-way multiple choice question, with options 1-4.\\
OUTPUT ONLY A SINGLE DIGIT (1-4). Put it on the first line. No other characters.\\[0.3em]
Question: \{question\}\\[0.3em]
1) \{choice 1\}\\
2) \{choice 2\}\\
3) \{choice 3\}\\
4) \{choice 4\}\\
Answer:
\end{quote}

\paragraph{Phase~0 / Phase~1: answer elicitation (reasoning models).}
For reasoning models, the answer-elicitation prompt instructed the model to ``think step by step and then produce your final answer after \texttt{**Final Answer:**}'' to elicit explicit chain-of-thought reasoning before the answer commit. Answer, confidence and abstention outputs were still constrained to an option, label, numeric value, or A/B decision.

\paragraph*{Phase~1b: verbal-confidence elicitation.}
In Phase~1b, the model was shown the original question together with its Phase~1 answer (and reasoning trace for reasoning-models) and asked to report how likely that answer was to be correct. The primary verbal-confidence measure, Class-VC, used a 10-class labelled confidence scale from ``No chance'' to ``Almost certain'', mapped to bin midpoints from 0.05 to 0.95. For non-reasoning models, we additionally elicited numeric confidence, Num-VC, by asking the model to report an integer from 0 to 100; this value was divided by 100 for analysis (see Supplemental Methods for exact prompt used). Phase~1b was run as a separate call after answer generation. For reasoning models in the trace condition, the Phase~1 reasoning trace was included alongside the question and answer when eliciting confidence.

\paragraph{Verbal confidence elicitation (Phase~1b).}
\begin{quote}\small\ttfamily
You will be shown a question, together with an answer you previously provided.\\[0.3em]
Question: \{question\}\\[0.3em]
1) \{choice 1\}\\
2) \{choice 2\}\\
3) \{choice 3\}\\
4) \{choice 4\}\\[0.3em]
Your previous answer: \{model answer\}\\[0.3em]
Classify your confidence into one of the following classes based on how likely the answer above is to be correct (NO REASONING OR EXPLANATION):\\[0.3em]
- "No chance" (0.0-0.1)\\
- "Really unlikely" (0.1-0.2)\\
\hspace*{1em}\textit{\dots six classes omitted \dots}\\
- "Highly likely" (0.8-0.9)\\
- "Almost certain" (0.9-1.0)\\[0.3em]
Each category reflects the probability that the answer is correct.\\[0.3em]
At the very end of your output, format your confidence as\\
\textbf{*}\textbf{*}Confidence\textbf{*}\textbf{*}: \$CLASS\\
where CLASS is one of the names (only the names without the probability ranges) of the classes above.\\[0.3em]
\textbf{*}\textbf{*}Confidence\textbf{*}\textbf{*}:
\end{quote}

\paragraph*{Phase~2: commit-versus-abstain decision.}
In Phase~2, the model was again shown the original question and its Phase~1 answer -- and its reasoning trace for reasoning models -- and was asked to choose between committing to that answer or abstaining. The response format was binary: A indicated commit, and B indicated abstain. For MCQ datasets, the answer options were re-presented; for HLE, the prompt was adapted to the freeform answer format. We used a standard abstention prompt containing cautionary language, and a neutral prompt omitting this language for robustness analyses. The Phase~2 output was a binary variable, used in all behavioural analyses.

\paragraph{Phase~2: commit-versus-abstain decision (standard prompt).}
\begin{quote}\small\ttfamily
You will be shown a question, together with an answer you previously provided.\\[0.3em]
Question: \{question\}\\[0.3em]
1) \{choice 1\}\\
2) \{choice 2\}\\
3) \{choice 3\}\\
4) \{choice 4\}\\[0.3em]
Your previous answer: \{model answer\}\\[0.3em]
You now need to decide whether to output your answer to the user (A) or abstain (B). Be very careful - some questions have NO correct answer among the options. If you don't see a clearly correct answer, choose 'B' to avoid making an error. It's better to be safe than sorry. Remember, that your answer should be a single letter, no spaces, no punctuation, no words.\\[0.3em]
Decision:
\end{quote}

A neutral prompt that omitted the cautionary clause is detailed in the Supplemental Methods. 

\subsection*{Sampling parameters}
NRMs used greedy decoding with temperature 0 for all phases. For RMs, we used a uniform temperature of 0.8 in Phase~1, the answer-and-reasoning generation phase, to provide a consistent baseline for comparing generative reasoning behaviour. For Phase~1b confidence elicitation and Phase~2 abstention decisions, we omitted the temperature parameter and used model-specific API defaults. 

\subsection*{Calibration of log-probability confidence}
Calibrated log-probability confidence (Cal-LP) is an option-distribution-derived confidence signal. It is computed by temperature-scaling the model's log-probabilities over the available answer options and then taking the calibrated probability assigned to the model's chosen answer \cite{guo2017calibration}.

\paragraph*{Non-reasoning models.}
For NRMs, Cal-LP was computed from the answer-option log-probabilities at the answer-token position. For each model--dataset cell, we fit a single temperature $T^{*}$ on the disjoint Phase~0 calibration set by minimising 15-bin equal-width expected calibration error (ECE). This temperature was then applied to the Phase~1 option-token log-probabilities. If $\ell$ denotes the vector of option log-probabilities and $\hat{y}$ the chosen option, then
\[
\mathrm{Cal\mbox{-}LP}
=
\mathrm{softmax}(\ell/T^{*})_{\hat{y}}.
\]

\paragraph*{Calibration in reasoning models: Gemini 3 Flash and Qwen 235B Think mode }
For RMs, final-answer log-probabilities can be heavily saturated, especially when the model has completed a long reasoning trace. We therefore computed Cal-LP at a checkpoint within or immediately after the chain-of-thought, depending on the dataset and model. Calibration parameters were fit on Phase~0 and then applied to Phase~1. The available Cal-LP cells are summarised in Table~\ref{tab:rm_vc_callp_calibration}.

For SimpleQA and MMLU-Pro, Cal-LP was computed at a 95\% chain-of-thought checkpoint. The checkpoint was defined as 95\% of the generated reasoning trace length, before the final answer token. Following completion of generation in Phase~0 and 1 we truncated the trace at this point and re-presented it to the model with the checkpoint probe shown below (in a separate API call). Hence at this checkpoint we appended a forced-format probe requesting a single-token current answer and recorded the model's option-token log-probabilities at the next-token position. This probe was used only to read out the model's option distribution and did not alter the final Phase~1 answer used in the behavioural task.

\paragraph*{95\% checkpoint probe.}
The checkpoint probe used the following structure:
\begin{quote}\small\ttfamily
\{base\_prompt\}\\[0.3em]
<think>\{cot[:k\_r]\}</think>\\[0.3em]
Based on your reasoning so far, output your current answer in the format\\
**current answer**: followed by the option label, nothing else,\\
otherwise the trial will be terminated.
\end{quote}
Here $k_r$ denotes the token index corresponding to 95\% of the reasoning trace. We recorded the option-token log-probabilities following the forced-format answer prefix and applied the Phase~0-fitted temperature to obtain Cal-LP for the model's final chosen answer.

For SuperGPQA-hard, the 95\% checkpoint was not sufficiently close to answer commitment. The model often had not yet converged on a final option by that point, and the option distribution remained unsaturated. In Gemini, for example, saturation, measured by mean $p_{\max}$, was approximately 9\% at the 95\% checkpoint but 71\% at the 100\% post-commit checkpoint. Correspondingly, Cal-LP at the 95\% checkpoint was near chance for correctness discrimination (AUROC $\approx 0.52$), whereas Cal-LP at the 100\% checkpoint was more informative (AUROC $0.58$). We therefore computed SuperGPQA-hard Cal-LP at the 100\% post-CoT checkpoint. At this checkpoint, the model had completed its reasoning and committed to an answer, but the same Phase~0-to-Phase~1 calibration procedure was used.

\paragraph*{HLE freeform calibration.}
HLE does not have a fixed option set, so per-option Cal-LP cannot be computed. For Gemini~3 Flash on HLE, we instead computed a scalar log-probability feature from the generated answer and calibrated it using Platt scaling fit on Phase~0 and applied to Phase~1. The fitted Platt parameters were $a=1.44$ and $b=-0.46$. We report this calibrated scalar as Cal-LP for HLE, while noting that it is not directly equivalent to the option-distribution Cal-LP used for MCQ datasets.

\subsection*{Statistical Procedures}

\paragraph{Overview of statistical tests.} 
All statistical analyses were performed in Python (\texttt{scipy.stats}, \texttt{statsmodels}, \texttt{scikit-learn}). Tests of the central directional hypothesis --- that verbal confidence predicts later commit/abstain behaviour more strongly than it predicts correctness --- used one-sided tests, with the alternative specified a priori in the direction of the hypothesis. Tests with no a priori directional prediction (e.g., comparisons of gap magnitude between non-reasoning and reasoning model classes) used two-sided tests. The tail specification for each test is noted explicitly below.

\subsubsection*{Joint logistic regression and variance partitioning.}
For each (dataset, model) cell, we fit nested logistic regression models predicting two binary outcomes separately: Phase~1 correctness, and Phase~2 abstention. Predictors were the three confidence signals (class-based VC, numeric VC, and Cal-LP), all $z$-standardised within cell. We fit all $2^3 - 1 = 7$ non-empty subsets of predictors reporting McFadden's pseudo-$R^2$ \citep{mcfadden1974conditional}, AIC, and standardised coefficients with two-sided Wald $p$-values. Total McFadden $R^2$ for the full M$_{\text{VC+NumC+LP}}$ model was decomposed into unique and shared components by inclusion--exclusion over the seven nested fits; unique-variance percentages are reported as $\text{unique}_X / \text{total } R^2$ within outcome. Joint analyses required a minimum of 100 trials with valid values for all three signals and both outcomes; cells below this threshold were excluded. With three predictors, this minimum corresponds to at least 10 events per predictor for the rarer outcome in all retained cells, in line with current sample-size guidance for stable logistic-regression estimation \citep{vansmeden2019sample}.

\subsubsection*{AUROC, cross-validation, and the decision-truth gap.}
We report AUROC alongside fit statistics since outcome base rates differed substantially across cells (e.g., Phase~2 abstention rates of 6\%--83\%). This means that McFadden's pseudo-$R^2$ values are not directly comparable across outcomes \citep{mcfadden1974conditional}; AUROC is invariant to the marginal class rate. To avoid in-sample optimism, we report 5-fold cross-validated AUROC. 95\% confidence intervals on AUROC and on $\Delta$AUROC between nested models use the percentile paired bootstrap (1000 resamples of trial indices, with both AUROCs recomputed on the resampled data each iteration) \citep{hastie2009elements,demler2012misuse}. 

For each confidence signal we additionally compute the decision-truth gap as $\text{AUROC}_{\text{abstention}} - \text{AUROC}_{\text{correctness}}$. Here, $\mathrm{AUROC}_{\mathrm{correctness}}$ measures how well the signal discriminates correct from incorrect Phase~1 answers, and $\mathrm{AUROC}_{\mathrm{abstention}}$ measures how well it discriminates Phase~2 abstain from commit decisions on the same trials. Positive decision-truth gaps indicate that the signal predicts Phase~2 abstention more strongly than Phase~1 correctness; CIs on the gap are computed as a paired statistic, using the same bootstrap-resampled trial indices for both AUROCs in each iteration. We additionally pre-specified $\Delta = 0.05$ AUROC as a minimum threshold for substantively meaningful decision-truth gap. Cross-cell tests treat each (model, dataset) cell as a unit of analysis ($N=8$); we use one-sample Wilcoxon signed-rank tests (one-sided, alternative: gap $> 0.05$) for testing whether gaps reliably exceed this threshold, and paired Wilcoxon signed-rank tests (one-sided, alternative: VC gap $>$ Cal-LP gap) for comparing verbal vs.\ Cal-LP gaps.

\subsubsection*{Reciprocal residual analysis of verbal confidence and Cal-LP.}
We used reciprocal residualisation to isolate the components of verbal confidence (VC) and calibrated log-probability confidence (Cal-LP) that are not linearly shared with one another. For each non-reasoning model--dataset cell, and separately for Class-VC and Num-VC, we fit linear models of $\mathrm{VC}\sim\mathrm{Cal\mbox{-}LP}$ and $\mathrm{Cal\mbox{-}LP}\sim\mathrm{VC}$. The resulting residuals were treated as VC-specific and Cal-LP-specific signal components, respectively, and evaluated as predictors of both Phase~1 correctness and Phase~2 abstention. Correctness AUROC was computed using the residual directly; abstention AUROC was computed after sign-flipping the residual so that lower confidence corresponds to greater abstention. We also report the residual decision--truth gap, defined as residual abstention AUROC minus residual correctness AUROC. Full implementation details and preserved-power calculations are provided in the Supplemental Methods.

Paired-bootstrap 95\% confidence intervals were computed using 1000 resamples, refitting the residualisation regressions within each bootstrap sample before recomputing residual AUROCs and residual decision--truth gaps. This ensures that uncertainty reflects both the residualisation step and downstream discrimination estimates.

Cell-level tests of residual AUROCs against chance used one-sample exact Wilcoxon signed-rank tests against 0.5, with tail specification given by the directional hypothesis under test. Comparisons of residual VC against residual Cal-LP correctness AUROC used paired one-sided Wilcoxon signed-rank tests (alternative: residual Cal-LP $>$ residual VC).

\subsubsection*{ANOVA and partial $\eta^2$.}
For each confidence signal --- and separately for each activation probe's output --- we fit a two-way ANOVA with correctness, decision, and their interaction as factors, using Type II sums of squares (\texttt{statsmodels.anova\_lm}, \texttt{typ=2}). Partial $\eta^2$ was computed as $\mathrm{SS}_{\mathrm{effect}} / (\mathrm{SS}_{\mathrm{effect}} + \mathrm{SS}_{\mathrm{residual}})$ for the correctness and decision main effects, and is reported descriptively to characterise the variance structure of each signal. Where significance tests for ANOVA effects are reported, these are two-sided $F$-tests. The ANOVA underlies the variance-decomposition results presented in Figures 4, 5b, and 7b.

\subsubsection*{Mixed-effects regression.}
For cross-cell analyses pooling across (model, dataset, prompt, format) combinations, we fit linear mixed-effects models with the relevant AUROC or decision--truth gap as the outcome, fixed-effect covariates for dataset, prompt, and (for verbal-confidence models) confidence format, and model identity as a random intercept, using \texttt{statsmodels.MixedLM} with restricted maximum likelihood (REML) estimation. The continuous correctness-AUROC predictor was mean-centred. Reported $p$-values are two-sided Wald tests of regression coefficients. The mixed-effects framework was used to explicitly account 
for non-independence of cells contributed by the same model; the intraclass correlation coefficient (ICC) is reported alongside as a descriptive measure of between-model variance. Cell-level Pearson correlations are reported alongside as descriptive statistics only; inference is carried by the mixed-effects regression.

\subsubsection*{One-sample $t$-tests and Mann--Whitney comparisons.}
For pooled-level summaries of the decision--truth gap across all VC cells, we report one-sample $t$-tests against zero, one-sided (alternative: mean gap $> 0$), reflecting the directional hypothesis. For comparisons of gap magnitude between non-reasoning and reasoning model classes, we used two-sided Mann--Whitney $U$ tests, since no directional prediction was specified.

\subsubsection*{Bimodality testing of activation projections.}
To test whether the decision-probe projection at the peak layer was bimodally distributed, we used Hartigan's dip test (\texttt{diptest}; two-sided null of unimodality) and confirmed by comparing one- and two-component Gaussian mixture models via $\Delta$BIC (\texttt{sklearn.mixture.GaussianMixture}).

\subsubsection*{Linear probes and activation steering.}
Linear probes (logistic regression) were trained and evaluated using 5-fold stratified cross-validation, with predictions pooled across folds before computing AUROC. Probe AUROC standard errors use the Hanley--McNeil formula. Activation-steering analyses were evaluated on held-out test trials ($N = 200$ per condition); error bars on steering effects are SEM across trials.

\subsubsection*{Cross-cell tests and multiple comparisons.}
Cross-cell statistical tests treat each (model, dataset) combination as a single unit of analysis ($N = 8$ for non-reasoning, $N = 16$ for reasoning, $N = 7$ for reasoning models with Cal-LP available). Because we test a small number of pre-specified contrasts grounded in the directional hypothesis, we do not apply formal multiple-comparisons correction across cells; the pattern of cell-level effects is reported descriptively alongside each cross-cell test.
%---------decoding/steering--------
\subsection*{Activation based Methods in Gemma 4 31B}
\subsubsection*{Activation decoding}
We recorded residual-stream activations from a single Phase~1 forward pass at every transformer layer at the token immediately following the model's answer. In Gemma~4 (with \texttt{enable\_thinking=False}) this token is the model end of turn token introduced by the chat template. We refer to this position as PA throughout. The PA position is the analogue of the post-answer-newline (PANL) token previously shown to causally generate class-based verbal confidence in Gemma 3 27B \citep{kumaran2026llms}. To extract activations, we presented the model with a prompt comprising the question, options, and committed Phase~1 answer (`Your previous answer: $X$') --- formatted via the same chat template used in the verbal-confidence elicitation pipeline --- and ran a single forward pass without generation. As a control, we also extract residual stream activations at the \emph{question-last-token} (QLT), the last token of the question stem prior to the option list, serving as a within-prompt control. 

\paragraph*{Linear probes.}
At each layer of Gemma~3 27B and Gemma~4 31B, we trained two linear logistic-regression probes on the residual-stream activation at the post-answer (PA) position: a \emph{correctness probe} predicting Phase~1 binary correctness (\texttt{is\_correct}), and a \emph{decision probe} predicting the Phase~2 commit/abstain outcome (\texttt{is\_abstain}). Both probes used L2-regularised logistic regression ($C{=}1.0$, 2000 maximum iterations), with features standardised within each training fold. Performance was evaluated via 5-fold stratified cross-validation and reported as area under the ROC curve (AUROC) with Hanley--McNeil standard errors (Figure~\ref{fig:activation}a). For comparison, we also computed a joint behavioural baseline for each target by fitting a logistic regression on Class-VC, Num-VC, and Cal-LP together (same cross-validation procedure; dashed lines in Figure~\ref{fig:activation}a). Per-trial out-of-fold predicted probabilities from each probe were retained for downstream variance and structural analyses.

\paragraph*{Variance decomposition of probe outputs.}
To quantify how each probe's output reflects correctness vs.\ decision factors, we fit a two-way ANOVA on the out-of-fold predicted probability of each probe, with correctness (\texttt{is\_correct}), decision (\texttt{is\_abstain}), and their interaction as fixed factors. We report partial $\eta^2$ for the correctness and decision main effects (Figure~\ref{fig:activation}b), computed as $\eta^2_{\mathrm{partial}} = SS_{\mathrm{effect}} / (SS_{\mathrm{effect}} + SS_{\mathrm{residual}})$. Partial $\eta^2$ is invariant to monotonic rescaling of the probe output and quantifies the proportion of probe-output variance attributable to each factor after accounting for the other.

\paragraph*{Trial-level decomposition (joint projection scatter).}
At the peak decision-probe layer (L32 in Gemma~3 and L38 in Gemma~4), we projected each trial's PA-position activation onto both probe weight vectors $\mathbf{w}_{\mathrm{corr}}$ and $\mathbf{w}_{\mathrm{dec}}$ (each unit-normalised) to obtain two scalar projections per trial. The cosine similarity between the two weight vectors was computed as $\cos(\mathbf{w}_{\mathrm{corr}}, \mathbf{w}_{\mathrm{dec}}) = \mathbf{w}_{\mathrm{corr}}^{\top} \mathbf{w}_{\mathrm{dec}} / (\lVert \mathbf{w}_{\mathrm{corr}} \rVert \lVert \mathbf{w}_{\mathrm{dec}} \rVert)$. Trials were assigned to one of four quadrants based on their true \texttt{is\_correct} $\times$ \texttt{is\_abstain} labels and plotted as a scatter (Figure~\ref{fig:activation}c).

\paragraph*{Bimodality analysis of decision projection.}
To characterise the distributional structure of the decision-probe projection at the peak abstention predicting layer (L32 in Gemma~3 and L38 in Gemma~4), we ran two complementary tests. First, Hartigan's dip test provided a non-parametric test of unimodality; we report the dip statistic $D$ and bootstrap $p$-value against a uniform null. Second, we fit Gaussian mixture models with $n \in \{1, 2\}$ components and compared fits via the Bayesian information criterion (BIC), reporting $\Delta\mathrm{BIC}$ for $n{=}1 \rightarrow n{=}2$. For the two-component fit, we report the mean and standard deviation of each component, the component weights, and the mode separation in pooled-SD units: $\Delta = (\mu_2 - \mu_1) / \sqrt{(\sigma_1^2 + \sigma_2^2)/2}$. We additionally compared component weights and means against the behavioural class proportions and class means to test whether the recovered modes correspond to the commit/abstain classes (Figure~\ref{fig:activation}d).

\subsection*{Activation steering in Gemma~4 31B}
\paragraph*{Steering vector construction.}
We constructed two confidence steering vectors at the post-answer (PA) position from Phase~1 activations during the SimpleQA task. The Cal-LP steering vector $\mathbf{w}_\mathrm{LP}$ is the difference between mean activations on the top and bottom $N{=}100$ trials ranked by Cal-LP (correct trials only, balanced across the four answer-option positions to avoid confounds with option-specific representations). The verbal confidence (Class-VC) steering vector $\mathbf{w}_\mathrm{VC}$ is computed analogously on trials ranked by class-based verbal confidence. To prevent overfitting, the trials used to construct the steering vectors were held out from the test set used to evaluate steering effects: vector construction used the top/bottom $N{=}100$ confidence-ranked trials, while steering effects were evaluated on a separate set of $200$ randomly sampled trials from the remaining pool.

\paragraph*{Decomposition of confidence steering vectors.}
Given the two confidence steering vectors $\mathbf{w}_\mathrm{LP}$ and $\mathbf{w}_\mathrm{VC}$ at a given layer, we compute their decomposition into shared and orthogonal components. Let $\hat{\mathbf{w}}_\mathrm{LP} = \mathbf{w}_\mathrm{LP} / \|\mathbf{w}_\mathrm{LP}\|$ and $\hat{\mathbf{w}}_\mathrm{VC} = \mathbf{w}_\mathrm{VC} / \|\mathbf{w}_\mathrm{VC}\|$ denote the unit vectors. The SHARED component is defined as the angular bisector of the two unit vectors:
\begin{equation}
\mathbf{w}_\mathrm{SHARED} = \frac{1}{2} \left( \hat{\mathbf{w}}_\mathrm{LP} + \hat{\mathbf{w}}_\mathrm{VC} \right),
\end{equation}
which captures the direction along which Cal-LP and VC vary jointly. The component of $\mathbf{w}_\mathrm{LP}$ orthogonal to $\mathbf{w}_\mathrm{VC}$ is computed via Gram--Schmidt as
\begin{equation}
\mathbf{w}_{\mathrm{LP},\mathrm{orth}} = \mathbf{w}_\mathrm{LP} - (\mathbf{w}_\mathrm{LP} \cdot \hat{\mathbf{w}}_\mathrm{VC}) \, \hat{\mathbf{w}}_\mathrm{VC},
\end{equation}
and symmetrically for $\mathbf{w}_{\mathrm{VC},\mathrm{orth}}$. By construction, $\mathbf{w}_{\mathrm{LP},\mathrm{orth}}$ is orthogonal to $\mathbf{w}_\mathrm{VC}$, capturing the Cal-LP-specific signal that is not aligned with the VC direction; $\mathbf{w}_{\mathrm{VC},\mathrm{orth}}$ is orthogonal to $\mathbf{w}_\mathrm{LP}$ analogously. The fraction of each vector's variance captured by the shared subspace is $\cos^2(\mathbf{w}_\mathrm{LP}, \mathbf{w}_\mathrm{VC})$; the fraction in the orthogonal complement is $1 - \cos^2(\mathbf{w}_\mathrm{LP}, \mathbf{w}_\mathrm{VC})$. We use the orthogonal components $\mathbf{w}_{\mathrm{LP},\mathrm{orth}}$ and $\mathbf{w}_{\mathrm{VC},\mathrm{orth}}$ as steering targets to probe the unique causal contribution of each confidence measure independent of the other.

\paragraph*{Steering procedure.}
Steering was applied during Phase~2 (commit-vs-abstain decision). At inference time, on each test trial, we modified the residual stream at a single specified layer by adding a scaled steering vector to the activation at the last token of the Phase~2 prompt (i.e. the answer-colon (AC) position) --- where the next-token logits over `A' (commit) and `B' (abstain) are produced. Critically, although the steering \emph{vector} is constructed from Phase~1 PA-position activations (the position causally implicated in verbal confidence generation in our prior work \citep{kumaran2026llms}), it is \emph{applied} at the AC position during Phase~2.

Note that the cross-position construction of the steering (i.e. extract activations at PA and steer at AC) is not an arbitrary design choice but follows from (a) the theoretical commitment to PA as the VC-relevant state \citep{kumaran2026llms} and (b) the architectural constraint that steering cannot occur after the decision-readout position.

Formally, let $\mathbf{h}^{(\ell)}_\mathrm{AC}$ denote the residual stream activation at layer $\ell$ at the AC token position during Phase~2. The steered activation is
\begin{equation}
\tilde{\mathbf{h}}^{(\ell)}_\mathrm{AC} = \mathbf{h}^{(\ell)}_\mathrm{AC} + \alpha \cdot s \cdot \|\mathbf{h}^{(\ell)}_\mathrm{AC}\| \cdot \hat{\mathbf{w}},
\end{equation}
where $\hat{\mathbf{w}}$ is the unit-normalised steering vector ($\mathbf{w}_{\mathrm{VC},\mathrm{orth}}$ or $\mathbf{w}_{\mathrm{LP},\mathrm{orth}}$); $s \in \{+1, -1\}$ is the steering direction (`high' or `low' along the confidence axis); and $\alpha$ is a magnitude parameter expressed as a fraction of the residual stream norm at the steering position. We used $\alpha = 0.03 \times c$, where $c{=}4$ is a scale factor determined empirically as the smallest value producing reliable behavioural effects. 

\paragraph*{Steering evaluation.}
For each combination of steering vector $\times$ layer $\times$ direction (high vs.\ low), we measured the abstention rate across 200 held-out test trials --- distinct from the trials used to construct the steering vectors. Baseline abstention was measured on the same 200 trials without any steering applied.

\clearpage

\bibliographystyle{naturemag}
\bibliography{references_2_step}

@article{mcfadden1974conditional,
  author    = {McFadden, Daniel},
  title     = {Conditional logit analysis of qualitative choice behavior},
  booktitle = {Frontiers in Econometrics},
  editor    = {Zarembka, Paul},
  publisher = {Academic Press},
  address   = {New York},
  year      = {1974},
  pages     = {105--142},
}

@article{vansmeden2019sample,
  author  = {van Smeden, Maarten and Moons, Karel G. M. and de Groot, Joris A. H. and Collins, Gary S. and Altman, Douglas G. and Eijkemans, Marinus J. C. and Reitsma, Johannes B.},
  title   = {Sample size for binary logistic prediction models: Beyond events per variable criteria},
  journal = {Statistical Methods in Medical Research},
  volume  = {28},
  number  = {8},
  pages   = {2455--2474},
  year    = {2019},
  doi     = {10.1177/0962280218784726},
}

@book{hastie2009elements,
  author    = {Hastie, Trevor and Tibshirani, Robert and Friedman, Jerome},
  title     = {The Elements of Statistical Learning: Data Mining, Inference, and Prediction},
  edition   = {2nd},
  publisher = {Springer},
  address   = {New York},
  year      = {2009},
  doi       = {10.1007/978-0-387-84858-7},
}

@article{demler2012misuse,
  author  = {Demler, Olga V. and Pencina, Michael J. and D'Agostino, Ralph B.},
  title   = {Misuse of {DeLong} test to compare {AUC}s for nested models},
  journal = {Statistics in Medicine},
  volume  = {31},
  number  = {23},
  pages   = {2577--2587},
  year    = {2012},
  doi     = {10.1002/sim.5328},
}

@article{kadavath2022language,
  title={Language models (mostly) know what they know},
  author={Kadavath, Saurav and Conerly, Tom and Askell, Amanda and Henighan, Tom and Drain, Dawn and Perez, Ethan and Schiefer, Nicholas and Hatfield-Dodds, Zac and DasSarma, Nova and Tran-Johnson, Eli and others},
  journal={arXiv preprint arXiv:2207.05221},
  year={2022}
}

@article{steyvers2025metacognition,
  title={Metacognition and uncertainty communication in humans and large language models},
  author={Steyvers, Mark and Peters, Megan AK},
  journal={Current Directions in Psychological Science},
  pages={09637214251391158},
  year={2025},
  publisher={SAGE Publications Sage CA: Los Angeles, CA}
}

@article{webb2023natural,
  title={Natural statistics support a rational account of confidence biases},
  author={Webb, Taylor W and Miyoshi, Kiyofumi and So, Tsz Yan and Rajananda, Sivananda and Lau, Hakwan},
  journal={Nature Communications},
  volume={14},
  number={1},
  pages={3992},
  year={2023},
  publisher={Nature Publishing Group UK London}
}

@article{yeung2012metacognition,
  title={Metacognition in human decision-making: confidence and error monitoring},
  author={Yeung, Nick and Summerfield, Christopher},
  journal={Philosophical Transactions of the Royal Society B: Biological Sciences},
  volume={367},
  number={1594},
  pages={1310--1321},
  year={2012},
  publisher={The Royal Society}
}

@article{lak2014orbitofrontal,
  title={Orbitofrontal cortex is required for optimal waiting based on decision confidence},
  author={Lak, Armin and Costa, Gil M and Romberg, Erin and Koulakov, Alexei A and Mainen, Zachary F and Kepecs, Adam},
  journal={Neuron},
  volume={84},
  number={1},
  pages={190--201},
  year={2014},
  publisher={Elsevier}
}

@article{turner2023steering,
  title={Steering language models with activation engineering},
  author={Turner, Alexander Matt and Thiergart, Lisa and Leech, Gavin and Udell, David and Vazquez, Juan J and Mini, Ulisse and MacDiarmid, Monte},
  journal={arXiv preprint arXiv:2308.10248},
  year={2023}
}

@article{kepecs2008neural,
  title={Neural correlates, computation and behavioural impact of decision confidence},
  author={Kepecs, Adam and Uchida, Naoshige and Zariwala, Hatim A and Mainen, Zachary F},
  journal={Nature},
  volume={455},
  number={7210},
  pages={227--231},
  year={2008},
  publisher={Nature Publishing Group UK London}
}

@article{kepecs2012computational,
  title={A computational framework for the study of confidence in humans and animals},
  author={Kepecs, Adam and Mainen, Zachary F},
  journal={Philosophical Transactions of the Royal Society B: Biological Sciences},
  volume={367},
  number={1594},
  pages={1322--1337},
  year={2012},
  publisher={The Royal Society}
}

@article{kiani2009representation,
  title={Representation of confidence associated with a decision by neurons in the parietal cortex},
  author={Kiani, Roozbeh and Shadlen, Michael N},
  journal={science},
  volume={324},
  number={5928},
  pages={759--764},
  year={2009},
  publisher={American Association for the Advancement of Science}
}

@inproceedings{guo2017calibration,
  title={On calibration of modern neural networks},
  author={Guo, Chuan and Pleiss, Geoff and Sun, Yu and Weinberger, Kilian Q},
  booktitle={International conference on machine learning},
  pages={1321--1330},
  year={2017},
  organization={PMLR}
}

@article{wei2024measuring,
  title={Measuring short-form factuality in large language models},
  author={Wei, Jason and Karina, Nguyen and Chung, Hyung Won and Jiao, Yunxin Joy and Papay, Spencer and Glaese, Amelia and Schulman, John and Fedus, William},
  journal={arXiv preprint arXiv:2411.04368},
  year={2024}
}

@article{folke2016explicit,
  title={Explicit representation of confidence informs future value-based decisions},
  author={Folke, Tomas and Jacobsen, Catrine and Fleming, Stephen M and De Martino, Benedetto},
  journal={Nature Human Behaviour},
  volume={1},
  number={1},
  pages={0002},
  year={2016},
  publisher={Nature Publishing Group UK London}
}

@article{tian2023just,
  title={Just ask for calibration: Strategies for eliciting calibrated confidence scores from language models fine-tuned with human feedback},
  author={Tian, Katherine and Mitchell, Eric and Zhou, Allan and Sharma, Archit and Rafailov, Rafael and Yao, Huaxiu and Finn, Chelsea and Manning, Christopher D},
  journal={arXiv preprint arXiv:2305.14975},
  year={2023}
}

@article{xiong2023can,
  title={Can llms express their uncertainty? an empirical evaluation of confidence elicitation in llms},
  author={Xiong, Miao and Hu, Zhiyuan and Lu, Xinyang and Li, Yifei and Fu, Jie and He, Junxian and Hooi, Bryan},
  journal={arXiv preprint arXiv:2306.13063},
  year={2023}
}

@article{stone2022second,
  title={On second thoughts: changes of mind in decision-making},
  author={Stone, Caleb and Mattingley, Jason B and Rangelov, Dragan},
  journal={Trends in Cognitive Sciences},
  volume={26},
  number={5},
  pages={419--431},
  year={2022},
  publisher={Elsevier}
}

@article{pouget2016confidence,
  title={Confidence and certainty: distinct probabilistic quantities for different goals},
  author={Pouget, Alexandre and Drugowitsch, Jan and Kepecs, Adam},
  journal={Nature neuroscience},
  volume={19},
  number={3},
  pages={366--374},
  year={2016},
  publisher={Nature Publishing Group US New York}
}

@article{fleming2017self,
  title={Self-evaluation of decision-making: A general Bayesian framework for metacognitive computation.},
  author={Fleming, Stephen M and Daw, Nathaniel D},
  journal={Psychological review},
  volume={124},
  number={1},
  pages={91},
  year={2017},
  publisher={American Psychological Association}
}

@inproceedings{kumaran2026llms,
  author = {Kumaran, Dharshan and Conmy, Arthur and Barbero, Federico and Osindero, Simon and Patraucean, Viorica and Veli{\v{c}}kovi{\'c}, Petar},
  title = {How do LLMs compute verbal confidence?},
  booktitle = {Proceedings of the 43rd International Conference on Machine Learning (ICML)},
  year = {2026},
  note = {In press. Preprint: \url{https://arxiv.org/abs/2603.17839}}
}

@article{yoon2025reasoning,
  title={Reasoning models better express their confidence},
  author={Yoon, Dongkeun and Kim, Seungone and Yang, Sohee and Kim, Sunkyoung and Kim, Soyeon and Kim, Yongil and Choi, Eunbi and Kim, Yireun and Seo, Minjoon},
  journal={arXiv preprint arXiv:2505.14489},
  year={2025}
}

@article{lin2022teaching,
  title={Teaching models to express their uncertainty in words},
  author={Lin, Stephanie and Hilton, Jacob and Evans, Owain},
  journal={arXiv preprint arXiv:2205.14334},
  year={2022}
}

@article{kumaran2026causal,
  author  = {Kumaran, Dharshan and Daw, Nathaniel and Osindero, Simon and Veli{\v{c}}kovi{\'c}, Petar and Patraucean, Viorica},
  title   = {Causal evidence that language models use confidence to drive behavior},
  journal = {Nature Machine Intelligence},
  year    = {2026},
  note    = {In press. Preprint: \url{https://arxiv.org/abs/2603.22161}}
}

@article{beran2006rhesus,
  title={Rhesus macaques (Macaca mulatta) monitor uncertainty during numerosity judgments.},
  author={Beran, Michael J and Smith, J David and Redford, Joshua S and Washburn, David A},
  journal={Journal of Experimental Psychology: Animal Behavior Processes},
  volume={32},
  number={2},
  pages={111},
  year={2006},
  publisher={American Psychological Association}
}

@article{maniscalco2012signal,
  title = {A signal detection theoretic approach for estimating metacognitive sensitivity from confidence ratings},
  author = {Maniscalco, Brian and Lau, Hakwan},
  journal = {Consciousness and Cognition},
  volume = {21},
  number = {1},
  pages = {422--430},
  year = {2012},
  doi = {10.1016/j.concog.2011.09.021}
}

@article{desender2019confidence,
title = {Confidence predicts speed-accuracy tradeoff for subsequent decisions},
author = {Desender, Kobe and Boldt, Annika and Verguts, Tom and Donner, Tobias H.},
journal = {eLife},
volume = {8},
pages = {e43499},
year = {2019},
doi = {10.7554/eLife.43499}
}

@article{wang2024mmlu,
  title={Mmlu-pro: A more robust and challenging multi-task language understanding benchmark},
  author={Wang, Yubo and Ma, Xueguang and Zhang, Ge and Ni, Yuansheng and Chandra, Abhranil and Guo, Shiguang and Ren, Weiming and Arulraj, Aaran and He, Xuan and Jiang, Ziyan and others},
  journal={Advances in Neural Information Processing Systems},
  volume={37},
  pages={95266--95290},
  year={2024}
}

@article{du2025supergpqa,
  title={Supergpqa: Scaling llm evaluation across 285 graduate disciplines},
  author={Du, Xinrun and Yao, Yifan and Ma, Kaijing and Wang, Bingli and Zheng, Tianyu and Zhu, King and Liu, Minghao and Liang, Yiming and Jin, Xiaolong and Wei, Zhenlin and others},
  journal={arXiv preprint arXiv:2502.14739},
  year={2025}
}

@article{center2026benchmark,
  title={A benchmark of expert-level academic questions to assess AI capabilities},
  author={Center for AI Safety, Scale AI, HLE Contributors Consortium},
  journal={Nature},
  volume={649},
  number={8099},
  pages={1139--1146},
  year={2026},
  publisher={Nature Publishing Group UK London}
}

@article{farquhar2024detecting,
  title={Detecting hallucinations in large language models using semantic entropy},
  author={Farquhar, Sebastian and Kossen, Jannik and Kuhn, Lorenz and Gal, Yarin},
  journal={Nature},
  volume={630},
  number={8017},
  pages={625--630},
  year={2024},
  publisher={Nature Publishing Group UK London}
}

@article{kang2026scalable,
  title={Scalable best-of-n selection for large language models via self-certainty},
  author={Kang, Zhewei and Zhao, Xuandong and Song, Dawn},
  journal={Advances in neural information processing systems},
  volume={38},
  pages={19720--19745},
  year={2026}
}

@article{fu2025deep,
  title={Deep think with confidence},
  author={Fu, Yichao and Wang, Xuewei and Tian, Yuandong and Zhao, Jiawei},
  journal={arXiv preprint arXiv:2508.15260},
  year={2025}
}

@article{fu2026multiple,
  title={Multiple Choice Questions: Reasoning Makes Large Language Models (LLMs) More Self-Confident, specially When They are Wrong},
  author={Fu, Tairan and Conde, Javier and Martinez, Gonzalo and Grandury, Maria and Reviriego, Pedro},
  journal={IEEE Intelligent Systems},
  year={2026},
  publisher={IEEE}
}

@article{kumaran2026competing,
  title={Competing Biases underlie Overconfidence and Underconfidence in LLMs},
  author={Kumaran, Dharshan and Fleming, Stephen M and Markeeva, Larisa and Heyward, Joe and Banino, Andrea and Mathur, Mrinal and Pascanu, Razvan and Osindero, Simon and De Martino, Benedetto and Veli{\v{c}}kovi{\'c}, Petar and others},
  journal={Nature Machine Intelligence},
  pages={1--14},
  year={2026},
  publisher={Nature Publishing Group UK London}
}

@article{balsdon2020confidence,
  title={Confidence controls perceptual evidence accumulation},
  author={Balsdon, Tarryn and Wyart, Valentin and Mamassian, Pascal},
  journal={Nature communications},
  volume={11},
  number={1},
  pages={1753},
  year={2020},
  publisher={Nature Publishing Group UK London}
}

@article{lindsey2026emergent,
  title={Emergent introspective awareness in large language models},
  author={Lindsey, Jack},
  journal={arXiv preprint arXiv:2601.01828},
  year={2026}
}

@article{macar2026mechanisms,
  title={Mechanisms of Introspective Awareness},
  author={Macar, Uzay and Yang, Li and Wang, Atticus and Wallich, Peter and Ameisen, Emmanuel and Lindsey, Jack},
  journal={arXiv preprint arXiv:2603.21396},
  year={2026}
}
% \section*{Author Contributions}  % or \section*{Acknowledgements} for British spelling
% DK conceived the project, carried out the experiments, analysis and wrote the paper.

\section*{Acknowledgments}  % or \section*{Acknowledgements} for British spelling
We thank Nathaniel Daw, Maks Ovsjanikov, Kim Stachenfeld and Beningo Uria for comments on an earlier version of the manuscript. We used Gemini to help improve the clarity and readability of the manuscript. 
% \section*{Funding}  % or \section*{Acknowledgements} for British spelling
% This was provided by Google DeepMind
% \section*{Competing Interests}  % or \section*{Acknowledgements} for British spelling
% There are no competing interests
% \section*{Data Availability}
% The SimpleQA MCQ dataset used in this study is publicly available at \url{https://osf.io/sh6z3/overview?view_only=323639b7ba714ec9926913269f3ae0b2}. The other datasets are publicly available. 

% \section*{Code Availability}
% Code required to reproduce the main and supplemental figures is available at \url{https://osf.io/sh6z3/overview?view_only=323639b7ba714ec9926913269f3ae0b2}. Code for the core non-reasoning model experiments and analyses reported in this manuscript will be made available upon publication. The exact configuration used for some reasoning-model experiments will not be released, as those experiments were carried out using Google-specific infrastructure and internal model-access code. Prompts, analysis procedures, and processed outputs will be provided where possible to support reproducibility of the reported results.

\clearpage

\section{Supplemental Material}
\subsection{Supplemental Methods}

\paragraph{Numeric verbal confidence elicitation.}
For the numeric-confidence (NumC) analysis, we elicited a self-reported numeric 
confidence rating (0--100) on the same trials as labelled VC, in a separate API 
call following Phase~1. The exact prompt used was:

\begin{quote}\small\ttfamily
State your confidence as an integer between 0 and 100 based on how likely your answer is to be correct. Do NOT output anything else at all!!\\[0.3em]
That is, if your confidence is 0, that means that your answer has almost no chance of being correct. If your confidence is 100, then you are totally certain that your answer is correct. If your confidence is 70, then you are moderately confident that your answer is correct.\\[0.3em]
At the very end of your output, format your confidence as:\\[0.3em]
Confidence:\$CONFIDENCE\\[0.3em]
where CONFIDENCE is an integer between 0 and 100. Do NOT output ANYTHING ELSE (e.g.\ reasoning, explanation)!\\[0.3em]
Confidence:
\end{quote}

The integer value following the final \texttt{Confidence:} marker was extracted, 
divided by 100 to yield a confidence score on $[0, 1]$, and used as the NumC 
predictor in all subsequent analyses.
The prompt structure for the 10-option MMLU-Pro, A--J SuperGPQA-hard, and freeform HLE follows the same template with appropriate option-format adjustments.

\paragraph{Neutral abstention prompt.}
\begin{quote}\small\ttfamily
You will be shown a question, together with an answer you previously provided.\\[0.3em]
Question: \{question\}\\[0.3em]
1) \{choice 1\}\\
2) \{choice 2\}\\
3) \{choice 3\}\\
4) \{choice 4\}\\[0.3em]
Your previous answer: \{model answer\}\\[0.3em]
You now need to decide whether to output your answer to the user (A) or abstain (B). Remember, that your answer should be a single letter, no spaces, no punctuation, no words.\\[0.3em]
Decision:
\end{quote}

\subsection*{Further details of residual analysis}
\subsubsection*{Reciprocal residual analysis of verbal confidence and Cal-LP.}
To distinguish the components of verbal confidence (VC) and calibrated log-probability confidence (Cal-LP) that are shared from those that are signal-specific, we performed a reciprocal residual analysis. The analysis asks two complementary questions: (i) after removing the linear component of VC explained by Cal-LP, is the remaining VC-specific signal more aligned with objective correctness or with the later commit/abstain decision? and (ii) after removing the linear component of Cal-LP explained by VC, does the remaining Cal-LP-specific signal show the same or a different profile?

For each non-reasoning model--dataset cell, and separately for Class-VC and Num-VC, we fit two ordinary least-squares regressions on the same paired-trial sample:
\[
\mathrm{VC} = \alpha + \beta\,\mathrm{Cal\mbox{-}LP} + \varepsilon_{\mathrm{VC}},
\]
\[
\mathrm{Cal\mbox{-}LP} = \alpha' + \beta'\,\mathrm{VC} + \varepsilon_{\mathrm{LP}}.
\]
The residuals $\varepsilon_{\mathrm{VC}}$ and $\varepsilon_{\mathrm{LP}}$ were treated as the VC-specific and Cal-LP-specific components, respectively. We then evaluated each residual signal as a predictor of (a) Phase~1 correctness and (b) Phase~2 abstention. For correctness AUROC, higher residual values were treated as predicting greater correctness; for abstention AUROC, lower residual values were treated as predicting greater abstention, so AUROC was computed on the sign-flipped residual. We also report the residual decision--truth gap,
\[
\Delta_{\mathrm{DT}}^{\mathrm{resid}}
=
\mathrm{AUROC}_{\mathrm{abstention}}^{\mathrm{resid}}
-
\mathrm{AUROC}_{\mathrm{correctness}}^{\mathrm{resid}},
\]
which summarises whether the signal-specific component is more strongly aligned with later commitment behaviour or with objective correctness.

For descriptive comparison with the raw signals, we additionally report preserved predictive power relative to chance. For example, preserved abstention-predictive power was computed as
\[
\frac{\mathrm{AUROC}^{\mathrm{resid}}_{\mathrm{abstention}} - 0.5}
{\mathrm{AUROC}^{\mathrm{raw}}_{\mathrm{abstention}} - 0.5}
\times 100,
\]
with an analogous calculation for preserved correctness-predictive power in the Cal-LP residual analysis. Ratios were treated as uninterpretable where the corresponding raw AUROC was at or below chance.

Paired-bootstrap 95\% confidence intervals were computed using 1000 resamples. Within each bootstrap sample, both residualisation regressions were refit before recomputing residual AUROCs, residual decision--truth gaps, and preserved-power statistics, so that uncertainty reflects both the residualisation step and downstream AUROC estimation.

\subsubsection*{Residualised abstention prediction: additional inclusion of binary correctness and difficulty.}
We tested whether each verbal-confidence signal carries abstention-relevant information beyond correctness-related answer evidence by residualising VC on increasingly stringent sets of control variables and computing the abstention AUROC of the residual. We used three nested specifications: (1) residualisation on binary \texttt{is\_correct} alone; (2) residualisation on \texttt{is\_correct} plus Cal-LP, which provides the strongest available model-internal proxy for graded answer evidence; and (3) residualisation on \texttt{is\_correct} plus Cal-LP plus a per-question difficulty score from \citet{kumaran2026competing} (SimpleQA only). For each specification, we refit a linear regression of VC on the control variables and computed the abstention AUROC of the resulting residuals. We report the percentage of predictive power preserved as
\[
\frac{\mathrm{AUROC}_{\mathrm{residual}} - 0.5}
{\mathrm{AUROC}_{\mathrm{raw}} - 0.5}
\times 100.
\]
Paired-bootstrap 95\% confidence intervals were computed using 1000 resamples, refitting the residualisation regression within each bootstrap sample so that intervals reflect uncertainty in both the residualisation step and the AUROC estimate. 

\paragraph*{Note on cross-outcome comparisons.}
Total McFadden $R^2$ values for predicting correctness and abstention are not directly comparable across outcomes because McFadden $R^2$ depends on the null-model log-likelihood, which itself depends on outcome base rate \citep{mcfadden1974conditional}. Phase~2 abstention rates differed substantially across cells (6\%--83\% in our data) and from Phase~1 correctness rates within the same cells, so larger total $R^2$ for abstention prediction does not imply that abstention is intrinsically more predictable. We therefore restrict cross-outcome comparisons to AUROC, which is invariant to base rate, and report McFadden $R^2$ only for within-outcome decomposition.

\paragraph*{Question-level covariates.}
As a control for question-intrinsic hardness independent of any single-trial confidence signal in the SimpleQA dataset, we computed item-level difficulty as the proportion of correct responses across $n=20$ GPT-4o runs (sampling temperature 0.8) with randomised option-position assignments \citep{kumaran2026causal}. 

\clearpage
%--------SUPPLEMENTAL RESULTS
\subsection{Supplemental Results}
\paragraph*{Further results on non-reasoning models}
\paragraph*{Details of residualization analyses}
Across the eight non-reasoning model cells, the Class-VC-specific residual showed no reliable positive alignment with correctness: its residual correctness AUROC was not greater than chance at the aggregate level (mean AUROC $=0.473$, median $=0.487$; one-sided exact Wilcoxon signed-rank test against $0.5$: $W=11$, $p=0.844$). By contrast, the Cal-LP-specific residual remained robustly correctness-aligned (mean AUROC $=0.686$, median $=0.685$; $W=36$, $p=0.0039$), and discriminated correctness significantly better than the Class-VC-specific residual in every cell (mean AUROC difference $=+0.213$, median $=+0.218$; one-sided exact Wilcoxon signed-rank test: $W=36$, $p=0.0039$). Both residual signals retained reliable abstention-related information: residual Class-VC abstention AUROC exceeded chance across cells (mean $=0.664$, median $=0.672$; $W=36$, $p=0.0039$), as did residual Cal-LP abstention AUROC (mean $=0.632$, median $=0.630$; $W=36$, $p=0.0039$). However, Class-VC was not reliably more abstention-predictive than Cal-LP after residualisation (mean AUROC difference $=+0.031$, median $=+0.053$; $W=24$, $p=0.230$). Thus, the clearest reciprocal dissociation lies in truth alignment: Cal-LP retains a robust correctness-related component after removing shared variance with VC, whereas VC does not, even though both residual signals continue to carry substantial information about the later commit/abstain decision.

As a complementary robustness analysis, we also asked whether VC retains abstention-predictive power after removing variance associated with increasingly stringent correctness-related proxies: binary correctness alone, binary correctness plus Cal-LP, and, for SimpleQA, binary correctness plus Cal-LP and question-level difficulty (Table~\ref{tab:residual_correctness_proxy_analysis}). Residualising on binary correctness alone preserved most of VC's abstention signal. Adding Cal-LP reduced the residual signal but left substantial abstention-predictive power in most cells, and adding item difficulty changed the preserved fraction by less than two percentage points relative to the Cal-LP specification. 

\subsection*{Robustness of the decision--truth dissociation under a neutral abstention prompt}
We re-ran the Phase 2 of the abstention pipeline using a neutral abstention prompt that omits the cautionary framing of the standard prompt (see exact prompt in Supplemental Methods). Phase 2 abstention rates differed substantially between prompts (Table~\ref{tab:nrm_abstention_by_prompt}); for example, GPT-4o on SimpleQA shifted from 0.722 abstention under the standard prompt to 0.346 under the neutral prompt. However, the truth-aligned discrimination of the abstention decision was preserved: $\Delta = $ P(abstain $|$ incorrect) $-$ P(abstain $|$ correct) was positive in 8/8 cells under both prompts.

We computed decision--truth gaps for all three confidence signals on the same trials (Table~\ref{tab:nrm_dt_gap_comparison}). Class-VC gaps exceeded the $0.05$ threshold in 7/8 cells (median $+0.16$, one-sample Wilcoxon $W=35$, $p=0.008$). Num-VC gaps exceeded the threshold in 6/8 cells (median $+0.20$, $W=33$, $p=0.020$). Cal-LP gaps exceeded the threshold in 2/8 cells only (median $+0.01$, $W=4$, $p=0.98$) and did not reliably exceed zero ($W=19$, $p=0.47$). Paired comparisons confirmed the dissociation: verbal-confidence gaps exceeded Cal-LP gaps in 8/8 cells for Class-VC (median $\Delta=+0.14$, $W=36$, $p=0.004$) and in 7/8 cells for Num-VC (median $\Delta=+0.18$, $W=34$, $p=0.012$; one-sided Wilcoxon signed-rank).

The full nested logistic regression and variance-partitioning analysis under the neutral prompt is reported in Table~\ref{tab:nrm_auroc_variance_neutral}, and shows the same pattern: VC and Num-VC carry the dominant unique abstention-related variance across cells, while Cal-LP carries little. 

\subsection{Reasoning-model results in detail}
\paragraph*{Confidence measures in reasoning models.}
We evaluated four reasoning models --- Gemini Flash~3, Qwen~235B Think, Kimi K2 Think, and GPT-Oss-120B --- on the two-phase abstention paradigm across SimpleQA, MMLU-Pro, SuperGPQA-hard, and HLE. Phase~1 permitted a chain-of-thought trace before the model committed to an answer; this trace was included both during verbal-confidence elicitation and in the Phase~2 commit/abstain context. We elicited only the class-based verbal confidence scale (VC; Figure~\ref{fig:all_RMmodels_vcplot}), not the numeric variant. Cal-LP was available for Gemini Flash~3 on all four datasets and for Qwen~235B Think on the three MCQ datasets; Kimi K2 Think and GPT-Oss-120B are evaluated on VC only (Table~\ref{tab:rm_vc_callp_calibration}).

Unlike in the non-reasoning models, Cal-LP did not uniformly outperform VC in correctness discrimination. VC's correctness AUROC exceeded Cal-LP's in 4 of the 7 reasoning-model cells. Gemini Flash~3 showed a difficulty-sensitive pattern: Cal-LP discriminated correctness better than VC on the easier MCQ benchmarks (SimpleQA: $0.69$ vs.\ $0.64$; MMLU-Pro: $0.74$ vs.\ $0.68$), but worse on SuperGPQA-hard ($0.58$ vs.\ $0.63$) and HLE ($0.55$ vs.\ $0.62$). Qwen~235B Think showed a weaker Cal-LP profile overall: Cal-LP slightly exceeded VC on SimpleQA ($0.56$ vs.\ $0.54$), but was weaker on MMLU-Pro ($0.54$ vs.\ $0.70$) and SuperGPQA-hard ($0.56$ vs.\ $0.64$).

This divergence from the non-reasoning models is plausible given how Cal-LP is measured in reasoning systems. In non-reasoning models, the answer-token distribution is a relatively clean readout of the evidence supporting the chosen option. In reasoning models, uncertainty is distributed across the full reasoning trajectory, while the probed answer-token distribution may already be partially or fully committed by the checkpoint at which it is read (see Methods for details). Consistent with this concern, sequence and distribution-level measures may be better indices of a model's confidence \citep{farquhar2024detecting, kang2026scalable, fu2025deep}. We therefore treat reasoning-model Cal-LP as a secondary comparator rather than as a fully analogous proxy to the non-reasoning Cal-LP signal.

\paragraph{Baseline behavioural performance.}
Phase~1 accuracy and Phase~2 abstention rates varied substantially across the 16 reasoning-model cells, as expected given the breadth of the benchmark suite (Table~\ref{tab:rm_behavior}). Accuracy was lowest on HLE and highest on MMLU-Pro in several models. Despite this heterogeneity, abstention was higher after incorrect than correct answers in all sixteen cells, with $\Delta=P(\mathrm{abstain}\mid\mathrm{incorrect})-P(\mathrm{abstain}\mid\mathrm{correct})$ ranging from $+0.047$ to $+0.259$. Thus, as in non-reasoning models, the abstention decision was consistently partially discriminatory with respect to objective correctness.

\paragraph{Decision--truth gaps in reasoning models.}
Across the 16 reasoning model--dataset cells, the VC decision--truth gap was positive in 15/16 cells (median $+0.16$, one-sample Wilcoxon $p=1.1\times10^{-4}$) and exceeded the pre-specified $+0.05$ threshold in 14/16 cells ($p=1.1\times10^{-3}$; Figure~\ref{fig:RM_fig_decision_truth_gap}a). Cal-LP, computable in seven cells, showed a near-zero distribution of gaps (median $-0.02$; 2/7 positive; $p=0.77$). In paired within-cell comparisons, VC's gap exceeded Cal-LP's in 7/7 cells (median $\Delta=+0.13$, $p=0.008$) and exceeded it by at least $0.05$ in 7/7 cells ($p=0.008$). Full AUROC and gap values are reported in Table~\ref{tab:rm_auroc_variance}.

\paragraph{Residualised abstention prediction in reasoning models.}
As a complementary robustness analysis, we applied the same correctness-proxy residualisation used for non-reasoning models to the reasoning-model cells where Cal-LP was available (Table~\ref{tab:rm_residual_abstention}). VC residuals retained most of their abstention-predictive power after controlling for binary correctness and Cal-LP, with a median preserved fraction of approximately $87\%$ across seven cells (range: 36--98\%). Adding SimpleQA difficulty changed the preserved fraction by approximately seven percentage points in the two SimpleQA cells where it was tested. Thus, as in non-reasoning models, the behavioural alignment of VC persisted after removing variance associated with Cal-LP -- though as noted previously, this is only a partial proxy for correctness in reasoning models. 

\paragraph{Joint regression and variance partitioning.}
We next asked whether VC and Cal-LP carried separable predictive information about abstention and correctness in the reasoning-model cells where both signals were available. Nested logistic regressions with VC and Cal-LP as predictors showed that VC carried the dominant unique abstention-related variance across all seven cells (Supplementary Table~\ref{tab:rm_auroc_variance}). Unique $R^2_{\mathrm{VC}}$ ranged from $0.05$ in Gemini Flash~3 on SimpleQA to $0.49$ in Qwen~235B Think on SuperGPQA-hard, accounting for a median of $96\%$ of total McFadden $R^2$. Cal-LP's unique contribution to abstention was negligible in most cells (median unique $R^2_{\mathrm{LP}}=0.002$; $\leq 9\%$ of total $R^2$ in six of the seven cells), with Gemini Flash~3 on SimpleQA the only setting showing a substantial Cal-LP abstention contribution ($33\%$ of total $R^2$).

The contrast with non-reasoning models was clearest for correctness prediction. In NRMs, Cal-LP carried substantial unique variance for correctness. In reasoning models, its unique correctness contribution was small or negligible in five of seven cells. Even in Gemini Flash~3 on MMLU-Pro, the cell with the highest Cal-LP correctness AUROC ($0.74$), Cal-LP added only $0.3\%$ of total $R^2$ to the joint VC+LP model, indicating that nearly all of its correctness discrimination was already captured by VC. The only clearly meaningful unique Cal-LP correctness contribution appeared in Gemini Flash~3 on SimpleQA ($29\%$ of total $R^2$). Thus, the reasoning-model dissociation is expressed primarily through VC's strong decision alignment, with single-token Cal-LP making a weaker and less stable unique contribution to either target.

\paragraph{Variance decomposition in reasoning models.}
We applied the same two-way ANOVA decomposition used for the non-reasoning models, fitting each confidence signal with correctness, decision, and their interaction as predictors and extracting partial $\eta^2$ for the correctness and decision main effects (Figure~\ref{fig:RM_fig_decision_truth_gap}b; Table~\ref{tab:rm_eta2_effect_sizes}). For VC, the decision-related effect size exceeded the correctness-related effect size in 13/16 cells. Decision $\eta^2$ ranged from $0.00$ to $0.52$ (median $0.14$), whereas correctness $\eta^2$ ranged from $0.00$ to $0.08$ (median $0.05$). The three exceptions were Gemini Flash~3 on SimpleQA and MMLU-Pro, where the decision and correctness effects were more comparable, and Kimi K2 Think on HLE, where VC was near-saturated and provided little discriminative variance for either factor.

Cal-LP, where available, showed little organised variance for either factor: correctness partial $\eta^2$ ranged from $0.00$ to $0.03$ (median $0.006$), and decision partial $\eta^2$ from $0.00$ to $0.03$ (median $0.005$). This differs from the non-reasoning models, where Cal-LP carried substantially more correctness-related variance than VC. The reasoning-model ANOVA therefore reinforces the broader pattern: VC remains strongly structured by later abstention behaviour, while checkpoint-based Cal-LP offers a weaker and less stable decomposition in this setting.

\subsection{Gemma~3 27B Activation-based analyses: decoding correctness and commitment decision}
The two probes diverge substantially with depth (Figure~\ref{fig:activation_G3}a). The correctness probe rises modestly to a peak AUROC of $0.64$ at L35, only slightly above the joint behavioural baseline obtained by combining all three confidence signals (Class-VC + Num-VC + Cal-LP; dashed line; AUROC $=0.59$). By contrast, the decision probe rises steeply through mid-network layers, reaching AUROC $=0.95$ at L32 --- nearly $0.30$ AUROC above its corresponding behavioural baseline (AUROC $=0.65$). PA activations therefore contain a strong commit/abstain-relevant signal that is only partially expressed through the model's own verbal and Cal-LP confidence channels.

A complementary 2$\times$2 ANOVA on each probe's output (factors: correctness $\times$ decision) confirms the dissociation in variance terms (Figure~\ref{fig:activation_G3}b). The decision probe's partial $\eta^2$ for the decision factor climbs steeply through mid layers, peaking at $\eta^2 = 0.58$ at L32. The correctness probe's partial $\eta^2$ for correctness remains much smaller, peaking at only $\eta^2 = 0.04$. At the peak decision layer, decision-related variance exceeds correctness-related variance by approximately $17$-fold ($\eta^2=0.58$ vs.\ $0.03$). The two signals are therefore not equally represented at the PA position: decision-related structure dominates correctness-related structure by a large margin in the layers where abstention is most decodable.

\subsection*{Geometric and structural relationships between the two representations in Gemma 3 27B}
\begin{figure}[!t]
    \centering
    \includegraphics[angle=-90, width=\textwidth]{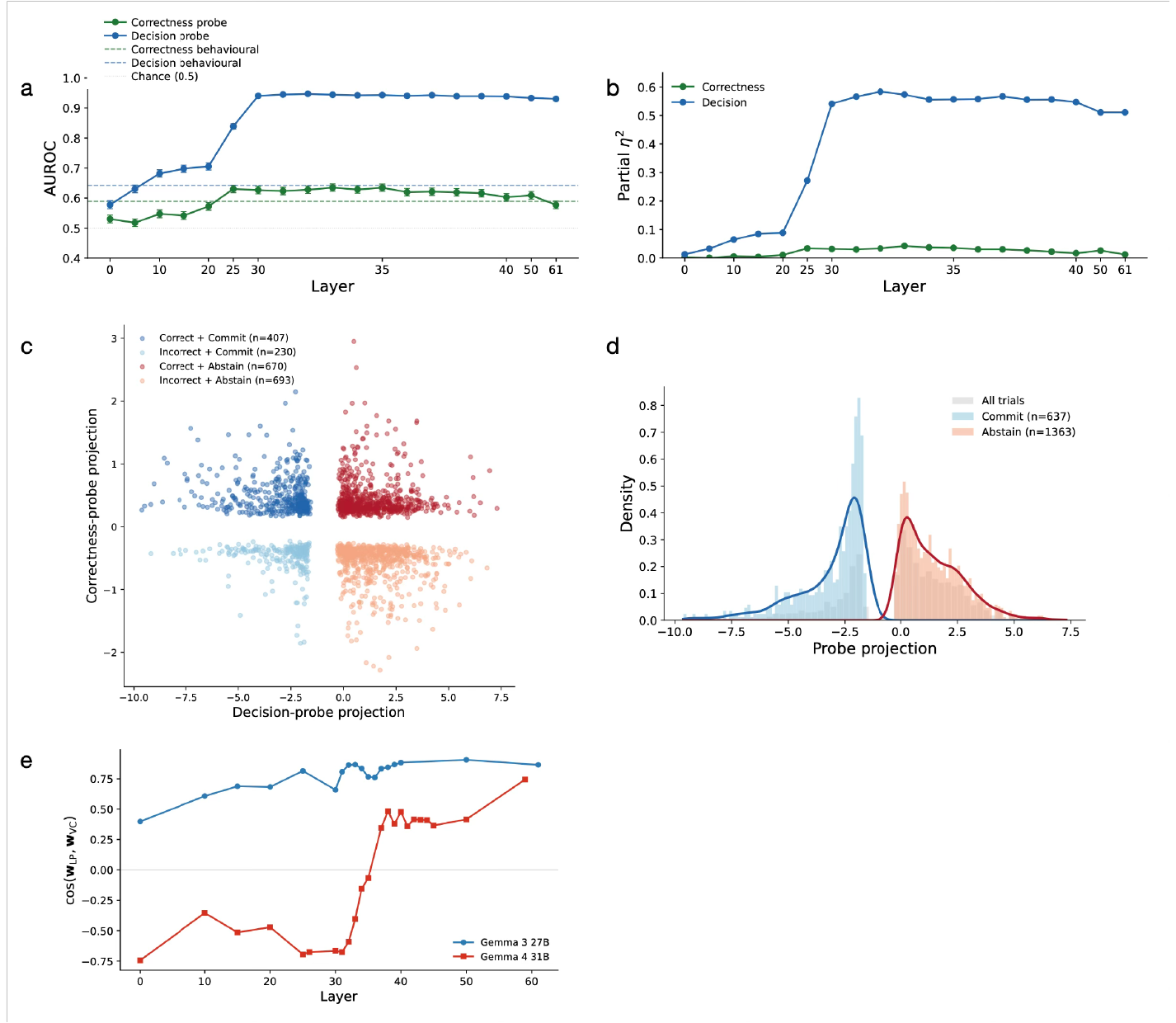}
    \caption{\textbf{Activation-level structure of decision and truth representations in Gemma~3 27B.} 
    Panels (a--d) analyse residual-stream activations at the post-answer (PA) position.
    \textbf{(a)} \textbf{Linear-probe AUROC across layers.} Decision probe (blue) and correctness probe (green), trained to predict Phase~2 abstention and Phase~1 correctness respectively. Error bars are Hanley-McNeil SEMs. Dashed lines indicate joint behavioural baselines obtained by fitting a logistic regression on Class-VC + Num-VC + Cal-LP for each target. 
    \textbf{(b)} \textbf{Partial $\eta^2$ from a 2$\times$2 ANOVA on each probe's output} (factors: correctness $\times$ decision). The decision probe's variance is dominated by the decision factor over the correctness factor (see Supplemental Results for details). 
    \textbf{(c)} \textbf{Trial-level decomposition at L32.} Each point is one of 2000 trials, plotted by its projection onto the decision-probe weight vector (x-axis) and the correctness-probe weight vector (y-axis)(see Methods). The four (correctness $\times$ decision) quadrants are populated approximately independently; clusters along the horizontal axis correspond to commit and abstain classes, separated along the direction defined by the decision probe (cosine similarity with correctness direction $= -0.01$).
    \textbf{(d)} \textbf{Distribution of the decision-probe projection at L32.} Histograms show all trials (grey) and class-conditional distributions for commit (blue) and abstain (red); coloured lines show class-conditional kernel density estimates.
    \textbf{(e)} \textbf{Layerwise alignment of Class-VC and Cal-LP steering vectors in Gemma~3 and Gemma~4.}Cosine similarity between the Class-VC and Cal-LP activation directions across layers. }
    \label{fig:activation_G3}
\end{figure}

\begin{figure}[!t]
    \centering
    \includegraphics[width=0.8\textwidth]{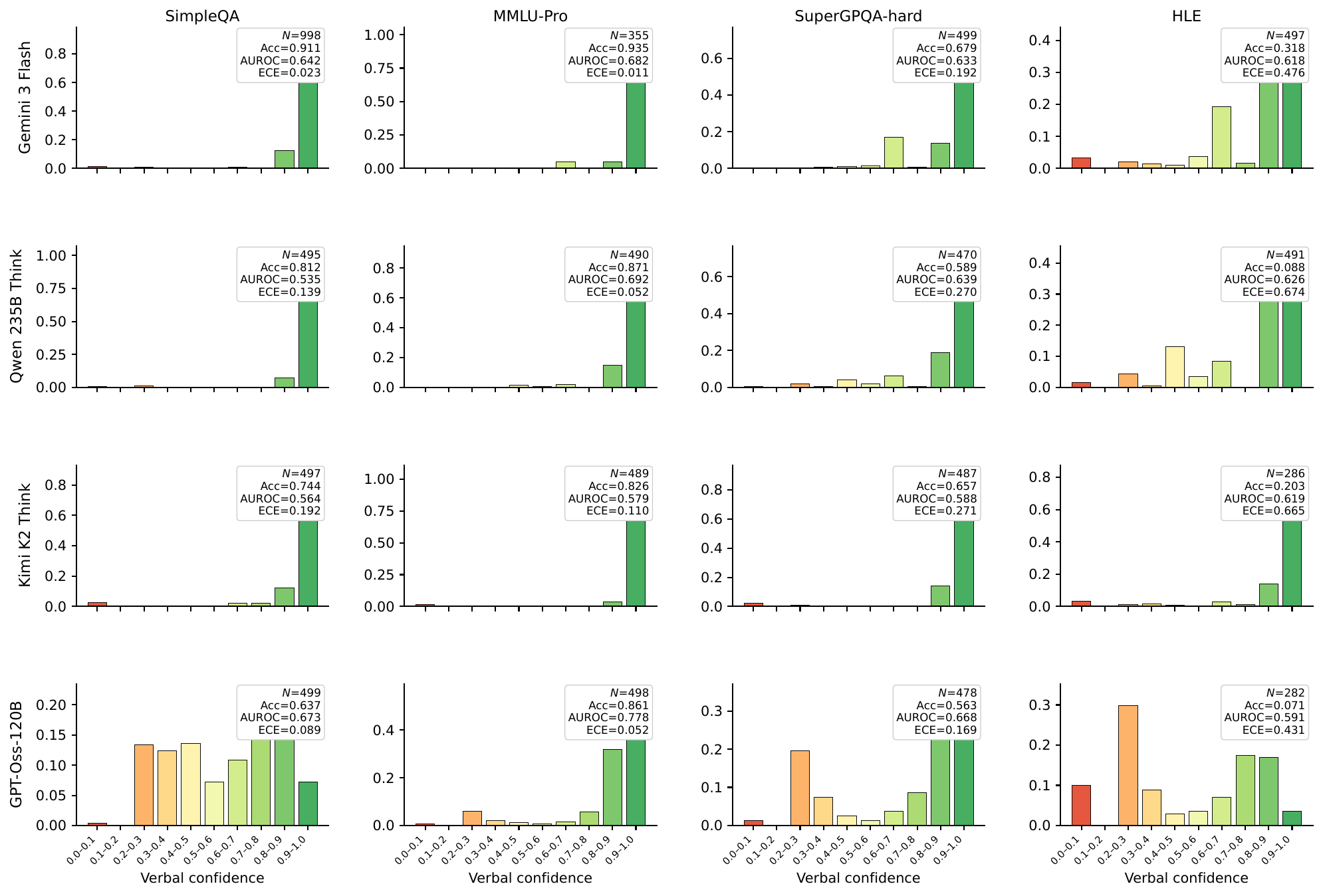}
    \caption{\textbf{Class-based verbal confidence: 4 reasoning models across 4 datasets}}
    \label{fig:all_RMmodels_vcplot}
\end{figure}

We next characterised the relationship between the two probe directions. The cosine similarity between the correctness-probe weight vector $\mathbf{w}_\mathrm{corr}$ and the decision-probe weight vector $\mathbf{w}_\mathrm{dec}$ at L32, the peak layer for predicting abstention, was $-0.025$ --- indicating that the two directions are approximately orthogonal in activation space. Visualising trials projected onto both directions confirms this: trials separate strongly along the decision axis (horizontal) while showing much weaker organisation along the correctness axis (vertical), with the four (correctness $\times$ decision) quadrants populated approximately independently (Figure~\ref{fig:activation_G3}c).

The decision projection at L32 was clearly non-unimodal (Hartigan's dip test $D = 0.051$, $p < 10^{-6}$; Figure~\ref{fig:activation}d). A two-component Gaussian mixture fit identified modes centred at $-2.67$ and $+1.31$, separated by $2.19$ pooled-SD units. The corresponding component weights (32.8\% and 67.2\%) closely tracked the behavioural split between commit and abstain trials (31.9\% and 68.2\%), and the class-conditioned projection means aligned with the same two regions of the axis ($-3.11$ for commit, $+1.45$ for abstain). The PA decision-aligned representation is therefore not merely a weak graded confidence gradient: it contains a strongly structured commit/abstain axis that is already present after answer production, before the model is ever presented with the Phase~2 abstention prompt.

%----SUPPL FIGURES 
%----------NRM verbal confidence distribution

%---------residualization analysis figure----

%-----raw confidence means to complement radar plot and eta squared table 

%%%%%%-----___RM FIGURES
%-----------RM verbal conf

%---------SUPPL TABLES
%NRM table with confidence metrics (and accuracy)
\begin{table}[t]
\centering
\small
\setlength{\tabcolsep}{4pt}
\renewcommand{\arraystretch}{1.1}
\begin{tabular}{llrrrrr}
\toprule
 &  &  & & & \multicolumn{2}{c}{Calibration} \\
\cmidrule(lr){6-7}
Dataset & Model & $N$ & Acc. & $T^{*}$ & ECE & AUROC \\
\midrule
\multicolumn{7}{l}{\textit{Verbal confidence (class-based)}} \\
\addlinespace[2pt]
SimpleQA & Gemma 3 27B          & 1999 & 0.538 & --  & 0.246 & 0.560 \\
         & Gemma 4 31B (no-think) & 1865 & 0.582 & --  & 0.351 & 0.533 \\
         & GPT-4o               & 2000 & 0.637 & --  & 0.244 & 0.587 \\
         & Qwen 3 235B (no-think) &  498 & 0.651 & --  & 0.268 & 0.557 \\
\addlinespace[2pt]
MMLU-Pro & Gemma 3 27B          &  500 & 0.292 & --  & 0.087 & 0.597 \\
         & Gemma 4 31B (no-think)  &  391 & 0.555 & --  & 0.316 & 0.637 \\
         & GPT-4o               &  500 & 0.316 & --  & 0.408 & 0.615 \\
         & Qwen 3 235B (no-think) &  496 & 0.435 & --  & 0.311 & 0.645 \\
\midrule
\multicolumn{7}{l}{\textit{Numeric confidence (0--100 scale)}} \\
\addlinespace[2pt]
SimpleQA & Gemma 3 27B          & 2000 & 0.538 & --  & 0.276 & 0.555 \\
         & Gemma 4 31B (no-think) & 1991 & 0.577 & --  & 0.421 & 0.507 \\
         & GPT-4o               & 2000 & 0.637 & --  & 0.217 & 0.606 \\
         & Qwen 3 235B (no-think) &  498 & 0.651 & --  & 0.176 & 0.629 \\
\addlinespace[2pt]
MMLU-Pro & Gemma 3 27B          &  307 & 0.293 & -- & 0.318 & 0.507 \\
         & Gemma 4 31B (no-think) &   296 & 0.600 & -- & 0.418 & 0.502 \\
         & GPT-4o               &  500 & 0.316 & --  & 0.504 & 0.552 \\
         & Qwen 3 235B (no-think) &  496 & 0.435 & --  & 0.347 & 0.675 \\
\midrule
\multicolumn{7}{l}{\textit{Calibrated log-probability (Cal-LP)}} \\
\addlinespace[2pt]
SimpleQA & Gemma 3 27B          & 2000 & 0.538 & 5.1 & 0.041 & 0.623 \\
         & Gemma 4 31B          & 2000 & 0.578 & 7.8 & 0.026 & 0.660 \\
         & GPT-4o               & 2000 & 0.637 & 4.1 & 0.019 & 0.742 \\
         & Qwen 3 235B no-think &  498 & 0.651 & 3.4 & 0.055 & 0.734 \\
\addlinespace[2pt]
MMLU-Pro & Gemma 3 27B          &  500 & 0.292 & 7.4 & 0.058 & 0.671 \\
         & Gemma 4 31B          &  500 & 0.494 & 4.5 & 0.071 & 0.800 \\
         & GPT-4o               &  500 & 0.316 & 5.3 & 0.046 & 0.728 \\
         & Qwen 3 235B no-think &  496 & 0.435 & 3.8 & 0.064 & 0.714 \\
\bottomrule
\addlinespace[6pt]
\end{tabular}
\caption{\textbf{Three confidence signals across non-reasoning models.}
For each model and dataset, $N$ is the number of valid Phase~1 trials, Acc.\ is Phase~1 accuracy on those trials, $T^{*}$ is the temperature applied to the option-token logprobs (fitted on a disjoint Phase~0 calibration set; pre-fitted values used for the original SimpleQA runs as published). ECE is 15-bin equal-width expected calibration error and AUROC measures discrimination of correct from incorrect Phase~1 answers.}
\label{tab:vc_numc_callp_calibration}
\end{table}

%---------------table of behavioral correlation Class-VC, Num-VC, Cal-LP. 
\begin{table}[t]
\centering
\small
\setlength{\tabcolsep}{6pt}
\renewcommand{\arraystretch}{1.15}
\begin{tabular}{lllll}
\toprule
 & & \multicolumn{3}{c}{Spearman $\rho$ [95\% CI]} \\
\cmidrule(lr){3-5}
Dataset & Model & Class-VC $\times$ Num-VC & Class-VC $\times$ Cal-LP & Num-VC $\times$ Cal-LP \\
\midrule
SimpleQA & Gemma 3      & 0.38 [0.33, 0.42] & 0.32 [0.28, 0.36] & 0.33 [0.29, 0.37] \\
         & Gemma 4 (NT) & 0.51 [0.44, 0.57] & 0.18 [0.14, 0.23] & 0.10 [0.06, 0.15] \\
         & GPT-4o       & 0.51 [0.47, 0.54] & 0.33 [0.29, 0.36] & 0.51 [0.48, 0.55] \\
         & Qwen 3 (NT)  & 0.36 [0.28, 0.44] & 0.24 [0.15, 0.31] & 0.51 [0.43, 0.57] \\
\addlinespace[3pt]
MMLU-Pro & Gemma 3      & 0.03 [$-$0.08, 0.12] & 0.17 [0.04, 0.27] & $-$0.11 [$-$0.23, $-$0.01] \\
         & Gemma 4 (NT)$^{\ddagger}$ & 0.09 [$-$0.06, 0.25] & 0.35 [0.24, 0.45] & 0.10 [0.01, 0.18] \\
         & GPT-4o                    & 0.39 [0.30, 0.46]    & 0.39 [0.31, 0.46] & 0.39 [0.31, 0.47] \\
         & Qwen 3 (NT)               & 0.46 [0.38, 0.53]    & 0.32 [0.24, 0.40] & 0.43 [0.35, 0.50] \\
\bottomrule
\addlinespace[6pt]
\end{tabular}
\caption{\textbf{Pairwise Spearman correlations between the three confidence signals across non-reasoning models.}
For each (dataset, model) cell, we report Spearman $\rho$ and its 95\% paired-bootstrap CI (1000 resamples) between Class-VC, Num-VC, and Cal-LP, computed on the joint sample where all three signals are valid. Gemma 4 31B and Qwen 3 235B are run in no-think mode (NT). $^{\ddagger}$Numeric confidence for Gemma~4 31B on MMLU-Pro is degenerate: 292/296 (98.6\%) of valid responses equal 100, producing near-zero variance. The two Num-VC correlations are accordingly attenuated; the Class-VC $\times$ Cal-LP correlation is unaffected.}
\label{tab:three_way_correlation}
\end{table}
%--Table with accuracy, abs rate, cond abs rates, delta incorr vs corr, AUROC DELETED
\begin{table}[t]
\centering
\small
\setlength{\tabcolsep}{5pt}
\renewcommand{\arraystretch}{1.1}
\begin{tabular}{llrrrrrrr}
\toprule
 & & & & & & \multicolumn{2}{c}{$P(\text{abstain}\mid\cdot)$} & \\
\cmidrule(lr){7-8}
Dataset & Model & $N$ & Acc.\ & Sim.\ \% & Abs.\ rate & correct & incorrect & $\Delta$ \\
\midrule
SimpleQA  & Gemma 3 27B                       & 2000 & 0.538 & 76.2 & 0.681 & 0.622 & 0.751 & $+0.129$ \\
          & Gemma 4 31B (no-think)            & 2000 & 0.578 & 80.1 & 0.153 & 0.141 & 0.169 & $+0.028$ \\
          & GPT-4o                            & 2000 & 0.637 & 78.0 & 0.722 & 0.651 & 0.846 & $+0.195$ \\
          & Qwen 3 235B (no-think)            &  500 & 0.648 & 84.1 & 0.090 & 0.068 & 0.131 & $+0.063$ \\
\addlinespace[3pt]
MMLU-Pro  & Gemma 3 27B                       &  500 & 0.292 & --   & 0.830 & 0.781 & 0.850 & $+0.069$ \\
          & Gemma 4 31B (no-think)            &  500 & 0.494 & --   & 0.064 & 0.024 & 0.103 & $+0.078$ \\
          & GPT-4o                            &  500 & 0.316 & --   & 0.732 & 0.608 & 0.789 & $+0.182$ \\
          & Qwen 3 235B (no-think)            &  500 & 0.432 & --   & 0.308 & 0.181 & 0.405 & $+0.224$ \\
\bottomrule
\addlinespace[6pt]
\end{tabular}
\caption{\textbf{Phase~1 accuracy and Phase~2 abstention behaviour across non-reasoning models.}
$N$ is the number of Phase~1 trials. Acc.\ is overall Phase~1 accuracy. Sim.\ \% (SimpleQA only) is the proportion of Phase~1 errors that selected the semantically similar foil rather than a dissimilar one. Abs.\ rate is the overall Phase~2 abstention rate; the next two columns give abstention rates conditional on Phase~1 correctness, and $\Delta = P(\text{abstain}\mid\text{incorrect}) - P(\text{abstain}\mid\text{correct})$. All four models abstain more often after incorrect responses ($\Delta > 0$ in every cell), but the magnitude of this truth-aligned gap varies substantially across cells.}
\label{tab:nrm_behavior}
\end{table}

%---NRM LOGISTIC REG -- FULL DECOMP-------
%---NRM LOGISTIC REG -- FULL DECOMP (STANDARD PROMPT) -------
\begin{table}[t]
\centering
\footnotesize
\setlength{\tabcolsep}{3pt}
\renewcommand{\arraystretch}{1.1}
\begin{tabular}{lllrrrrrrrr}
\toprule
 & & & \multicolumn{3}{c}{AUROC [95\% CI]} & \multicolumn{4}{c}{\% unique variance} & \\
\cmidrule(lr){4-6} \cmidrule(lr){7-10}
Dataset & Model & Outcome & Class-VC & Num-VC & Cal-LP & Class-VC & Num-VC & Cal-LP & shared & McF $R^2$ \\
\midrule
SimpleQA & Gemma 3 27B  & corr & 0.56 [0.53,0.58] & 0.55 [0.52,0.57] & 0.62 [0.60,0.65] &  4.3 &  2.6 & 72.4 & 20.7 & 0.035 \\
         &              & abs  & 0.79 [0.76,0.81] & 0.78 [0.76,0.80] & 0.77 [0.74,0.79] & 28.8 & 10.1 & 10.5 & 50.6 & 0.389 \\
\addlinespace[2pt]
         & Gemma 4 31B  & corr & 0.52 [0.50,0.55] & 0.51 [0.48,0.53] & 0.66 [0.64,0.69] &  3.2 &  0.2 & 86.7 &  9.9 & 0.063 \\
         &              & abs  & 0.80 [0.77,0.82] & 0.60 [0.58,0.63] & 0.62 [0.59,0.66] & 59.1 &  3.1 &  0.8 & 37.0 & 0.409 \\
\addlinespace[2pt]
         & GPT-4o       & corr & 0.58 [0.55,0.60] & 0.60 [0.57,0.62] & 0.74 [0.72,0.76] &  0.2 &  0.2 & 81.8 & 17.8 & 0.137 \\
         &              & abs  & 0.71 [0.69,0.73] & 0.84 [0.83,0.86] & 0.84 [0.83,0.86] &  6.7 & 16.4 & 21.3 & 55.6 & 0.438 \\
\addlinespace[2pt]
         & Qwen 3 235B  & corr & 0.55 [0.49,0.60] & 0.62 [0.56,0.67] & 0.73 [0.69,0.78] &  2.0 &  0.2 & 79.3 & 18.5 & 0.128 \\
         &              & abs  & 0.65 [0.57,0.74] & 0.88 [0.81,0.94] & 0.67 [0.59,0.75] &  1.2 & 63.6 &  0.5 & 34.7 & 0.397 \\
\midrule
MMLU-Pro & Gemma 3 27B  & corr & 0.61 [0.55,0.67] & 0.46 [0.40,0.52] & 0.66 [0.61,0.72] & 12.7 &  1.2 & 58.1 & 28.0 & 0.093 \\
         &              & abs  & 0.72 [0.66,0.78] & 0.46 [0.40,0.52] & 0.69 [0.64,0.75] & 35.3 &  3.2 & 22.1 & 39.3 & 0.163 \\
\addlinespace[2pt]
         & Gemma 4 31B  & corr & 0.61 [0.55,0.68] & 0.49 [0.42,0.55] & 0.79 [0.74,0.84] &  2.3 &  0.9 & 82.3 & 14.5 & 0.218 \\
         &              & abs  & 0.82 [0.70,0.93] & 0.49 [0.49,0.50] & 0.82 [0.75,0.88] & 33.2 &  2.2 & 24.1 & 40.4 & 0.224 \\
\addlinespace[2pt]
         & GPT-4o       & corr & 0.62 [0.57,0.66] & 0.53 [0.48,0.58] & 0.73 [0.68,0.77] &  1.8 &  1.0 & 78.0 & 19.2 & 0.133 \\
         &              & abs  & 0.76 [0.71,0.80] & 0.75 [0.71,0.80] & 0.74 [0.69,0.79] & 23.6 & 13.6 & 11.1 & 51.7 & 0.284 \\
\addlinespace[2pt]
         & Qwen 3 235B  & corr & 0.64 [0.59,0.68] & 0.66 [0.62,0.71] & 0.71 [0.67,0.76] & 11.1 &  2.6 & 42.2 & 44.1 & 0.154 \\
         &              & abs  & 0.69 [0.64,0.75] & 0.81 [0.77,0.85] & 0.67 [0.62,0.72] &  5.4 & 41.5 &  3.4 & 49.7 & 0.281 \\
\bottomrule
\addlinespace[6pt]
\end{tabular}
\caption{\textbf{Full decomposition of confidence-signal contributions to predicting correctness and abstention under the standard Phase~2 prompt, across non-reasoning models.}
For each (dataset, model) cell, two rows: predicting Phase~1 correctness (corr) and Phase~2 abstention (abs). AUROCs and 95\% CIs from paired non-parametric bootstrap (1000 resamples). Decision-truth gap is reported separately in Table~\ref{tab:nrm_decision_truth_gap}. Percentage-unique-variance figures are McFadden $R^2$ decomposition components from the joint M$_{\text{VC+NumC+LP}}$ logit, expressed as a fraction of total $R^2$; shared variance pools all pairwise and three-way overlaps. Total McFadden $R^2$ shown in the last column for transparency, but is not directly comparable across outcomes because it depends on outcome base rate. AUROCs use single-signal valid samples; variance partitioning uses the joint sample where all three signals are valid. Gemma 4 31B and Qwen 3 235B are run in no-think mode.}
\label{tab:nrm_auroc_variance}
\end{table}
%-------------NRM D-T GAPS----
\begin{table}[t]
\centering
\small
\setlength{\tabcolsep}{4pt}
\renewcommand{\arraystretch}{1.15}
\begin{tabular}{llrlll}
\toprule
 & & & \multicolumn{3}{c}{Decision-truth gap [95\% CI]} \\
\cmidrule(lr){4-6}
Dataset & Model & $N$ & Class-VC & Num-VC & Cal-LP \\
\midrule
SimpleQA & Gemma 3 27B & 1999 & $+0.23$ $[+0.19,+0.26]$ & $+0.23$ $[+0.20,+0.26]$ & $+0.14$ $[+0.11,+0.17]$ \\
         & Gemma 4 31B & 1865 & $+0.26$ $[+0.23,+0.29]$ & $+0.10$ $[+0.08,+0.12]$ & $-0.04$ $[-0.07,+0.01]$ \\
         & GPT-4o      & 2000 & $+0.12$ $[+0.09,+0.15]$ & $+0.24$ $[+0.21,+0.27]$ & $+0.10$ $[+0.08,+0.13]$ \\
         & Qwen 3 235B & 498 & $+0.10$ $[+0.00,+0.19]$ & $+0.25$ $[+0.17,+0.32]$ & $-0.06$ $[-0.14,+0.02]$ \\
\addlinespace[3pt]
MMLU-Pro & Gemma 3 27B & 307 & $+0.12$ $[+0.04,+0.20]$ & $-0.05$ $[-0.12,+0.02]$ & $+0.02$ $[-0.06,+0.10]$ \\
         & Gemma 4 31B & 253 & $+0.18$ $[+0.07,+0.29]$ & $-0.01$ $[-0.03,+0.00]$ & $+0.02$ $[-0.06,+0.10]$ \\
         & GPT-4o      & 500 & $+0.14$ $[+0.09,+0.20]$ & $+0.20$ $[+0.15,+0.26]$ & $+0.02$ $[-0.05,+0.09]$ \\
         & Qwen 3 235B & 496 & $+0.05$ $[-0.02,+0.11]$ & $+0.14$ $[+0.09,+0.20]$ & $-0.05$ $[-0.11,+0.02]$ \\
\bottomrule
\addlinespace[6pt]
\end{tabular}
\caption{\textbf{Decision-truth gaps for each confidence signal across non-reasoning models.}
For each (dataset, model) cell, the decision-truth gap is $\text{AUROC}_{\text{abstention}} - \text{AUROC}_{\text{correctness}}$, computed on the same trials with paired-bootstrap 95\% CIs (1000 resamples; resampled trial indices used for both AUROCs in each iteration). $N$ is the number of trials on which all three signals (Class-VC, Num-VC, Cal-LP) are jointly valid; AUROCs use single-signal valid samples (per-signal $N$ may be slightly higher and is reported in Table~\ref{tab:nrm_behavior}). Positive values indicate decision-aligned signal (better predicts abstention than correctness); near-zero or negative values indicate truth-aligned signal. Gemma 4 31B and Qwen 3 235B are run in no-think mode. Verbal measures (Class-VC, Num-VC) consistently show positive gaps; calibrated log-probability (Cal-LP) shows small or negative gaps.}
\label{tab:nrm_decision_truth_gap}
\end{table}

%-----------VC predicting ABS with residuals -- reciprocal results
\begin{table}[t]
\centering
\scriptsize
\setlength{\tabcolsep}{3.5pt}
\renewcommand{\arraystretch}{1.08}

\begin{tabular}{lllrrrrrrr}
\toprule
\multicolumn{10}{l}{\textbf{A. VC residualised on Cal-LP}} \\
\midrule
Dataset & Model & Signal
& Corr. raw & Corr. resid.
& Abst. raw & Abst. resid.
& Resid. $\Delta_{\mathrm{DT}}$ [95\% CI]
& \% abst. pres.
& VC-Cal-LP shared variance ($R^2$) \\
\midrule
\multicolumn{10}{l}{\textit{SimpleQA}} \\
\addlinespace[2pt]
 & Gemma 3        & Class-VC & 0.560 & 0.502 & 0.786 & 0.731 & +0.229 [+0.194, +0.261] & 81.0 & 0.155 \\
 &                & Num-VC   & 0.555 & 0.506 & 0.780 & 0.687 & +0.180 [+0.148, +0.213] & 66.5 & 0.086 \\
\addlinespace[1pt]
 & Gemma 4        & Class-VC & 0.533 & 0.410 & 0.796 & 0.768 & +0.357 [+0.312, +0.402] & 90.5 & 0.036 \\
 &                & Num-VC   & 0.508 & 0.358 & 0.608 & 0.511 & +0.153 [+0.108, +0.203] & 9.9  & 0.009 \\
\addlinespace[1pt]
 & GPT-4o         & Class-VC & 0.587 & 0.472 & 0.707 & 0.536 & +0.064 [+0.031, +0.097] & 17.2 & 0.139 \\
 &                & Num-VC   & 0.606 & 0.474 & 0.842 & 0.679 & +0.205 [+0.169, +0.237] & 52.3 & 0.163 \\
\addlinespace[1pt]
 & Qwen 3 (NT)    & Class-VC & 0.557 & 0.424 & 0.653 & 0.571 & +0.148 [+0.036, +0.267] & 46.7 & 0.018 \\
 &                & Num-VC   & 0.629 & 0.494 & 0.879 & 0.849 & +0.355 [+0.261, +0.446] & 91.9 & 0.142 \\
\midrule
\multicolumn{10}{l}{\textit{MMLU-Pro}} \\
\addlinespace[2pt]
 & Gemma 3        & Class-VC & 0.629 & 0.575 & 0.694 & 0.649 & +0.074 [$-$0.020, +0.182] & 76.9 & 0.110 \\
 &                & Num-VC   & 0.507 & 0.617 & 0.455 & 0.604 & $-$0.013 [$-$0.136, +0.079] & --   & 0.007 \\
\addlinespace[1pt]
 & Gemma 4        & Class-VC & 0.626 & 0.402 & 0.781 & 0.639 & +0.237 [$-$0.069, +0.527] & 49.5 & 0.112 \\
 &                & Num-VC   & 0.504 & 0.215 & 0.492 & 0.177 & $-$0.038 [$-$0.172, +0.109] & --   & 0.005 \\
\addlinespace[1pt]
 & GPT-4o         & Class-VC & 0.615 & 0.505 & 0.759 & 0.664 & +0.159 [+0.087, +0.219] & 63.4 & 0.116 \\
 &                & Num-VC   & 0.552 & 0.441 & 0.753 & 0.669 & +0.228 [+0.158, +0.306] & 66.8 & 0.115 \\
\addlinespace[1pt]
 & Qwen 3 (NT)    & Class-VC & 0.645 & 0.525 & 0.691 & 0.633 & +0.108 [+0.039, +0.176] & 69.5 & 0.084 \\
 &                & Num-VC   & 0.675 & 0.586 & 0.814 & 0.768 & +0.182 [+0.119, +0.248] & 85.5 & 0.123 \\

\addlinespace[8pt]
\toprule
\multicolumn{10}{l}{\textbf{B. Cal-LP residualised on verbal confidence}} \\
\midrule
Dataset & Model & Residualised on
& Corr. raw & Corr. resid.
& Abst. raw & Abst. resid.
& Resid. $\Delta_{\mathrm{DT}}$ [95\% CI]
& \% corr. pres.
& \% abst. pres. \\
\midrule
\multicolumn{10}{l}{\textit{SimpleQA}} \\
\addlinespace[2pt]
 & Gemma 3        & Class-VC & 0.623 & 0.607 & 0.765 & 0.631 & +0.024 [$-$0.006, +0.056] & 87.0 & 49.5 \\
 &                & Num-VC   & 0.623 & 0.621 & 0.765 & 0.709 & +0.088 [+0.058, +0.118] & 98.9 & 78.9 \\
\addlinespace[1pt]
 & Gemma 4        & Class-VC & 0.663 & 0.654 & 0.635 & 0.528 & $-$0.126 [$-$0.163, $-$0.090] & 94.4 & 20.7 \\
 &                & Num-VC   & 0.663 & 0.661 & 0.635 & 0.601 & $-$0.059 [$-$0.103, $-$0.019] & 98.6 & 75.0 \\
\addlinespace[1pt]
 & GPT-4o         & Class-VC & 0.742 & 0.719 & 0.845 & 0.776 & +0.057 [+0.026, +0.087] & 90.5 & 80.1 \\
 &                & Num-VC   & 0.742 & 0.728 & 0.845 & 0.767 & +0.039 [+0.008, +0.070] & 94.1 & 77.4 \\
\addlinespace[1pt]
 & Qwen 3 (NT)    & Class-VC & 0.734 & 0.727 & 0.673 & 0.635 & $-$0.092 [$-$0.174, $-$0.017] & 97.1 & 78.0 \\
 &                & Num-VC   & 0.734 & 0.710 & 0.673 & 0.473 & $-$0.236 [$-$0.317, $-$0.157] & 89.5 & $-$15.5 \\
\midrule
\multicolumn{10}{l}{\textit{MMLU-Pro}} \\
\addlinespace[2pt]
 & Gemma 3        & Class-VC & 0.672 & 0.644 & 0.720 & 0.663 & +0.019 [$-$0.083, +0.122] & 84.0 & 74.0 \\
 &                & Num-VC   & 0.672 & 0.673 & 0.720 & 0.717 & +0.044 [$-$0.059, +0.142] & 100.6 & 98.5 \\
\addlinespace[1pt]
 & Gemma 4        & Class-VC & 0.794 & 0.766 & 0.819 & 0.713 & $-$0.054 [$-$0.258, +0.111] & 90.5 & 66.7 \\
 &                & Num-VC   & 0.794 & 0.795 & 0.819 & 0.820 & +0.025 [$-$0.123, +0.160] & 100.3 & 100.4 \\
\addlinespace[1pt]
 & GPT-4o         & Class-VC & 0.728 & 0.698 & 0.743 & 0.628 & $-$0.069 [$-$0.142, +0.006] & 86.8 & 52.9 \\
 &                & Num-VC   & 0.728 & 0.725 & 0.743 & 0.647 & $-$0.078 [$-$0.162, +0.004] & 98.6 & 60.5 \\
\addlinespace[1pt]
 & Qwen 3 (NT)    & Class-VC & 0.714 & 0.673 & 0.667 & 0.586 & $-$0.087 [$-$0.144, $-$0.026] & 80.6 & 51.5 \\
 &                & Num-VC   & 0.714 & 0.675 & 0.667 & 0.552 & $-$0.123 [$-$0.187, $-$0.066] & 81.6 & 30.8 \\
\bottomrule
\end{tabular}

\caption{\textbf{Reciprocal residual analyses dissociate the VC-specific and Cal-LP-specific components.}
\textbf{(A)} Verbal confidence residualised on Cal-LP. For each non-reasoning model cell, we remove the linear component of Class-VC or Num-VC shared with Cal-LP and evaluate the residual signal as a predictor of Phase~1 correctness and Phase~2 abstention. Residual $\Delta_{\mathrm{DT}} = \mathrm{AUROC}_{\mathrm{abstention}} - \mathrm{AUROC}_{\mathrm{correctness}}$; 95\% CIs are paired-bootstrap intervals over trials. \% abst.\ pres.\ is the fraction of raw abstention-predictive power preserved after residualisation, relative to a chance baseline of 0.5. VC-Cal-LP shared variance ($R^2$): the coefficient of determination from the OLS regression of VC on Cal-LP at the trial level, indicating the fraction of VC's trial-to-trial variance that is linearly explained by Cal-LP. This quantifies the overall shared variance between the two signals, independent of correctness or abstention targets. Higher values indicate greater overlap between VC and Cal-LP; lower values indicate that VC and Cal-LP capture more largely distinct trial-level information. \textbf{(B)} Cal-LP residualised on verbal confidence. We perform the converse analysis, removing the linear component of Cal-LP shared with Class-VC or Num-VC and evaluating the residual Cal-LP signal against the same two targets. \% corr.\ pres.\ and \% abst.\ pres.\ are calculated relative to the corresponding raw AUROC above chance. Positive residual $\Delta_{\mathrm{DT}}$ indicates stronger behavioural than truth alignment; negative values indicate stronger truth than behavioural alignment. Gemma~4 31B and Qwen~3 235B are run in no-think mode. Dashes indicate cases where preserved-fraction estimates are uninterpretable because the raw AUROC is at or below chance.}
\label{tab:reciprocal_residual_analysis}
\end{table}
%-------original residual analysis with difficult/1-0 correctness

\begin{table}[t]
\centering
\small
\setlength{\tabcolsep}{4pt}
\renewcommand{\arraystretch}{1.1}
\begin{tabular}{llrrrrrr}
\toprule
 & & & \multicolumn{2}{c}{AUROC} & & \multicolumn{2}{c}{Δ} \\
\cmidrule(lr){4-5} \cmidrule(lr){7-8}
Model & Signal & Spec & Raw & Residual & \% pres. & delta AUROC & 95\% CI \\
\midrule
\multicolumn{8}{l}{\textit{SimpleQA}} \\
\addlinespace[2pt]
Gemma 3        & Class-VC & corr only         & 0.79 & 0.77 & 95.4 & 0.01 & [0.01, 0.02] \\
               &          & + Cal-LP          & 0.79 & 0.73 & 80.6 & 0.06 & [0.04, 0.07] \\
               &          & + Cal-LP + diff.  & 0.79 & 0.73 & 80.3 & 0.06 & [0.04, 0.07] \\
\addlinespace[1pt]
               & Num-VC   & corr only         & 0.78 & 0.77 & 97.1 & 0.01 & [0.00, 0.01] \\
               &          & + Cal-LP          & 0.78 & 0.69 & 67.0 & 0.09 & [0.07, 0.11] \\
               &          & + Cal-LP + diff.  & 0.78 & 0.69 & 67.4 & 0.09 & [0.07, 0.11] \\
\addlinespace[2pt]
Gemma 4        & Class-VC & corr only         & 0.80 & 0.80 & 102.1 & −0.01 & [−0.02, 0.01] \\
               &          & + Cal-LP          & 0.80 & 0.77 & 91.8 & 0.02 & [0.00, 0.05] \\
               &          & + Cal-LP + diff.  & 0.80 & 0.77 & 90.4 & 0.03 & [0.00, 0.05] \\
\addlinespace[1pt]
               & Num-VC   & corr only         & 0.61 & 0.59 & 82.0 & 0.02 & [−0.01, 0.05] \\
               &          & + Cal-LP          & 0.61 & 0.51 & 12.1 & 0.10 & [0.06, 0.12] \\
               &          & + Cal-LP + diff.  & 0.61 & 0.51 & 6.8  & 0.10 & [0.05, 0.13] \\
\addlinespace[2pt]
GPT-4o         & Class-VC & corr only         & 0.71 & 0.66 & 74.7 & 0.05 & [0.03, 0.07] \\
               &          & + Cal-LP          & 0.71 & 0.54 & 17.3 & 0.17 & [0.16, 0.19] \\
               &          & + Cal-LP + diff.  & 0.71 & 0.54 & 17.2 & 0.17 & [0.15, 0.19] \\
\addlinespace[1pt]
               & Num-VC   & corr only         & 0.84 & 0.83 & 95.0 & 0.02 & [0.01, 0.02] \\
               &          & + Cal-LP          & 0.84 & 0.68 & 52.3 & 0.16 & [0.14, 0.19] \\
               &          & + Cal-LP + diff.  & 0.84 & 0.68 & 52.3 & 0.16 & [0.14, 0.20] \\
\addlinespace[2pt]
Qwen 3 (NT)    & Class-VC & corr only         & 0.65 & 0.62 & 76.0 & 0.04 & [−0.01, 0.08] \\
               &          & + Cal-LP          & 0.65 & 0.58 & 49.1 & 0.08 & [0.03, 0.12] \\
               &          & + Cal-LP + diff.  & 0.65 & 0.58 & 49.5 & 0.08 & [0.02, 0.12] \\
\addlinespace[1pt]
               & Num-VC   & corr only         & 0.88 & 0.87 & 98.6 & 0.01 & [−0.01, 0.02] \\
               &          & + Cal-LP          & 0.88 & 0.85 & 92.0 & 0.03 & [0.01, 0.06] \\
               &          & + Cal-LP + diff.  & 0.88 & 0.85 & 92.1 & 0.03 & [0.01, 0.06] \\
\midrule
\multicolumn{8}{l}{\textit{MMLU-Pro}} \\
\addlinespace[2pt]
Gemma 3        & Class-VC & corr only         & 0.69 & 0.69 & 98.8 & 0.00 & [−0.02, 0.05] \\
               &          & + Cal-LP          & 0.69 & 0.65 & 76.1 & 0.05 & [0.01, 0.09] \\
\addlinespace[1pt]
               & Num-VC   & corr only         & 0.46$^{\ddagger}$ & 0.38 & --   & 0.07 & [−0.13, 0.12] \\
               &          & + Cal-LP          & 0.46$^{\ddagger}$ & 0.57 & --   & −0.12 & [−0.20, 0.14] \\
\addlinespace[2pt]
Gemma 4        & Class-VC & corr only         & 0.78 & 0.77 & 98.1 & 0.01 & [−0.03, 0.01] \\
               &          & + Cal-LP          & 0.78 & 0.68 & 63.8 & 0.10 & [−0.00, 0.28] \\
\addlinespace[1pt]
               & Num-VC   & corr only         & 0.49$^{\ddagger}$ & 0.51 & --   & −0.02 & [−0.21, 0.21] \\
               &          & + Cal-LP          & 0.49$^{\ddagger}$ & 0.12 & --   & 0.37 & [−0.27, 0.42] \\
\addlinespace[2pt]
GPT-4o         & Class-VC & corr only         & 0.76 & 0.74 & 94.3 & 0.02 & [0.00, 0.06] \\
               &          & + Cal-LP          & 0.76 & 0.66 & 63.2 & 0.10 & [0.06, 0.15] \\
\addlinespace[1pt]
               & Num-VC   & corr only         & 0.75 & 0.73 & 90.6 & 0.02 & [0.01, 0.04] \\
               &          & + Cal-LP          & 0.75 & 0.67 & 67.8 & 0.08 & [0.04, 0.13] \\
\addlinespace[2pt]
Qwen 3 (NT)    & Class-VC & corr only         & 0.69 & 0.64 & 75.2 & 0.05 & [0.02, 0.08] \\
               &          & + Cal-LP          & 0.69 & 0.62 & 64.6 & 0.07 & [0.04, 0.10] \\
\addlinespace[1pt]
               & Num-VC   & corr only         & 0.81 & 0.79 & 93.0 & 0.02 & [0.01, 0.07] \\
               &          & + Cal-LP          & 0.81 & 0.75 & 79.0 & 0.07 & [0.04, 0.10] \\
\bottomrule
\addlinespace[6pt]
\end{tabular}
\caption{\textbf{Residual abstention-prediction analysis after controlling for correctness-related proxies.}
For each (model, signal) cell, we residualise verbal confidence on increasingly stringent proxies for correctness-related answer evidence and measure the abstention AUROC of the residual. Three nested specifications are tested: \emph{corr only} (binary $is\_correct$); \emph{$+$ Cal-LP} (continuous calibrated log-probability confidence); and \emph{$+$ Cal-LP $+$ difficulty} (additional question-level difficulty proxy, SimpleQA only). \% pres.\ = $(\text{AUROC}_{\text{residual}} - 0.5) / (\text{AUROC}_{\text{raw}} - 0.5) \times 100$. $\Delta$ AUROC and 95\% CI are obtained by paired bootstrap with residualisation refit in each of 1000 resamples. Gemma~4 31B and Qwen~3 235B are run in no-think mode. $^{\ddagger}$Raw AUROC near or below chance: \% preserved is uninterpretable.}
\label{tab:residual_correctness_proxy_analysis}
\end{table}

%------------eta squared for Radar plots
\begin{table}[t]
\centering
\small
\setlength{\tabcolsep}{6pt}
\renewcommand{\arraystretch}{1.15}
\begin{tabular}{llrrr}
\toprule
 & & & \multicolumn{2}{c}{Partial $\eta^{2}$} \\
\cmidrule(lr){4-5}
Dataset & Model & Signal & Correctness & Decision \\
\midrule
\multicolumn{5}{l}{\textit{SimpleQA}} \\
\addlinespace[2pt]
SimpleQA & Gemma 3       & Class-VC & 0.002 & 0.351 \\
         &               & Num-VC   & 0.000 & 0.142 \\
         &               & Cal-LP   & 0.029 & 0.167 \\
\addlinespace[1pt]
         & Gemma 4 (NT)  & Class-VC & 0.011 & 0.484 \\
         &               & Num-VC   & 0.001 & 0.173 \\
         &               & Cal-LP   & 0.076 & 0.022 \\
\addlinespace[1pt]
         & GPT-4o        & Class-VC & 0.009 & 0.133 \\
         &               & Num-VC   & 0.005 & 0.170 \\
         &               & Cal-LP   & 0.126 & 0.271 \\
\addlinespace[1pt]
         & Qwen 3 (NT)   & Class-VC & 0.008 & 0.109 \\
         &               & Num-VC   & 0.022 & 0.332 \\
         &               & Cal-LP   & 0.144 & 0.019 \\
\midrule
\multicolumn{5}{l}{\textit{MMLU-Pro}} \\
\addlinespace[2pt]
MMLU-Pro & Gemma 3       & Class-VC & 0.027 & 0.137 \\
         &               & Num-VC   & 0.001 & 0.006 \\
         &               & Cal-LP   & 0.094 & 0.063 \\
\addlinespace[1pt]
         & Gemma 4 (NT)  & Class-VC & 0.049$^{\dagger}$ & 0.132$^{\dagger}$ \\
         &               & Num-VC   & $<$0.001$^{\dagger\ddagger}$ & $<$0.001$^{\dagger\ddagger}$ \\
         &               & Cal-LP   & 0.250$^{\dagger}$ & 0.059$^{\dagger}$ \\
\addlinespace[1pt]
         & GPT-4o        & Class-VC & 0.012 & 0.144 \\
         &               & Num-VC   & 0.003 & 0.112 \\
         &               & Cal-LP   & 0.141 & 0.162 \\
\addlinespace[1pt]
         & Qwen 3 (NT)   & Class-VC & 0.050 & 0.141 \\
         &               & Num-VC   & 0.034 & 0.256 \\
         &               & Cal-LP   & 0.122 & 0.040 \\
\bottomrule
\addlinespace[6pt]
\end{tabular}
\caption{\textbf{Effect sizes from a 2$\times$2 (correctness $\times$ decision) ANOVA on each confidence signal across non-reasoning models.}
For each (dataset, model, signal) cell, partial $\eta^{2}$ quantifies the proportion of variance in confidence attributable to a main effect, after partialling out the other main effect and the interaction: $\eta^{2}_{\text{partial}} = SS_{\text{effect}} / (SS_{\text{effect}} + SS_{\text{residual}})$. Values range 0 to 1; conventional thresholds are 0.01 (small), 0.06 (medium), 0.14 (large). The Correctness column shows the partial $\eta^{2}$ for the correctness main effect (correct vs.\ incorrect); the Decision column shows the partial $\eta^{2}$ for the abstention decision main effect (commit vs.\ abstain). Larger Decision $\eta^{2}$ relative to Correctness $\eta^{2}$ indicates that the signal differs more across abstention decisions than across actual correctness --- the signature of a decision-aligned signal.
$^{\dagger}$Underpowered: only 4--14 trials in the correct--abstain or incorrect--abstain quadrant of the 2$\times$2.
$^{\ddagger}$Numeric confidence for Gemma~4 31B on MMLU-Pro is degenerate: 292/296 (98.6\%) of valid responses equal 100, leaving near-zero variance for either main effect to explain (raw $\eta^{2}$ values $\sim 10^{-4}$).}
\label{tab:eta2_effect_sizes}
\end{table}

%_-----------_TABLES FOR NEUTRAL PROMPT NRM (with neutral AUROC only)
\begin{table}[t]
\centering
\footnotesize
\setlength{\tabcolsep}{5pt}
\renewcommand{\arraystretch}{1.1}
\begin{tabular}{llcccccccc}
\toprule
 &  & \multicolumn{4}{c}{Standard prompt} & \multicolumn{4}{c}{Neutral prompt} \\
\cmidrule(lr){3-6} \cmidrule(lr){7-10}
Dataset & Model & AbsR & P\textsubscript{cor} & P\textsubscript{inc} & $\Delta$ & AbsR & P\textsubscript{cor} & P\textsubscript{inc} & $\Delta$ \\
\midrule
SimpleQA & Gemma 3 27B          & 0.681 & 0.622 & 0.751 & +0.129 & 0.464 & 0.399 & 0.540 & +0.140 \\
         & Gemma 4 31B          & 0.153 & 0.141 & 0.169 & +0.028 & 0.033 & 0.021 & 0.050 & +0.029 \\
         & GPT-4o               & 0.722 & 0.651 & 0.846 & +0.195 & 0.346 & 0.283 & 0.456 & +0.173 \\
         & Qwen 3 235B no-think & 0.090 & 0.068 & 0.131 & +0.063 & 0.104 & 0.077 & 0.153 & +0.076 \\
\midrule
MMLU-Pro & Gemma 3 27B          & 0.830 & 0.781 & 0.850 & +0.069 & 0.750 & 0.747 & 0.751 & +0.005 \\
         & Gemma 4 31B          & 0.064 & 0.024 & 0.103 & +0.078 & 0.040 & 0.016 & 0.063 & +0.047 \\
         & GPT-4o               & 0.732 & 0.608 & 0.789 & +0.182 & 0.470 & 0.342 & 0.529 & +0.187 \\
         & Qwen 3 235B no-think & 0.308 & 0.181 & 0.405 & +0.224 & 0.324 & 0.190 & 0.426 & +0.236 \\
\bottomrule
\addlinespace[6pt]
\end{tabular}
\caption{\textbf{Phase 2 abstention behaviour under standard and neutral prompts, across non-reasoning models.}
For each (dataset, model) cell, abstention rates are reported under the standard prompt (caution clause: ``some questions have NO correct answer; if you don't see a clearly correct answer, choose B'') and the neutral prompt (no caution clause; see Methods). \emph{AbsR} is the overall Phase 2 abstention rate. P\textsubscript{cor} and P\textsubscript{inc} are abstention rates conditional on Phase 1 correctness. $\Delta = $ P\textsubscript{inc} $-$ P\textsubscript{cor} is an index of the discrimination of the abstention decision. All sixteen cells (8 model$\times$dataset$\times$prompt combinations) show $\Delta > 0$, indicating that abstention is partially discriminatory of correctness regardless of prompt framing. Base abstention rates and the magnitude of the discriminatory gap vary across models and prompt variants. Gemma 4 31B and Qwen 3 235B are run in no-think mode.}
\label{tab:nrm_abstention_by_prompt}
\end{table}
%_-----------_NEUTRAL PROMPT LOGISTIC REGRESSION
%---NRM LOGISTIC REG -- FULL DECOMP (NEUTRAL PROMPT) -------

%---NRM LOGISTIC REG -- FULL DECOMP (NEUTRAL PROMPT) -------
\begin{table}[t]
\centering
\footnotesize
\setlength{\tabcolsep}{3pt}
\renewcommand{\arraystretch}{1.1}
\begin{tabular}{lllrrrrrrrr}
\toprule
 & & & \multicolumn{3}{c}{AUROC [95\% CI]} & \multicolumn{4}{c}{\% unique variance} & \\
\cmidrule(lr){4-6} \cmidrule(lr){7-10}
Dataset & Model & Outcome & Class-VC & Num-VC & Cal-LP & Class-VC & Num-VC & Cal-LP & shared & McF $R^2$ \\
\midrule
SimpleQA & Gemma 3 27B  & corr & 0.56 [0.53,0.58] & 0.55 [0.52,0.57] & 0.62 [0.60,0.65] &  4.3 &  2.6 & 72.4 & 20.7 & 0.035 \\
         &              & abs  & 0.74 [0.72,0.76] & 0.75 [0.73,0.77] & 0.71 [0.69,0.73] & 33.4 & 14.0 &  8.8 & 43.8 & 0.284 \\
\addlinespace[2pt]
         & Gemma 4 31B  & corr & 0.52 [0.50,0.55] & 0.51 [0.48,0.53] & 0.66 [0.64,0.69] &  3.2 &  0.2 & 86.7 &  9.9 & 0.063 \\
         &              & abs  & 0.94 [0.91,0.97] & 0.77 [0.71,0.83] & 0.72 [0.66,0.77] & 29.0 & 10.8 &  4.3 & 55.9 & 0.567 \\
\addlinespace[2pt]
         & GPT-4o       & corr & 0.58 [0.55,0.60] & 0.60 [0.57,0.62] & 0.74 [0.72,0.76] &  0.2 &  0.2 & 81.8 & 17.8 & 0.137 \\
         &              & abs  & 0.76 [0.74,0.79] & 0.82 [0.81,0.84] & 0.76 [0.74,0.78] & 14.5 & 17.3 &  7.9 & 60.2 & 0.368 \\
\addlinespace[2pt]
         & Qwen 3 235B  & corr & 0.55 [0.49,0.60] & 0.62 [0.56,0.67] & 0.73 [0.69,0.78] &  2.0 &  0.2 & 79.3 & 18.5 & 0.128 \\
         &              & abs  & 0.62 [0.54,0.71] & 0.90 [0.85,0.95] & 0.63 [0.56,0.70] &  2.2 & 69.4 &  1.5 & 26.8 & 0.504 \\
\midrule
MMLU-Pro & Gemma 3 27B  & corr & 0.61 [0.55,0.67] & 0.46 [0.40,0.52] & 0.66 [0.61,0.72] & 12.7 &  1.2 & 58.1 & 28.0 & 0.093 \\
         &              & abs  & 0.64 [0.58,0.69] & 0.48 [0.43,0.53] & 0.64 [0.58,0.69] & 40.3 &  2.5 & 21.0 & 36.1 & 0.036 \\
\addlinespace[2pt]
         & Gemma 4 31B  & corr & 0.61 [0.55,0.68] & 0.49 [0.42,0.55] & 0.79 [0.74,0.84] &  2.3 &  0.9 & 82.3 & 14.5 & 0.218 \\
         &              & abs  & 0.97 [0.96,0.98] & 0.49 [0.49,0.50] & 0.81 [0.70,0.90] & 72.0 &  1.4 &  2.7 & 24.0 & 0.598 \\
\addlinespace[2pt]
         & GPT-4o       & corr & 0.62 [0.57,0.66] & 0.53 [0.48,0.58] & 0.73 [0.68,0.77] &  1.8 &  1.0 & 78.0 & 19.2 & 0.133 \\
         &              & abs  & 0.75 [0.70,0.79] & 0.72 [0.68,0.76] & 0.74 [0.69,0.78] & 28.2 & 13.8 & 12.3 & 45.7 & 0.301 \\
\addlinespace[2pt]
         & Qwen 3 235B  & corr & 0.64 [0.59,0.68] & 0.66 [0.62,0.71] & 0.71 [0.67,0.76] & 11.1 &  2.6 & 42.2 & 44.1 & 0.154 \\
         &              & abs  & 0.72 [0.67,0.77] & 0.89 [0.86,0.91] & 0.66 [0.62,0.71] &  3.8 & 56.9 &  0.1 & 39.2 & 0.425 \\
\bottomrule
\addlinespace[6pt]
\end{tabular}
\caption{\textbf{Full decomposition of confidence-signal contributions to predicting correctness and abstention under the neutral Phase~2 prompt, across non-reasoning models.}
Identical structure to Table~\ref{tab:nrm_auroc_variance} but using the neutral abstention prompt for Phase~2 (`commit to A vs.\ abstain to B') rather than the standard prompt (see Supplemental Methods). For each (dataset, model) cell, two rows: predicting Phase~1 correctness (corr; identical to standard-prompt table) and Phase~2 abstention (abs; neutral prompt). AUROCs and 95\% CIs from paired non-parametric bootstrap (1000 resamples). Percentage-unique-variance figures are McFadden $R^2$ decomposition components from the joint M$_{\text{VC+NumC+LP}}$ logit, expressed as a fraction of total $R^2$; shared variance pools all pairwise and three-way overlaps. AUROCs use single-signal valid samples; variance partitioning uses the joint sample where all three signals are valid. Gemma 4 31B and Qwen 3 235B are run in no-think mode.}
\label{tab:nrm_auroc_variance_neutral}
\end{table}
%----------------------DT gap CIs AUROCS comparison with standard
\begin{table}[t]
\centering
\footnotesize
\setlength{\tabcolsep}{3pt}
\renewcommand{\arraystretch}{1.1}
\begin{tabular}{llrrrrrr}
\toprule
 & & \multicolumn{3}{c}{Standard prompt} & \multicolumn{3}{c}{Neutral prompt} \\
\cmidrule(lr){3-5} \cmidrule(lr){6-8}
Dataset & Model & VC gap & NumC gap & Cal-LP gap & VC gap & NumC gap & Cal-LP gap \\
\midrule
SimpleQA & Gemma 3 27B          & $+0.23$ & $+0.23$ & $+0.14$ & $+0.18$ & $+0.19$ & $+0.09$ \\
         & Gemma 4 31B          & $+0.26$ & $+0.10$ & $-0.04$ & $+0.41$ & $+0.27$ & $+0.06$ \\
         & GPT-4o               & $+0.12$ & $+0.24$ & $+0.10$ & $+0.18$ & $+0.22$ & $+0.02$ \\
         & Qwen 3 235B no-think & $+0.10$ & $+0.25$ & $-0.06$ & $+0.07$ & $+0.27$ & $-0.10$ \\
\midrule
MMLU-Pro & Gemma 3 27B          & $+0.12$ & $-0.05$ & $+0.02$ & $+0.04$ & $-0.03$ & $-0.04$ \\
         & Gemma 4 31B          & $+0.18$ & $-0.01^{\ddagger}$ & $+0.02$ & $+0.33$ & $-0.01^{\ddagger}$ & $+0.01$ \\
         & GPT-4o               & $+0.14$ & $+0.20$ & $+0.02$ & $+0.14$ & $+0.17$ & $+0.01$ \\
         & Qwen 3 235B no-think & $+0.05$ & $+0.14$ & $-0.05$ & $+0.07$ & $+0.21$ & $-0.05$ \\
\bottomrule
\addlinespace[6pt]
\end{tabular}
\caption{\textbf{Decision-truth gaps under standard and neutral abstention prompts, across non-reasoning models.}
For each (dataset, model) cell, the decision-truth gap is AUROC$_{\text{abstention}} - $ AUROC$_{\text{correctness}}$, computed on the same trials. Verbal-confidence gaps (Class-VC, Num-VC) remain positive in nearly every cell under both prompt variants; Cal-LP gaps remain small or near-zero across prompts. The dissociation pattern is preserved across prompt framings: verbal confidence is consistently more decision-aligned than truth-aligned, while Cal-LP discriminates correctness and abstention more comparably. 
$^{\ddagger}$ Num-VC for Gemma 4 31B on MMLU-Pro is degenerate due to anchoring at the top of the scale (98.6\% of valid responses = 100), yielding near-zero gap under either prompt. Gemma 4 31B and Qwen 3 235B are run in no-think mode.}
\label{tab:nrm_dt_gap_comparison}
\end{table}

%-----__TABLES REASONING MODELS

%----metrics calibration 
\begin{table}[t]
\centering
\small
\setlength{\tabcolsep}{4pt}
\renewcommand{\arraystretch}{1.1}
\begin{tabular}{llrrr}
\toprule
 & & & \multicolumn{2}{c}{Calibration} \\
\cmidrule(lr){4-5}
Dataset & Model & $T^{*}$ & ECE & AUROC \\
\midrule
\multicolumn{5}{l}{\textit{Verbal confidence (class-based)}} \\
\addlinespace[2pt]
SimpleQA       & Gemini 3 Flash     & -- & 0.023 & 0.642 \\
               & Qwen 235B Think    & -- & 0.139 & 0.535 \\
               & Kimi K2 Think      & -- & 0.192 & 0.564 \\
               & GPT-Oss-120B       & -- & 0.089 & 0.673 \\
\addlinespace[2pt]
MMLU-Pro       & Gemini 3 Flash     & -- & 0.011 & 0.682 \\
               & Qwen 235B Think    & -- & 0.052 & 0.692 \\
               & Kimi K2 Think      & -- & 0.110 & 0.579 \\
               & GPT-Oss-120B       & -- & 0.052 & 0.778 \\
\addlinespace[2pt]
SuperGPQA-hard & Gemini 3 Flash     & -- & 0.192 & 0.633 \\
               & Qwen 235B Think    & -- & 0.270 & 0.639 \\
               & Kimi K2 Think      & -- & 0.271 & 0.588 \\
               & GPT-Oss-120B       & -- & 0.169 & 0.668 \\
\addlinespace[2pt]
HLE            & Gemini 3 Flash     & -- & 0.476 & 0.618 \\
               & Qwen 235B Think    & -- & 0.674 & 0.626 \\
               & Kimi K2 Think      & -- & 0.665 & 0.619 \\
               & GPT-Oss-120B       & -- & 0.431 & 0.591 \\
\midrule
\multicolumn{5}{l}{\textit{Calibrated log-probability (Cal-LP)}} \\
\addlinespace[2pt]
SimpleQA       & Gemini 3 Flash     & 3.3 & 0.015 & 0.686 \\
               & Qwen 235B Think    & 5.0 & 0.049 & 0.561 \\
               & Kimi K2 Think      & --  & --    & --    \\
               & GPT-Oss-120B       & --  & --    & --    \\
\addlinespace[2pt]
MMLU-Pro       & Gemini 3 Flash     & 2.6 & 0.029 & 0.792 \\
               & Qwen 235B Think    & 2.1 & 0.386 & 0.544 \\
               & Kimi K2 Think      & --  & --    & --    \\
               & GPT-Oss-120B       & --  & --    & --    \\
\addlinespace[2pt]
SuperGPQA-hard & Gemini 3 Flash     & 1.0 & 0.333 & 0.586 \\
               & Qwen 235B Think    & 3.9 & 0.288 & 0.536 \\
               & Kimi K2 Think      & --  & --    & --    \\
               & GPT-Oss-120B       & --  & --    & --    \\
\addlinespace[2pt]
HLE            & Gemini 3 Flash     & --$^{\dagger}$ & 0.022 & 0.553 \\
               & Qwen 235B Think    & --$^{\ddagger}$ & --    & --    \\
               & Kimi K2 Think      & --             & --    & --    \\
               & GPT-Oss-120B       & --             & --    & --    \\
\bottomrule
\addlinespace[6pt]
\end{tabular}
\caption{\textbf{Verbal and Cal-LP confidence signals across reasoning models.}
For each model and dataset, $T^{*}$ is the temperature fit on the Phase~0 option-token logprobs to minimise expected calibration error. ECE is 15-bin equal-width expected calibration error; AUROC measures discrimination of correct from incorrect Phase~1 answers. Cal-LP requires per-option logprobs at a checkpoint within or following the reasoning trace, which are available for Gemini~3~Flash on all four datasets and for Qwen~235B~Think on the three MCQ datasets; Kimi K2 Think and GPT-Oss-120B do not expose per-option logprobs and are therefore not included in the Cal-LP block.
$^{\dagger}$Gemini~3~Flash on HLE: freeform answers so Cal-LP is computed via Platt scaling; Platt parameters: $a{=}1.44$, $b{=}-0.46$.
For SimpleQA and MMLU-Pro (see Methods), Cal-LP is computed at the 95\% CoT checkpoint, where saturation (mean $p_{\max}$) is already high ($>$0.99). On SuperGPQA-hard the model has not committed by 95\% of its (longer) CoT (saturation 9\% at cp\_95 vs 71\% at cp\_100), so we compute Cal-LP at the 100\% (post-commit) checkpoint instead. At cp\_95 SuperGPQA-hard Cal-LP collapses to chance (AUROC $\approx 0.52$); at cp\_100 it is informative (AUROC $0.58$). VC is unaffected by checkpoint choice. Calibration parameters are fitted on Phase~0 (calibration set) and applied to Phase~1.}
\label{tab:rm_vc_callp_calibration}
\end{table}
%------ behavioral performance, conditional abs, with AUROC
\begin{table}[t]
\centering
\small
\setlength{\tabcolsep}{5pt}
\renewcommand{\arraystretch}{1.1}
\begin{tabular}{llrrrrrr}
\toprule
 & & & & & \multicolumn{2}{c}{$P(\text{abstain}\mid\cdot)$} & \\
\cmidrule(lr){6-7}
Dataset & Model & $N$ & Acc.\ & Abs.\ rate & correct & incorrect & $\Delta$ \\
\midrule
SimpleQA       & Gemini 3 Flash    & 1000 & 0.911 & 0.095 & 0.076 & 0.292 & $+0.216$ \\
               & Qwen 235B Think   &  500 & 0.812 & 0.087 & 0.076 & 0.132 & $+0.056$ \\
               & Kimi K2 Think     &  500 & 0.743 & 0.137 & 0.125 & 0.172 & $+0.047$ \\
               & GPT-Oss-120B      &  500 & 0.637 & 0.220 & 0.176 & 0.298 & $+0.122$ \\
\addlinespace[3pt]
MMLU-Pro       & Gemini 3 Flash    &  500 & 0.940 & 0.051 & 0.038 & 0.259 & $+0.221$ \\
               & Qwen 235B Think   &  499 & 0.871 & 0.046 & 0.026 & 0.177 & $+0.151$ \\
               & Kimi K2 Think     &  496 & 0.826 & 0.037 & 0.022 & 0.105 & $+0.082$ \\
               & GPT-Oss-120B      &  498 & 0.861 & 0.068 & 0.033 & 0.292 & $+0.259$ \\
\addlinespace[3pt]
SuperGPQA-hard & Gemini 3 Flash    &  500 & 0.680 & 0.108 & 0.068 & 0.193 & $+0.125$ \\
               & Qwen 235B Think   &  494 & 0.589 & 0.112 & 0.076 & 0.164 & $+0.088$ \\
               & Kimi K2 Think     &  489 & 0.658 & 0.068 & 0.041 & 0.120 & $+0.080$ \\
               & GPT-Oss-120B      &  485 & 0.565 & 0.200 & 0.115 & 0.310 & $+0.195$ \\
\addlinespace[3pt]
HLE            & Gemini 3 Flash    &  500 & 0.318 & 0.178 & 0.071 & 0.228 & $+0.157$ \\
               & Qwen 235B Think   &  492 & 0.088 & 0.251 & 0.070 & 0.269 & $+0.199$ \\
               & Kimi K2 Think     &  302 & 0.195 & 0.216 & 0.135 & 0.235 & $+0.101$ \\
               & GPT-Oss-120B      &  285 & 0.070 & 0.462 & 0.389 & 0.468 & $+0.079$ \\
\bottomrule
\addlinespace[6pt]
\end{tabular}
\caption{\textbf{Phase~1 accuracy and Phase~2 abstention behaviour across reasoning models.}
$N$ is the number of Phase~1 trials. Acc.\ is overall Phase~1 accuracy. Abs.\ rate is the overall Phase~2 abstention rate; the next two columns give abstention rates conditional on Phase~1 correctness, and $\Delta = P(\text{abstain}\mid\text{incorrect}) - P(\text{abstain}\mid\text{correct})$. All sixteen model~$\times$~dataset cells show higher abstention on incorrect than correct trials ($\Delta > 0$). Base accuracies span 0.07--0.94 across cells, and base abstention rates span 0.04--0.46, indicating that the truth-aligned gap is preserved despite substantial heterogeneity in model behaviour. Phase 2 includes the model's reasoning trace from Phase 1.}
\label{tab:rm_behavior}
\end{table}
%-----

%WITH KIMI
\begin{table}[t]
\centering
\footnotesize
\setlength{\tabcolsep}{2.5pt}
\renewcommand{\arraystretch}{1.1}
\begin{tabular}{llllllcrrrr}
\toprule
 & & & & \multicolumn{2}{c}{AUROC [95\% CI]} & D-T gap & \multicolumn{3}{c}{\% unique variance} & \\
\cmidrule(lr){5-6} \cmidrule(lr){8-10}
Dataset & Model & $N$ & Outcome & VC & Cal-LP & VC [CI] & VC & Cal-LP & shared & McF $R^2$ \\
\midrule
SimpleQA       & Gemini 3 Flash    & 833 & corr & 0.64 [0.59,0.70] & 0.69 [0.61,0.76] & --                  & 57.2 & 29.6 & 13.2 & 0.112 \\
               &                   &     & abs  & 0.68 [0.63,0.74] & 0.67 [0.59,0.73] & +0.041 [-0.029,+0.115] & 53.0 & 33.4 & 13.6 & 0.093 \\
\addlinespace[2pt]
               & Qwen 235B Think   & 485 & corr & 0.54 [0.50,0.58] & 0.56 [0.49,0.63] & --                  & 12.3 & 78.5 &  7.7 & 0.007 \\
               &                   &     & abs  & 0.69 [0.61,0.78] & 0.59 [0.49,0.69] & +0.155 [+0.067,+0.234] & 96.1 &  0.4 &  3.5 & 0.227 \\
\addlinespace[2pt]
               & Kimi K2 Think     & 495 & corr & 0.57 [0.52,0.61] & --               & --                  & --   & --   & --   & --    \\
               &                   &     & abs  & 0.69 [0.62,0.76] & --               & +0.126 [+0.050,+0.205] & --   & --   & --   & --    \\
\addlinespace[2pt]
               & GPT-Oss-120B      & 499 & corr & 0.67 [0.63,0.72] & --               & --                  & --   & --   & --   & --    \\
               &                   &     & abs  & 0.75 [0.70,0.80] & --               & +0.075 [+0.007,+0.139] & --   & --   & --   & --    \\
\midrule
MMLU-Pro       & Gemini 3 Flash    & 355 & corr & 0.68 [0.57,0.79] & 0.74 [0.64,0.82] & --                  & 93.6 &  0.3 &  6.1 & 0.124 \\
               &                   &     & abs  & 0.74 [0.63,0.84] & 0.71 [0.59,0.83] & +0.052 [-0.088,+0.176] & 79.4 &  8.8 & 11.7 & 0.134 \\
\addlinespace[2pt]
               & Qwen 235B Think   & 483 & corr & 0.70 [0.63,0.76] & 0.54 [0.47,0.61] & --                  & 96.5 &  1.5 &  2.0 & 0.094 \\
               &                   &     & abs  & 0.88 [0.79,0.96] & 0.51 [0.39,0.63] & +0.183 [+0.082,+0.275] & 99.6 &  0.4 &  0.0 & 0.337 \\
\addlinespace[2pt]
               & Kimi K2 Think     & 484 & corr & 0.58 [0.54,0.62] & --               & --                  & --   & --   & --   & --    \\
               &                   &     & abs  & 0.76 [0.64,0.87] & --               & +0.179 [+0.055,+0.305] & --   & --   & --   & --    \\
\addlinespace[2pt]
               & GPT-Oss-120B      & 488 & corr & 0.77 [0.70,0.84] & --               & --                  & --   & --   & --   & --    \\
               &                   &     & abs  & 0.96 [0.93,0.98] & --               & +0.186 [+0.117,+0.263] & --   & --   & --   & --    \\
\midrule
SuperGPQA-hard & Gemini 3 Flash    & 473 & corr & 0.63 [0.58,0.68] & 0.58 [0.53,0.63] & --                  & 86.6 &  8.6 &  4.8 & 0.069 \\
               &                   &     & abs  & 0.78 [0.72,0.85] & 0.62 [0.54,0.71] & +0.154 [+0.073,+0.234] & 93.7 &  3.0 &  3.3 & 0.176 \\
\addlinespace[2pt]
               & Qwen 235B Think   & 463 & corr & 0.64 [0.59,0.69] & 0.56 [0.50,0.62] & --                  & 90.7 &  8.8 &  0.5 & 0.071 \\
               &                   &     & abs  & 0.91 [0.85,0.96] & 0.52 [0.44,0.60] & +0.266 [+0.197,+0.330] & 100.0 &  0.4 & $-0.4$ & 0.493 \\
\addlinespace[2pt]
               & Kimi K2 Think     & 485 & corr & 0.59 [0.55,0.63] & --               & --                  & --   & --   & --   & --    \\
               &                   &     & abs  & 0.79 [0.71,0.88] & --               & +0.200 [+0.111,+0.289] & --   & --   & --   & --    \\
\addlinespace[2pt]
               & GPT-Oss-120B      & 457 & corr & 0.67 [0.63,0.72] & --               & --                  & --   & --   & --   & --    \\
               &                   &     & abs  & 0.84 [0.80,0.89] & --               & +0.171 [+0.108,+0.232] & --   & --   & --   & --    \\
\midrule
HLE            & Gemini 3 Flash    & 361 & corr & 0.62 [0.57,0.66] & 0.55 [0.50,0.62] & --                  & 63.7 & 32.2 &  3.4 & 0.015 \\
               &                   &     & abs  & 0.71 [0.66,0.77] & 0.50 [0.42,0.58] & +0.093 [+0.024,+0.164] & 99.3 &  0.0 &  0.7 & 0.104 \\
\addlinespace[2pt]
               & Qwen 235B Think   & 482 & corr & 0.62 [0.53,0.71] & --               & --                  & --   & --   & --   & --    \\
               &                   &     & abs  & 0.79 [0.74,0.83] & --               & +0.160 [+0.071,+0.250] & --   & --   & --   & --    \\
\addlinespace[2pt]
               & Kimi K2 Think     & 261 & corr & 0.62 [0.56,0.66] & --               & --                  & --   & --   & --   & --    \\
               &                   &     & abs  & 0.54 [0.48,0.61] & --               & $-$0.078 [$-$0.158,+0.002] & --   & --   & --   & --    \\
\addlinespace[2pt]
               & GPT-Oss-120B      & 264 & corr & 0.59 [0.44,0.74] & --               & --                  & --   & --   & --   & --    \\
               &                   &     & abs  & 0.83 [0.79,0.88] & --               & +0.245 [+0.094,+0.409] & --   & --   & --   & --    \\
\bottomrule
\addlinespace[6pt]
\end{tabular}
\caption{\textbf{Full decomposition of VC and Cal-LP contributions to predicting correctness and abstention, across reasoning models.}
For each (dataset, model) cell, two rows: predicting Phase~1 correctness (corr) and Phase~2 abstention (abs). AUROCs and 95\% CIs from paired non-parametric bootstrap (1000 resamples). Decision-truth gap (D-T gap) is $\text{AUROC}_{\text{abstain}} - \text{AUROC}_{\text{correct}}$ for VC, computed on the same trials with paired-bootstrap 95\% CIs; positive values indicate decision-aligned signal. The D-T gap is reported on the abstention row only. Percentage-unique-variance figures are McFadden $R^2$ decomposition components from the joint M$_{\text{VC+LP}}$ logit, expressed as a fraction of total $R^2$; shared variance pools VC$\times$Cal-LP overlap. Total McFadden $R^2$ shown in the last column for transparency. AUROCs use single-signal valid samples; variance partitioning uses the joint sample where both signals are valid; $N$ refers to the joint sample for Cal-LP-available cells, otherwise to the VC-valid sample. Cells without Cal-LP: Kimi K2 Think and GPT-Oss-120B (all four datasets) and Qwen 235B Think on HLE; per-option logprobs at a chain-of-thought checkpoint are not exposed in the available run files for these cells.}
\label{tab:rm_auroc_variance}
\end{table}

%-------------RM residual analysis
%NEW WITH KIMI
\begin{table}[t]
\centering
\footnotesize
\setlength{\tabcolsep}{4pt}
\renewcommand{\arraystretch}{1.1}
\begin{tabular}{llrlrl}
\toprule
 & & \multicolumn{2}{c}{AUROC$_\mathrm{residual}$ [95\% CI]} & \multicolumn{2}{c}{\% preserved} \\
\cmidrule(lr){3-4} \cmidrule(lr){5-6}
Dataset & Model & corr only & corr + Cal-LP & corr & + Cal-LP \\
\midrule
SimpleQA       & Gemini 3 Flash    & 0.60 [0.53, 0.67] & 0.57 [0.49, 0.66] & 52 & 36 \\
               & Qwen 235B Think   & 0.67 [0.58, 0.79] & 0.66 [0.56, 0.78] & 89 & 84 \\
               & Kimi K2 Think     & 0.68 [0.60, 0.75] & --                & 94 & -- \\
               & GPT-Oss-120B      & 0.74 [0.68, 0.79] & --                & 95 & -- \\
\addlinespace[3pt]
MMLU-Pro       & Gemini 3 Flash    & 0.70 [0.57, 0.84] & 0.68 [0.52, 0.83] & 85 & 78 \\
               & Qwen 235B Think   & 0.73 [0.62, 0.94] & 0.85 [0.62, 0.94] & 59 & 90 \\
               & Kimi K2 Think     & 0.68 [0.47, 0.85] & --                & 71 & -- \\
               & GPT-Oss-120B      & 0.88 [0.77, 0.96] & --                & 83 & -- \\
\addlinespace[3pt]
SuperGPQA-hard & Gemini 3 Flash    & 0.78 [0.69, 0.84] & 0.75 [0.67, 0.83] & 99 & 87 \\
               & Qwen 235B Think   & 0.90 [0.84, 0.96] & 0.90 [0.85, 0.96] & 99 & 98 \\
               & Kimi K2 Think     & 0.76 [0.64, 0.87] & --                & 89 & -- \\
               & GPT-Oss-120B      & 0.81 [0.77, 0.87] & --                & 91 & -- \\
\addlinespace[3pt]
HLE            & Gemini 3 Flash    & 0.70 [0.63, 0.75] & 0.68 [0.60, 0.74] & 93 & 95 \\
               & Qwen 235B Think   & 0.78 [0.72, 0.82] & --                & 97 & -- \\
               & Kimi K2 Think     & 0.52 [0.44, 0.60] & --                & 45$^{\ddagger}$ & -- \\
               & GPT-Oss-120B      & 0.84 [0.78, 0.88] & --                & 101 & -- \\
\bottomrule
\addlinespace[6pt]
\end{tabular}
\caption{\textbf{Residual analysis: VC's abstention-prediction signal after stripping correctness-related variance, across reasoning models.}
For each (model, dataset) cell we residualised VC against (i) binary \texttt{is\_correct} alone (\emph{corr only}); and (ii) \texttt{is\_correct} plus Cal-LP (\emph{corr + Cal-LP}, where computable). For each residualisation we computed the AUROC of the residual VC for predicting Phase 2 abstention (paired-bootstrap 95\% CI from 1000 resamples), and the percentage of raw abstention-prediction power preserved relative to chance, $(\mathrm{AUROC}_\mathrm{residual} - 0.5) / (\mathrm{AUROC}_\mathrm{raw} - 0.5) \times 100$. Cells without Cal-LP (Kimi K2 Think and GPT-Oss on all four datasets; Qwen on HLE) report only the binary-correctness specification. For SimpleQA, an additional specification including a per-question difficulty score changed preserved fraction by 6 percentage points or fewer relative to corr+LP.
$^{\ddagger}$ Kimi K2 Think on HLE: raw VC AUROC for abstention is near chance (0.54), so the \% preserved metric is unreliable.}
\label{tab:rm_residual_abstention}
\end{table}

%----------eta squared cross cell analysis RMs
%NEW with KIMI
\begin{table}[t]
\centering
\footnotesize
\setlength{\tabcolsep}{5pt}
\renewcommand{\arraystretch}{1.1}
\begin{tabular}{llrrrr}
\toprule
 &  & \multicolumn{2}{c}{VC} & \multicolumn{2}{c}{Cal-LP} \\
\cmidrule(lr){3-4} \cmidrule(lr){5-6}
Dataset & Model & corr.\ $\eta^{2}$ & dec.\ $\eta^{2}$ & corr.\ $\eta^{2}$ & dec.\ $\eta^{2}$ \\
\midrule
SimpleQA       & Gemini 3 Flash    & 0.06 & 0.04 & 0.03 & 0.03 \\
               & Qwen 235B Think   & 0.00 & 0.25 & 0.01 & 0.01 \\
               & Kimi K2 Think     & 0.00 & 0.14 & --   & --   \\
               & GPT-Oss-120B      & 0.07 & 0.12 & --   & --   \\
\addlinespace[3pt]
MMLU-Pro       & Gemini 3 Flash    & 0.08 & 0.07 & 0.00 & 0.01 \\
               & Qwen 235B Think   & 0.06 & 0.27 & 0.00 & 0.00 \\
               & Kimi K2 Think     & 0.04 & 0.10 & --   & --   \\
               & GPT-Oss-120B      & 0.06 & 0.38 & --   & --   \\
\addlinespace[3pt]
SuperGPQA-hard & Gemini 3 Flash    & 0.06 & 0.14 & 0.01 & 0.01 \\
               & Qwen 235B Think   & 0.08 & 0.52 & 0.01 & 0.00 \\
               & Kimi K2 Think     & 0.01 & 0.12 & --   & --   \\
               & GPT-Oss-120B      & 0.05 & 0.27 & --   & --   \\
\addlinespace[3pt]
HLE            & Gemini 3 Flash    & 0.01 & 0.11 & 0.01 & 0.00 \\
               & Qwen 235B Think   & 0.00 & 0.24 & --   & --   \\
               & Kimi K2 Think     & 0.02 & 0.00 & --   & --   \\
               & GPT-Oss-120B      & 0.00 & 0.35 & --   & --   \\
\bottomrule
\addlinespace[6pt]
\end{tabular}
\caption{\textbf{Partial $\eta^{2}$ from a two-way ANOVA on each confidence signal, by reasoning model and dataset.}
Each cell reports the proportion of variance in the signal explained by correctness (corr.) and by the abstention decision (dec.), after accounting for the other factor and their interaction. For VC, the decision-related effect substantially exceeds the correctness-related effect in 13/16 cells (median decision $\eta^{2}=0.14$ vs.\ median correctness $\eta^{2}=0.05$). For Cal-LP, partial $\eta^{2}$ is small for both factors in every cell where Cal-LP is computable (median 0.006 for correctness, 0.005 for decision). Dashes indicate cells without Cal-LP available (Kimi K2 Think and GPT-Oss on all four datasets; Qwen on HLE).}
\label{tab:rm_eta2_effect_sizes}
\end{table}

%---------Mixed effects table to go with scatters
\begin{table}[ht]
\centering
\small
\caption{\textbf{Mixed-effects regression of confidence-signal AUROCs and decision-truth gap, with model as random intercept.} Fixed effects for dataset (4 levels: HLE, MMLU-Pro, SimpleQA, SuperGPQA-hard; HLE as reference), prompt (2 levels: neutral, standard; neutral as reference), and for VC only, confidence format (labelled vs.\ numeric; labelled as reference). The key fixed effect is \texttt{vc\_corr\_c} (or \texttt{lp\_corr\_c}), the mean-centred correctness AUROC. Q1 models predict abstention AUROC; Q2 models predict the decision-truth gap (abstention~$-$~correctness AUROC). For Cal-LP, the format covariate is omitted (Cal-LP is unitary). VC cells: $n = 46$ across 8 models; Cal-LP cells: $n = 23$ across 6 models. Reasoning models contribute standard-prompt, labelled-VC cells only. Bootstrap analyses are by-cell; mixed model is REML.}
\label{tab:mixed-effects}
\setlength{\tabcolsep}{6pt}
\begin{tabular}{l c c c c c c c}
\toprule
 & \multicolumn{3}{c}{\textbf{VC} ($n=46$, 8 models)} & & \multicolumn{3}{c}{\textbf{Cal-LP} ($n=23$, 6 models)} \\
\cmidrule(lr){2-4} \cmidrule(lr){6-8}
Predictor & $\beta$ & 95\% CI & $p$ & & $\beta$ & 95\% CI & $p$ \\
\midrule
\multicolumn{8}{l}{\textit{Q1: Abstention AUROC}} \\
\midrule
Correctness AUROC (centred) & $+1.31$ & $[+0.72, +1.89]$ & $<0.001$ & & $+0.91$ & $[+0.43, +1.39]$ & $<0.001$ \\
Dataset: MMLU-Pro             & $-0.00$ & $[-0.10, +0.10]$ & $0.961$ & & $+0.05$ & $[-0.08, +0.19]$ & $0.447$ \\
Dataset: SimpleQA             & $+0.05$ & $[-0.05, +0.15]$ & $0.306$ & & $+0.08$ & $[-0.05, +0.21]$ & $0.220$ \\
Dataset: SuperGPQA-hard       & $+0.09$ & $[-0.03, +0.20]$ & $0.149$ & & $+0.06$ & $[-0.07, +0.19]$ & $0.350$ \\
Prompt: standard              & $-0.03$ & $[-0.09, +0.02]$ & $0.253$ & & $+0.01$ & $[-0.04, +0.06]$ & $0.795$ \\
Format: numeric               & $+0.03$ & $[-0.03, +0.09]$ & $0.318$ & & --- & --- & --- \\
\midrule
\multicolumn{8}{l}{\textit{Q2: Decision-truth gap}} \\
\midrule
Correctness AUROC (centred) & $+0.31$ & $[-0.28, +0.89]$ & $0.303$ & & $-0.09$ & $[-0.57, +0.39]$ & $0.704$ \\
\midrule
ICC (between-model variance) & \multicolumn{3}{c}{$0.194$} & & \multicolumn{3}{c}{$0.413$} \\
\bottomrule
\end{tabular}
\end{table}

\end{document}